\pdfoutput=1
\documentclass[11pt,letterpaper]{article}
\usepackage[T1]{fontenc}
\usepackage[utf8]{inputenc}

\usepackage{amsmath,amssymb,amsthm,mathtools}
\usepackage{newpxtext}                 % Palatino text
\usepackage{newpxmath}                 % Palatino math
\usepackage[scaled=0.90]{FiraSans}     % headings, labels, title block

\usepackage[letterspace=80]{microtype}

\usepackage[letterpaper,left=1.15in,right=1.15in,top=1.1in,bottom=1.15in,headsep=0.32in,footskip=0.45in]{geometry}
\usepackage{xcolor}
\definecolor{ink}{HTML}{1A1F2B}
\definecolor{navy}{HTML}{1F3A63}         % headings, labels
\definecolor{refcool}{HTML}{2C5AA0}      % cross-references, numbers, theorem bars
\definecolor{citewarm}{HTML}{B5403B}     % citations, subtitle, definition bars
\definecolor{muted}{HTML}{6B6A66}        % running heads, meta text
\definecolor{rulegray}{HTML}{D3D6DC}
\definecolor{thmback}{HTML}{F2F6FC}
\definecolor{lemback}{HTML}{F5F6F8}
\definecolor{lembar}{HTML}{6B7A90}
\definecolor{absback}{HTML}{F7F8FA}

\usepackage{graphicx,booktabs,array,tabularx,multicol}
\usepackage{enumitem}
\setlist{itemsep=1pt,parsep=0pt,topsep=3pt,partopsep=0pt}
\usepackage[font=small,labelfont={sf,bf,color=navy},labelsep=period,skip=6pt]{caption}
\usepackage[round]{natbib}

\usepackage[most]{tcolorbox}
\usepackage{titlesec,titletoc}
\usepackage{fancyhdr}
\usepackage{amsfonts}
\usepackage{needspace,etoolbox,placeins}
\usepackage{multirow,hhline}
\usepackage{colortbl}
\usepackage{algorithm}
\usepackage[noend]{algpseudocode}
\usepackage{tikz}
\usepackage{pgfplots}
\usepackage{refcount}

\titleformat{\section}{\sffamily\Large\bfseries\color{navy}}{\textcolor{refcool}{\thesection}}{0.7em}{}
\titlespacing*{\section}{0pt}{3.2ex plus 1ex minus .2ex}{1.5ex plus .2ex}
\titleformat{\subsection}{\sffamily\large\bfseries\color{navy}}{\textcolor{refcool}{\thesubsection}}{0.6em}{}
\titlespacing*{\subsection}{0pt}{2.4ex plus .8ex minus .2ex}{1.0ex plus .2ex}
\titleformat{\subsubsection}{\sffamily\normalsize\bfseries\color{navy}}{\textcolor{refcool}{\thesubsubsection}}{0.5em}{}
\titleformat{\paragraph}[runin]{\sffamily\bfseries\color{navy}}{}{0pt}{}
\titlespacing*{\paragraph}{0pt}{1.4ex plus .5ex minus .2ex}{0.55em}

\titlecontents{section}[2.0em]{\addvspace{0.6pt}\sffamily\fontsize{8.4}{10.2}\selectfont}
  {\contentslabel[\textcolor{refcool}{\bfseries\thecontentslabel}]{2.0em}}{\hspace*{-2.0em}}
  {\titlerule*[0.6pc]{\textcolor{rulegray}{.}}\contentspage}

\fancypagestyle{first}{\fancyhf{}\fancyfoot[C]{\sffamily\footnotesize\color{muted}\thepage}}

\newtheoremstyle{boxed}{0pt}{0pt}{\itshape}{}{\sffamily\bfseries\color{navy}}{.}{0.5em}
  {\thmname{#1}\thmnumber{ #2}\thmnote{\ \textmd{(#3)}}}
\newtheoremstyle{boxedrm}{0pt}{0pt}{\normalfont}{}{\sffamily\bfseries\color{navy}}{.}{0.5em}
  {\thmname{#1}\thmnumber{ #2}\thmnote{\ \textmd{(#3)}}}
\theoremstyle{boxed}
\newtheorem{theorem}{Theorem}[section]
\newtheorem{proposition}[theorem]{Proposition}
\newtheorem{lemma}[theorem]{Lemma}
\newtheorem{corollary}[theorem]{Corollary}

\theoremstyle{boxedrm}
\newtheorem{definition}[theorem]{Definition}
\newtheorem{assumption}[theorem]{Assumption}
\newtheorem{remark}[theorem]{Remark}

\tcbset{thmbox/.style={enhanced,breakable,frame hidden,sharp corners,boxrule=0pt,
  left=8pt,right=8pt,top=5pt,bottom=5pt,before skip=9pt,after skip=9pt,lines before break=5}}
\tcbset{coolbox/.style={thmbox,colback=thmback,borderline west={2.4pt}{0pt}{refcool}}}
\tcbset{slatebox/.style={thmbox,colback=lemback,borderline west={2.4pt}{0pt}{lembar}}}
\tcbset{warmbox/.style={thmbox,colback=white,borderline west={2.4pt}{0pt}{citewarm}}}
\tcbset{linebox/.style={blanker,breakable,left=8pt,before skip=7pt,after skip=7pt,borderline west={1pt}{0pt}{rulegray}}}
\tcolorboxenvironment{theorem}{coolbox}
\tcolorboxenvironment{proposition}{coolbox}
\tcolorboxenvironment{corollary}{coolbox}
\tcolorboxenvironment{conjecture}{coolbox}
\tcolorboxenvironment{lemma}{slatebox}
\tcolorboxenvironment{claim}{slatebox}
\tcolorboxenvironment{fact}{slatebox}
\tcolorboxenvironment{observation}{slatebox}
\tcolorboxenvironment{definition}{warmbox}
\tcolorboxenvironment{assumption}{warmbox}
\tcolorboxenvironment{remark}{linebox}
\tcolorboxenvironment{example}{linebox}
\tcolorboxenvironment{problem}{linebox}
\tcolorboxenvironment{question}{linebox}

\makeatletter
\renewenvironment{proof}[1][\proofname]{\par\pushQED{\qed}\normalfont\topsep6\p@\@plus6\p@\relax
  \trivlist\item[\hskip\labelsep{\sffamily\bfseries\color{navy}#1.}]\ignorespaces}{\popQED\endtrivlist\@endpefalse}
\makeatother

\makeatletter
\usepackage{thm-restate}
\usetikzlibrary{arrows.meta,decorations.pathreplacing,calc,positioning}
\pgfplotsset{compat=1.18}
\usepgfplotslibrary{groupplots,fillbetween}
\definecolor{bluec}{HTML}{2566C4}
\definecolor{aquac}{HTML}{12A07A}
\definecolor{orangec}{HTML}{E8622D}
\definecolor{amberc}{HTML}{C28000}
\definecolor{bluetext}{HTML}{1B4F99}
\definecolor{aquatext}{HTML}{0B7A5C}
\definecolor{orangetext}{HTML}{B5471F}
\definecolor{ambertext}{HTML}{8F5F00}
\definecolor{inkc}{HTML}{1D2733}
\definecolor{ink2c}{HTML}{4C5664}
\definecolor{mutec}{HTML}{8B939C}
\definecolor{axisc}{HTML}{A4ACB5}
\definecolor{gridc}{HTML}{E6E9ED}
\definecolor{rulec}{HTML}{D3D8DE}
\definecolor{panelc}{HTML}{F6F7F9}
\definecolor{headc}{HTML}{ECEFF3}
\definecolor{stmtc}{gray}{0.92}
\colorlet{tealc}{aquac}
\colorlet{coralc}{orangec}
\colorlet{goldc}{amberc}
\definecolor{refwarm}{HTML}{B5471F}
\definecolor{citecool}{HTML}{1D56A6}
\definecolor{urlcool}{HTML}{0E7F63}
\newcommand{\figtitlefont}{\fontsize{7.6}{9}\selectfont}
\newcommand{\figlabelfont}{\fontsize{7}{8.2}\selectfont}
\newcommand{\figtickfont}{\fontsize{6.4}{7.4}\selectfont}
\newcommand{\figannotfont}{\fontsize{6.6}{7.6}\selectfont}
\newcommand{\paneltitle}[2]{\textbf{#1}\hspace{0.45em}#2}
\pgfplotsset{
  paper/.style={
    scale only axis,
    axis x line*=bottom, axis y line*=left,
    axis line style={axisc, line width=.45pt},
    every tick/.style={axisc, line width=.45pt},
    major tick length=2.2pt, minor tick length=1.2pt, tick align=outside,
    ymajorgrids, grid style={gridc, line width=.4pt},
    label style={font=\figlabelfont, text=ink2c},
    xlabel style={yshift=1.5pt}, ylabel style={yshift=-1pt},
    tick label style={font=\figtickfont, text=ink2c},
    title style={font=\figtitlefont, text=inkc, anchor=base west, at={(0,1)}, xshift=-16pt, yshift=4.5pt},
    every axis plot/.append style={line join=round, line cap=round},
    legend style={draw=none, fill=white, fill opacity=.9, text opacity=1, font=\figannotfont, text=ink2c,
      inner sep=1pt, row sep=-1pt, legend cell align=left},
    legend image post style={scale=.8},
    clip mode=individual,
  },
  data/.style={line width=1pt},
  refline/.style={mutec, line width=.6pt, densely dashed, forget plot},
  ring/.style={mark options={draw=white, line width=.35pt, solid}},
  dot/.style={mark=*, mark size=1.7pt, ring},
  diam/.style={mark=diamond*, mark size=2.1pt, ring},
  sq/.style={mark=square*, mark size=1.45pt, ring},
  opendot/.style={mark=o, mark size=1.6pt, mark options={solid, fill=white, line width=.7pt}},
  opendiam/.style={mark=diamond, mark size=2.1pt, mark options={solid, fill=white, line width=.7pt}},
}
\tikzset{
  dlabel/.style={font=\figannotfont, inner sep=1pt},
  dlabel blue/.style={dlabel, text=bluetext},
  dlabel aqua/.style={dlabel, text=aquatext},
  dlabel orange/.style={dlabel, text=orangetext},
  dlabel amber/.style={dlabel, text=ambertext},
  dlabel mute/.style={dlabel, text=mutec},
}
\newcommand{\headrow}{\rowcolor{headc}}
\newcommand{\beat}{\cellcolor{aquac!13}}
\newcolumntype{L}[1]{>{\raggedright\arraybackslash}p{#1}}
\AddToHook{env/table/begin}{\setlength{\aboverulesep}{0pt}\setlength{\belowrulesep}{0pt}%
  \setlength{\extrarowheight}{2pt}\setlength{\arrayrulewidth}{.5pt}}
\tcbset{statementbox/.style={enhanced,breakable,frame hidden,colback=stmtc,sharp corners,
  grow sidewards by=5pt,left=5pt,right=5pt,top=2pt,bottom=2pt,boxsep=0pt,
  before skip=4pt,after skip=4pt,lines before break=4,
  before upper={\setlength{\topsep}{0pt}\setlength{\partopsep}{0pt}}}}
\newcommand{\boxstatement}[1]{%
  \expandafter\let\csname stmtbox@begin@#1\expandafter\endcsname\csname #1\endcsname
  \expandafter\let\csname stmtbox@end@#1\expandafter\endcsname\csname end#1\endcsname
  \expandafter\def\csname #1\endcsname{\begin{tcolorbox}[statementbox]\csname stmtbox@begin@#1\endcsname}%
  \expandafter\def\csname end#1\endcsname{\csname stmtbox@end@#1\endcsname\end{tcolorbox}}}
\newcommand{\itemhead}[1]{\textbf{\upshape #1}}
\newcommand{\proofstep}[1]{\textit{#1}\enspace}
\newcommand{\E}{\mathbb{E}}
\newcommand{\Pp}{\mathbb{P}}
\newcommand{\R}{\mathbb{R}}

\newcommand{\calT}{\mathcal{T}}
\newcommand{\calU}{\mathcal{U}}

\newcommand{\calZ}{\mathcal{Z}}
\newcommand{\KL}{\mathrm{KL}}
\newcommand{\MI}{I}
\newcommand{\gap}{\mathrm{gap}}
\newcommand{\keff}{k_{\mathrm{eff}}}
\newcommand{\Lhat}{\widehat L}
\newcommand{\Vhat}{\widehat V}
\newcommand{\muhat}{\widehat\mu}
\newcommand{\Qbar}{\bar Q}
\newcommand{\Kbar}{\bar K}
\newcommand{\Pstore}{\mathbf{P}}
\newcommand{\supp}{\operatorname{supp}}
\newcommand{\defeq}{\vcentcolon=}
\newcommand{\dd}{\mathrm{d}}
\newcommand{\cond}{\,\vert\,}
\DeclareMathOperator*{\argmin}{arg\,min}
\DeclareMathOperator{\Var}{Var}
\DeclareMathOperator*{\esssup}{ess\,sup}
\newcommand{\Leak}{\mathcal{L}}
\makeatother

\usepackage[colorlinks=true,linkcolor=refcool,citecolor=citewarm,urlcolor=refcool,
  pdftitle={The Price of Choosing Examples: Context Overfitting in Adaptive In-Context Learning},pdfauthor={Xiaoyu Li, Jiangxuan Long},pdfsubject={Context Overfitting in Adaptive In-Context Learning},pdfkeywords={}]{hyperref}
\usepackage[nameinlink,capitalise,noabbrev]{cleveref}
\crefname{assumption}{Assumption}{Assumptions}
\crefname{appendix}{Appendix}{Appendices}
\crefname{algorithm}{Algorithm}{Algorithms}

\begin{document}
\thispagestyle{first}

\noindent{\color{refcool}\rule{\linewidth}{2.2pt}}\par\vspace{7pt}
\noindent{\sffamily\footnotesize\color{muted}\lsstyle NeurIPS 2026 $\cdot$ Extended Version\quad\textbullet\quad \ifcase\month\or January\or February\or March\or April\or May\or June\or July\or August\or September\or October\or November\or December\fi\ \number\year}\par\vspace{8pt}
\noindent{\sffamily\bfseries\fontsize{23}{28}\selectfont\color{navy}The Price of Choosing Examples\par}\vspace{7pt}
\noindent{\sffamily\large\color{citewarm}Context Overfitting in Adaptive In-Context Learning\par}\vspace{8pt}
\noindent\hspace*{0\linewidth}\begin{tabular*}{0.74\linewidth}{@{}l@{\extracolsep{\fill}}l@{}}
{\sffamily\bfseries\color{ink}Xiaoyu Li$^{*}$} & {\sffamily\bfseries\color{ink}Jiangxuan Long$^{*}$} \\[1pt]
{\small University of New South Wales} & {\small The University of Hong Kong} \\[1pt]
{\sffamily\footnotesize\href{mailto:xiaoyu.li2@unsw.edu.au}{\color{muted}xiaoyu.li2@unsw.edu.au}} & {\sffamily\footnotesize\href{mailto:jxlong@connect.hku.hk}{\color{muted}jxlong@connect.hku.hk}}
\end{tabular*}

\begingroup\renewcommand\thefootnote{}\footnotetext{$^{*}$Equal contribution. A preliminary version of this paper was accepted at the 40th Conference on Neural Information Processing Systems (NeurIPS 2026).}\endgroup

\vspace{5pt}

\begin{tcolorbox}[enhanced,colback=absback,frame hidden,sharp corners,boxrule=0pt,
  borderline west={3pt}{0pt}{refcool},left=11pt,right=11pt,top=6pt,bottom=6pt]
{\sffamily\bfseries\footnotesize\color{navy}\lsstyle ABSTRACT}\par\vspace{3pt}
\small Evaluating a prompt on the same examples used to choose it biases its reported loss downward, a phenomenon we
call \emph{context overfitting}. We study this gap by comparing a chosen prompt's deployment risk with its
\emph{selected score}, the empirical loss over the $k$ examples used in its evaluation, whether reused validation
queries or demonstrations scored inside the prompt itself. Every adaptive pipeline is captured by a \emph{transcript}
fixing both the predictor and the scored indices. Pairing that transcript with an independent copy of the scoring store
defines a ghost law under which the selected score concentrates at scale $1/\sqrt{k}$; transporting back to the actual
pipeline charges an information price, the \emph{selection capacity} $\kappa$, yielding expected, high-probability, and
PAC-Bayesian certificates at scale $\sqrt{\kappa/k}$. We prove that this rate is minimax optimal for every ambient store
size $n\ge k$, establish its exact leading constant in the moderate-capacity regime, and show that without a capacity
constraint deterministic cherry-picking forces a nearly vacuous $1-O(k/n)$ correction. A compositional \emph{capacity
calculus} budgets $\kappa$ directly from pipeline logs across candidate menus, data-dependent shortlists, noisy score
releases, and multi-round edits. Optimizing the PAC-Bayesian certificate itself yields \emph{stability-constrained
adaptive randomization} (SCAR), a Gibbs selector that trades empirical fit against divergence from a fixed prior. Across
24{,}000 selection trials on Banking77 spanning 64 prompt recipes and two model families (Qwen2.5-7B-Instruct and
GPT-4.1-mini), empirical coverage is $100\%$, and SCAR achieves the tightest certified risk at every sample size,
certifying gains over zero-shot deployment risk on more than $99\%$ of validation draws from $k=64$ and $k=128$
examples, respectively.\par

\end{tcolorbox}

\vspace{2pt}
\begin{tcolorbox}[enhanced,colback=white,colframe=rulegray,boxrule=0.5pt,arc=2pt,left=10pt,right=10pt,top=5pt,bottom=4pt]
{\sffamily\bfseries\footnotesize\color{navy}\lsstyle CONTENTS}\par
\begin{multicols}{2}
{\hypersetup{linkcolor=ink}\makeatletter\@starttoc{toc}\makeatother}
\end{multicols}
\end{tcolorbox}
\suppressfloats[t]
\clearpage

\section{Introduction}\label{sec:intro}

Evaluate $M$ prompt templates on the same $k$ validation examples and deploy the winner. Even if every template carries
identical population risk, taking the minimum converts random fluctuations in the validation losses into an apparent
improvement of roughly $\sigma\sqrt{2\log M/k}$, where $\sigma$ is the single-example standard deviation. Modern
in-context and retrieval-augmented pipelines \citep{brown2020language,garg2022what} amplify this reuse across every
stage of design: practitioners tune retrieval metrics, filter and reorder demonstrations, refine system instructions,
and benchmark successive iterations on a shared evaluation split. We call the resulting downward bias in a chosen
prompt's reported loss \emph{context overfitting}.

\paragraph{A running example.}
Suppose a practitioner compares 64 prompt recipes, each specifying an instruction template, a retrieval rule, and a
demonstration layout. Every recipe retrieves $m$ demonstrations per query from a bank fixed prior to validation, is
scored on the same $k$ validation queries, and yields a deployment candidate via the lowest validation error. Two sample
sizes play distinct roles here: the demonstration count $m$ shapes the predictor, whereas the \emph{scoring sample size}
$k$ governs the statistical precision of its reported error. More generally, an adaptive pipeline may select both the
prompt and the subset or weighting of examples on which its performance is reported. The central question of this paper
is:
\begin{center}
\emph{How much correction does a reported score require after adaptive prompt and example selection?}
\end{center}

Existing theory characterizes what a frozen transformer computes \emph{given} a context: it identifies when attention
implements gradient-based, ridge, or Bayesian estimators
\citep{akyurek2022what,vonoswald2023transformers,he2025demystified}, derives asymptotic risk curves
\citep{lu2025asymptotic,bordelon2025scaling}, and bounds estimation error under a prescribed retrieval law
\citep{guo2025rag}. Our object of study is orthogonal. Rather than analyzing the forward pass on a fixed context, we
price the statistical optimism induced by \emph{choosing} the context and its evaluation data, comparing deployment risk
with the empirical loss on the examples reused in selection, which we call the \emph{selected score}.

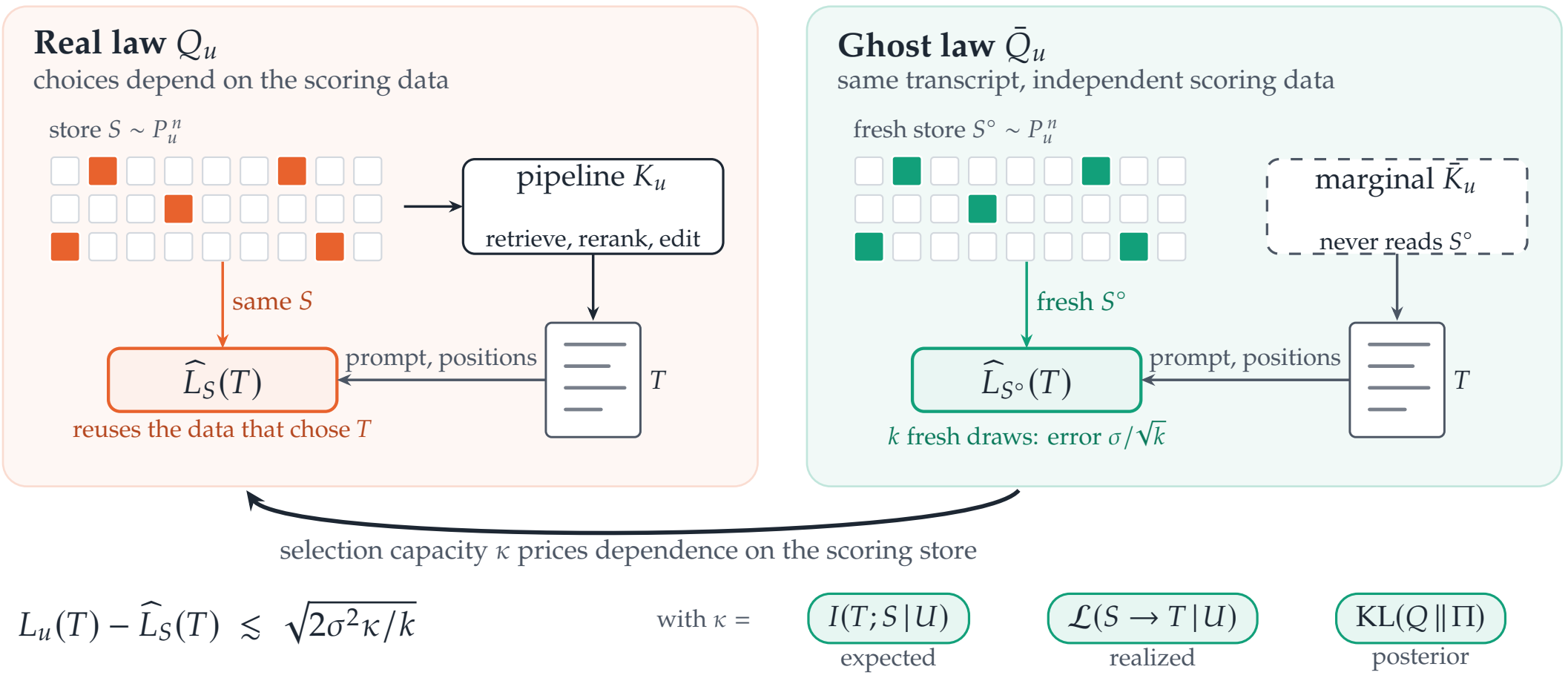
\begin{figure}[t]
  \centering
  \begin{tikzpicture}[x=1pt,y=1pt,
    every node/.style={inner sep=0pt,outer sep=0pt,text=inkc},
    panel/.style={rounded corners=6pt,line width=.5pt},
    cell/.style={draw=rulec,line width=.4pt,fill=white,minimum size=7pt,rounded corners=.8pt},
    rsel/.style={draw=white,line width=.5pt,fill=orangec,minimum size=7.6pt,rounded corners=.8pt},
    gsel/.style={draw=white,line width=.5pt,fill=aquac,minimum size=7.6pt,rounded corners=.8pt},
    arr/.style={-{Stealth[length=3.6pt,width=3pt]},line width=.65pt},
    box/.style={rounded corners=3pt,line width=.6pt,minimum width=66pt,minimum height=24pt,align=center},
    score/.style={rounded corners=3pt,line width=.8pt,minimum width=58pt,minimum height=16pt,
                  font=\fontsize{8.4}{9}\selectfont},
    head/.style={font=\fontsize{8.6}{10}\selectfont,anchor=west},
    sub/.style={font=\fontsize{6.8}{8}\selectfont,anchor=west,text=ink2c},
    lab/.style={font=\fontsize{6.6}{7.8}\selectfont},
    small/.style={font=\fontsize{6.2}{7.2}\selectfont,text=ink2c},
    pill/.style={draw=aquac,fill=aquac!10,rounded corners=5.5pt,line width=.6pt,minimum height=12.5pt,
                 inner xsep=5pt,font=\fontsize{7.4}{8.5}\selectfont}]
  \path[use as bounding box] (0,0) rectangle (396,166);

  \filldraw[panel,fill=orangec!5,draw=orangec!22] (0.25,44) rectangle (191.75,165.75);
  \filldraw[panel,fill=aquac!6,draw=aquac!26] (204.25,44) rectangle (395.75,165.75);
  \node[head] at (8,156) {\textbf{Real law} $Q_u$};
  \node[sub] at (8,146.5) {choices depend on the scoring data};
  \node[head] at (212,156) {\textbf{Ghost law} $\bar Q_u$};
  \node[sub] at (212,146.5) {same transcript, independent scoring data};

  \foreach \ox/\selst/\stl/\scol/\sctext/\sname/\note/\arrlab in {
      0/rsel/{store $S\sim P_u^{\,n}$}/orangec/orangetext/{\widehat L_S(T)}/{reuses the data that chose $T$}/{same $S$},
      204/gsel/{fresh store $S^{\circ}\sim P_u^{\,n}$}/aquac/aquatext/{\widehat L_{S^{\circ}}(T)}/{$k$ fresh draws: error $\sigma/\sqrt{k}$}/{fresh $S^{\circ}$}}{
    \begin{scope}[xshift=\ox]
      \node[small,anchor=south west] at (12,131) {\stl};
      \foreach \r in {0,1,2}{\foreach \c in {0,...,8}{\node[cell] at ({16+9.6*\c},{124-9.6*\r}) {};}}
      \foreach \r/\c in {0/1,0/6,1/3,2/0,2/7}{\node[\selst] at ({16+9.6*\c},{124-9.6*\r}) {};}
      \node[draw=ink2c,fill=white,line width=.55pt,rounded corners=1.5pt,minimum width=24pt,minimum height=29pt] (T) at (150,71) {};
      \foreach \y/\w in {80/15,74.5/11,69/15,63.5/9}{\draw[ink2c!75,line width=.9pt,line cap=round] (143,\y) -- ++(\w,0);}
      \node[lab,anchor=west] at (164.5,71) {$T$};
      \node[score,draw=\scol,fill=\scol!9] (L) at (56,71) {$\sname$};
      \draw[arr,\scol] (56,101) -- node[lab,right=2.5pt,pos=.45,text=\sctext] {\arrlab} (L.north);
      \draw[arr,ink2c] (T.west) -- node[small,above=2pt] {prompt, positions} (L.east);
      \node[small,anchor=north,text=\sctext] at (56,61) {\note};
    \end{scope}
  }
  \node[box,draw=inkc,fill=white] (K) at (150,115)
       {\fontsize{7.6}{9}\selectfont pipeline $K_u$\\[1pt]\fontsize{6.1}{7}\selectfont retrieve, rerank, edit};
  \draw[arr,inkc] (102,115) -- (K.west);
  \draw[arr,inkc] (K.south) -- (150,86);
  \node[box,draw=ink2c,dashed,fill=white] (Kb) at (354,115)
       {\fontsize{7.6}{9}\selectfont marginal $\bar K_u$\\[1pt]\fontsize{6.1}{7}\selectfont never reads $S^{\circ}$};
  \draw[arr,ink2c] (Kb.south) -- (354,86);

  \draw[-{Stealth[length=5pt,width=4.4pt]},line width=1.05pt,inkc] (258,43) .. controls (246,29) and (72,29) .. (62,43);
  \node[font=\fontsize{6.8}{8}\selectfont,text=ink2c] at (159,27.2)
       {selection capacity $\kappa$ prices dependence on the scoring store};
  \node[font=\fontsize{8.6}{10}\selectfont,anchor=west] at (4,9.5)
       {$L_u(T)-\widehat L_S(T)\;\lesssim\;\sqrt{2\sigma^2\kappa/k}$};
  \node[small,anchor=east] at (190,10.5) {with $\kappa=$};
  \node[pill] (p1) at (225,10.5) {$\MI(T;S\cond U)$};
  \node[pill] (p2) at (292,10.5) {$\Leak(S\to T\cond U)$};
  \node[pill] (p3) at (360,10.5) {$\KL(Q\,\|\,\Pi)$};
  \node[small,anchor=north,yshift=-1.2pt] at (p1.south) {expected};
  \node[small,anchor=north,yshift=-1.2pt] at (p2.south) {realized};
  \node[small,anchor=north,yshift=-1.2pt] at (p3.south) {posterior};
\end{tikzpicture}
  \caption{\textbf{Decoupling selection from scoring isolates the price of adaptivity.} Left (\emph{real law}): the
  pipeline reads the scoring store $S$ and emits a transcript $T$ fixing both the prompt and the scored positions,
  reusing $S$ to compute $\Lhat_S(T)$. Right (\emph{ghost law}): a transcript with the identical marginal distribution
  is evaluated on an independent store $S^\circ$, averaging $k$ fresh draws at scale $\sigma/\sqrt{k}$. Change of
  measure prices the dependence removed by this refresh through the selection capacities of \cref{thm:transfer}.}
  \label{fig:hero}
\end{figure}

\paragraph{Decoupling through transcripts.}
We represent an adaptive pipeline by a \emph{transcript} $T$ that records every data-dependent choice determining the
predictor and its selected score, such as the winning recipe together with the indices of the scored queries.
Certification rests on comparing two joint distributions over the scoring store $S$ and the transcript $T$
(\cref{fig:hero}). Under the \emph{real law}, $T$ is generated by reading $S$, coupling the predictor and its evaluation
indices to the losses being averaged. Under the \emph{ghost law}, a transcript drawn from the same marginal
distribution is paired with an independent copy $S^\circ$ of the store. Once $T$ is fixed, its scoring positions are
deterministic and the losses evaluated at $S^\circ$ are fresh draws, concentrating around deployment risk at scale
$\sigma/\sqrt{k}$. Transporting this concentration back to the real law exacts an information price, the
\emph{selection capacity} $\kappa$, measuring how strongly $T$ depends on $S$.

\paragraph{Main results.}
This decouple-and-transport principle yields three contributions:
\begin{enumerate}[leftmargin=1.7em,label=\arabic*.,itemsep=4pt]
\item \textbf{Sharp certificates for selected scores.} A single change-of-measure template (\cref{thm:transfer})
delivers guarantees for three targets: expected optimism across runs, high-probability risk of the realized transcript,
and PAC-Bayesian risk of a randomized selector. For an i.i.d.\ store $S$ of $n$ items, losses in $[0,1]$, public side
information $U$, and uniform weights on $k$ scored examples, deployment risk $L_U(T)$ and selected score $\Lhat_S(T)$
satisfy
\begin{equation}\label{eq:headline}
  \bigl|\E\bigl[L_U(T)-\Lhat_S(T)\bigr]\bigr|\;\le\;\sqrt{\frac{\MI(T;S\cond U)}{2k}} .
\end{equation}
Concentration scales with the $k$ scored items rather than the ambient pool size $n$; nonuniform weights or
within-block dependence replace $k$ by an effective sample size $\keff$. A matching winner's-curse lower bound proves
that the $\sqrt{\kappa/k}$ rate is unimprovable for every $n\ge k$ and that its leading constant is exact in the
moderate-capacity regime (\cref{thm:lower}). Without a capacity constraint, by contrast, deterministic cherry-picking can
drive the selected score to zero while deployment risk approaches $1-2k/n$, forcing a nearly full-range correction
(\cref{rem:cherry}).
\item \textbf{A compositional capacity calculus.} To make these bounds computable without inspecting model weights, we
develop an accounting calculus (\cref{prop:calculus}) that prices candidate menus, noisy score releases, randomized
selections, and multi-round prompt edits directly from execution logs, while charging zero for post-processing or fresh
evaluation splits. In retrieve-then-rerank pipelines, confining a selector to a data-dependent shortlist decomposes
into an in-shortlist divergence plus a prior-mass gate charge, showing why first-stage filtering cannot be treated as
free preprocessing.
\item \textbf{Certification-optimal randomized selection.} Certifying a greedy winner post hoc pays the full menu cost
$\log M$ even when several candidates perform comparably. \emph{Stability-constrained adaptive randomization} (SCAR)
instead samples from the Gibbs posterior that minimizes the PAC-Bayesian certificate relative to a prior fixed before
scoring (\cref{thm:scar}), achieving the minimax selection rate up to constants. A variance-adaptive extension,
second-order SCAR (\cref{thm:second}), replaces worst-case loss range with a predictable plug-in variance estimate,
penalizing high-variance ``brittle'' prompts that masquerade as winners on small validation splits.
\end{enumerate}

\needspace{4\baselineskip}
\paragraph{Empirical findings.}
Experiments on Banking77 intent classification with Qwen2.5-7B-Instruct and GPT-4.1-mini confirm the predicted
$\sqrt{\log M/k}$ scaling of selection optimism and yield $100\%$ empirical certificate coverage across $24{,}000$
selection trials (\cref{sec:experiments}). SCAR achieves the lowest certified risk among all compared selectors at every
tested $k$: from $k=64$ reused validation queries on Qwen2.5-7B-Instruct and $k=128$ on GPT-4.1-mini, its certified
deployment risk beats zero-shot holdout error in over $99\%$ of validation draws while matching greedy holdout accuracy
within $0.01$. Controlled synthetic studies independently verify the roles of $k$ versus $n$, multi-round composition,
shortlist gate charges, and variance penalization.

\section{Related work}\label{sec:related}

\paragraph{In-context learning and retrieval.}
Theoretical analyses of in-context learning largely ask what a frozen model computes when conditioned on a prescribed
context. Transformers trained on synthetic families learn linear, smooth, and algorithmic function classes in context
\citep{garg2022what}, with attention layers capable of implementing gradient descent, ridge regression, or exact Bayesian
updates \citep{akyurek2022what,vonoswald2023transformers,he2025demystified}, while high-dimensional limits yield sharp
scaling laws for task diversity and representation alignment \citep{lu2025asymptotic,letey2025alignment,bordelon2025scaling}.
Closer to retrieval practice, \citet{guo2025rag} characterize estimation risk for retrieval-augmented linear regression
under a fixed, non-adaptive retrieval kernel. Whereas this literature studies forward-pass inference given a context
distribution, we study the statistical cost of \emph{selecting} the context and its evaluation examples on shared data.

\paragraph{Adaptive data analysis and information-theoretic generalization.}
Maintaining valid statistical guarantees across adaptively chosen queries is the central problem of adaptive data
analysis \citep{dwork2015generalization,dwork2015reusable,bassily2016algorithmic}, developed through approximate
max-information \citep{dwork2015generalization,rogers2016maxinformation}, maximal leakage
\citep{issa2020operational,esposito2021generalization}, and variance-calibrated stability
\citep{feldman2018calibrating}. In parallel, mutual information bounds both the bias of adaptively selected estimators
\citep{russo2016controlling} and the generalization error of learning algorithms \citep{xu2017information}, with
cumulant-generating refinements beyond sub-Gaussian tails \citep{jiao2017dependence,bu2020tightening}, conditional
ghost-sample formulations \citep{steinke2020reasoning}, and hierarchical decompositions across tasks in meta-learning
\citep{jose2021information} akin to \cref{prop:calculus}(d). Unlike standard empirical risk bounds that average over an
entire training sample, adaptive prompt pipelines frequently choose \emph{which} subset or weighted combination of store
items forms the reported score, making the transcript-determined effective size $\keff$ rather than the store size $n$
the governing scale.

\paragraph{PAC-Bayes and empirical variance control.}
PAC-Bayesian theory certifies randomized predictors via their Kullback--Leibler divergence from a data-independent prior
\citep{mcallester1999pacbayesian,catoni2007pacbayesian}, extending to data-dependent hypothesis classes
\citep{dupuis2024uniform} and variance-sensitive bounds that replace worst-case range with empirical second moments
\citep{tolstikhin2013pacbayes,mhammedi2019pacbayes,chugg2023unified}. Our second-order selector builds on the
predictable plug-in supermartingales of \citet{howard2021time} and \citet{waudbysmith2024estimating}, embedding them
inside a transcript-level change of measure that simultaneously accounts for shortlist gates and nonuniform index
selection. Operationally, multi-round prompt engineering on a fixed validation split mirrors leaderboard reuse, for
which the Ladder mechanism \citep{blum2015ladder} restricts information leakage by releasing scores only upon
significant improvement.

\paragraph{Post-selection inference and the winner's curse.}
Selecting the most promising noisy estimate induces the classical winner's curse, long recognized in genome-wide
association studies \citep{zollner2007overcoming} and formalized for prompt search in \cref{thm:lower}. Two traditions
address post-selection validity: simultaneous inference pays a universal constant valid across all selection rules
\citep{berk2013valid}, whereas conditional selective inference conditions on the polyhedral or algebraic structure of
the selection event when tractable \citep{fithian2014optimal,lee2016exact}. Bootstrap reality checks test the best rule
in a searched family \citep{white2000reality}, and \citet{rogers2016maxinformation} apply approximate max-information
(\cref{rem:maxinfo}) to adjust post-selection $p$-values. Our capacity $\kappa$ plays the role of a compositional,
distribution-free selection surcharge, measured in nats and read directly off the execution log of an arbitrary prompt
pipeline.

\section{Transcripts and their two laws}\label{sec:setup}

The statistical problem is defined by whichever data are reused to choose and score a prompt. An adaptive pipeline may
benchmark candidate recipes on a shared validation split, or select and reweight examples inside a larger corpus and
report performance on those chosen items. Both regimes ask the same question: how far can the reported empirical score
depart from expected loss on a fresh deployment draw? We collect all data touched during selection and scoring into a
\emph{scoring store} $S$, while any resource fixed before $S$ is drawn (such as a static demonstration bank or task
specification) enters as public side information $U$.

\subsection{Scoring data and selection choices}
Let $U\in\calU$ denote the task descriptor and public side information fixed prior to selection; all measurable spaces
below are standard Borel, ensuring regular conditional laws. Conditional on $U=u$, the scoring store
$S=(Z_1,\dots,Z_n)$ is drawn from a \emph{store law} $\Pstore_u$, either a product measure $P_u^n$ or an
independent-block law with common item marginal $P_u$ (\cref{ass:store}), and deployment performance is measured on an
independent draw $Z_\star\sim P_u$. Depending on the pipeline, store items $Z_i$ may represent labeled validation
queries, retrieved passages, candidate demonstrations, or tool-execution traces. A possibly randomized selection
pipeline is a Markov kernel $K_u(\dd t\mid s)$ that inspects the store $S=s$ and emits a \emph{transcript} $T\in\calT$.

\needspace{6\baselineskip}
\begin{definition}[Transcript and selected score]\label[definition]{def:transcript}
A transcript $t$ is \emph{complete} if every data-dependent object used to construct or evaluate the predictor (prompt
text, retrieved items, candidate menus, released scores, random seeds, critiques, and edit histories) is a measurable
function of $(t,u)$. Each transcript $t$ specifies a loss function $\ell_t:\calZ\to\R$ (the frozen model prompted by
$t$), distinct \emph{selected indices} $i_1(t),\dots,i_m(t)\in[n]$, and nonnegative weights $w_j(t)$ summing to $1$.
The \emph{selected score}, deployment risk, and generalization gap are
\[
  \Lhat_S(t)\defeq\sum_{j=1}^m w_j(t)\,\ell_t\bigl(Z_{i_j(t)}\bigr),\qquad L_u(t)\defeq\E\,\ell_t(Z_\star),\qquad
  \gap_u(t,S)\defeq L_u(t)-\Lhat_S(t).
\]
Under uniform weights $w_j(t)=1/k$ over $m=k$ items, $\Lhat_S(t)$ is the empirical average over those $k$ selected
examples, and we call $k$ the \emph{scoring sample size}. Throughout, scores are oriented as losses so that smaller
values are preferred.
\end{definition}

\Cref{def:transcript} unifies two modes of reuse. In \emph{prompt selection on a validation split} (as in the 64-recipe
Banking77 pipeline of \cref{sec:experiments}), every candidate shares the same $k$ validation indices while selection
varies $\ell_t$; demonstrations retrieved from an external bank fixed in $U$ shape $\ell_t$ without entering $S$. In
\emph{adaptive subset or self-scored selection}, the pipeline also chooses which indices $i_j(t)$ or weights $w_j(t)$
within $S$ form the reported loss. Completeness simply demands that nothing read from $S$ bypass the transcript.

\subsection{Real and ghost laws}

Under the actual pipeline, $\Lhat_S(T)$ is hard to analyze directly because both the loss function $\ell_T$ and the
indices $i_j(T)$ depend on $S$. Pairing the marginal distribution of $T$ with a fresh store $S^\circ$ breaks this
feedback loop, reducing selection bias to a divergence between two joint laws.

\begin{definition}[The two laws and selection capacity]\label[definition]{def:laws}
Given $U=u$, the \emph{real law} of $(S,T)$ is $Q_u(\dd s,\dd t)=\Pstore_u(\dd s)\,K_u(\dd t\mid s)$, and the
\emph{ghost law} is the product measure $\Qbar_u=\Pstore_u\otimes\Kbar_u$, where
$\Kbar_u\defeq\int K_u(\cdot\mid s)\,\Pstore_u(\dd s)$ is the marginal law of $T$. The \emph{expected selection
capacity} is the conditional mutual information, and the \emph{realized-gap selection capacity} is the conditional
maximal leakage \citep{issa2020operational}:
\[
  \MI(T;S\cond U)=\E_U\,\KL\bigl(Q_U\,\|\,\Qbar_U\bigr),\qquad
  \Leak(S\to T\cond U=u)=\log\int_{\calT}\esssup_{s\sim\Pstore_u}k_u(t\mid s)\,\mu(\dd t),
\]
where $k_u(t\mid s)$ is a jointly measurable density of $K_u(\cdot\mid s)$ (for $\Pstore_u$-almost every $s$) with
respect to a $\sigma$-finite reference measure $\mu$ on $\calT$, and $\Leak(S\to T\cond U)\defeq\esssup_u\Leak(S\to T\cond U=u)$.
The integral is invariant to the choice of $\mu$ and density version, equals $+\infty$ when no dominating $\sigma$-finite
measure exists, and simplifies to $\log\sum_t\esssup_s K_u(t\mid s)$ when $\calT$ is countable.
\end{definition}

Mutual information governs average optimism across repeated draws of the store, whereas maximal leakage, an $L^\infty$
information measure over the full selection kernel, delivers high-probability tail control for the single prompt chosen
in a run. Both capacities vanish if and only if $T$ is independent of $S$, and satisfy
$\MI(T;S\cond U=u)\le\Leak(S\to T\cond U=u)$. At the opposite extreme, deterministic selection of $k$ indices out of $n$
can expend up to $\log\binom nk$ nats, and writing unquantized continuous store values into $T$ can drive both
capacities to infinity.

\subsection{Independent units and effective sample size}

\begin{assumption}[Store and losses]\label[assumption]{ass:store}
For every $u$, one of the following holds:
(a) \emph{i.i.d.\ store}: $\Pstore_u=P_u^n$, and for every fixed $t$ the centered loss $\ell_t(Z)-L_u(t)$ under
$Z\sim P_u$ is $\sigma^2$-sub-Gaussian.
(b) \emph{Block store}: the store is partitioned into independent public blocks $B_1,\dots,B_G$, each item has marginal
law $P_u$ with arbitrary dependence inside a block, and $\ell_t\in[0,1]$, for which we set $\sigma^2=1/4$.
Defining the total weight placed on block $g$ by $\pi_g(t)\defeq\sum_{j:\,i_j(t)\in B_g}w_j(t)$ (with singleton blocks
in case (a)), the \emph{effective scoring sample size} of $t$ is $\keff(t)\defeq1/\sum_g\pi_g(t)^2$, and for a support
set $\calT_0\subset\calT$ we write $\keff\defeq\inf_u\inf_{t\in\calT_0}\keff(t)$.
\end{assumption}

The effective sample size $\keff(t)$ simultaneously captures nonuniform weighting and intra-block correlation. Uniform
weights across $k$ items drawn from distinct blocks yield $\keff(t)=k$; concentrating weight on multiple chunks of the
same document accumulates mass inside a single block $B_g$ and reduces $\keff(t)$ accordingly.

\paragraph{Mapping to deployed pipelines.}
Each primitive in \cref{def:transcript,ass:store} corresponds directly to standard deployment practice. Treating $\ell_t$
as a fixed function of $t$ reflects inference with a version-pinned API or frozen model weights, with any LLM-generated
critiques or prompt rewrites logged inside $T$. Classification error, exact-match failure, and bounded rubric losses lie
in $[0,1]$, immediately satisfying sub-Gaussianity with $\sigma^2=1/4$ via Hoeffding's lemma. Grouping passages by
source document or conversation thread enforces cross-block independence under \cref{ass:store}(b), while the audit
checklist in \cref{app:checklist} translates transcript completeness into concrete logging requirements.

\section{The transfer principle}\label{sec:transfer}

Certification proceeds in two steps: establish sub-Gaussian concentration for a fixed transcript on a fresh store, and
transport that concentration across the change of measure from the ghost law back to the real law. Varying the
divergence in the second step yields three guarantees, for average optimism, for the single deployed prompt, and for a
randomized selector.

\subsection{Concentration for a fixed transcript}

\begin{restatable}[Ghost concentration]{lemma}{ghostlemma}\label[lemma]{lem:ghost}
Under \cref{ass:store}, for every $u$ and every fixed transcript $t$, the gap $\gap_u(t,S^\circ)$ evaluated on an
independent store $S^\circ\sim\Pstore_u$ is centered and $\sigma^2/\keff(t)$-sub-Gaussian.
\end{restatable}

With $t$ held fixed, the selected indices $i_j(t)$ and weights $w_j(t)$ are deterministic constants. Under a fresh store
$S^\circ$, the scored losses are drawn marginally from $P_u$ and are independent across blocks (contributing block sums
in $[0,\pi_g(t)]$ under \cref{ass:store}(b)), so their weighted average concentrates at scale $\sigma/\sqrt{\keff(t)}$
(\cref{app:ghost}). When $k$ items carry uniform weight $1/k$, the remaining $n-k$ store items never enter
$\Lhat_{S^\circ}(t)$: ghost concentration is governed by the scored sample size, not by the ambient pool size $n$.

\needspace{6\baselineskip}
\subsection{Certificates after selection}

\begin{restatable}[Selected-score transfer]{theorem}{transferthm}\label[theorem]{thm:transfer}
Let \cref{ass:store} hold, and let $\keff$ be taken over a set $\calT_0$ of transcripts that contains every output
of the pipeline: $K_u(\calT_0\mid s)=1$ for every $s$ and $P_U$-almost every $u$.
\begin{enumerate}[label=(\roman*),leftmargin=2em]
\item \itemhead{Expected gap.} $\bigl|\E\,\gap_U(T,S)\bigr|\le\sqrt{2\sigma^2\,\MI(T;S\cond U)/\keff}$.
\item \itemhead{Realized gap.} If $\Leak(S\to T\cond U=u)\le\kappa$ for $P_U$-almost every $u$, then for every
$\delta\in(0,1)$, with probability at least $1-\delta$,
\(
  \gap_U(T,S)\le\sqrt{2\sigma^2\bigl(\kappa+\log(1/\delta)\bigr)/\keff};
\)
the same holds for $-\gap_U(T,S)$.
\item \itemhead{Posterior risk.} Let $\Pi_u$ be a law on $\calT$ fixed before $S$ is drawn, let $\keff$ be taken over
a set of $\Pi_u$-probability one, and fix $\lambda>0$ and $\delta\in(0,1)$. With probability at least $1-\delta$ over
$S\sim\Pstore_u$,
simultaneously for every law $Q\ll\Pi_u$ on $\calT$, including laws that depend on $S$,
\[
  \E_{t\sim Q}L_u(t)\;\le\;\E_{t\sim Q}\Lhat_S(t)+\frac{\KL(Q\,\|\,\Pi_u)+\log(1/\delta)}{\lambda}
  +\frac{\lambda\sigma^2}{2\keff}.
\]
\end{enumerate}
\end{restatable}

\paragraph{Three guarantees from one decoupling.}
Each clause of \cref{thm:transfer} matches a distinct operational target to its natural information price
(\cref{tab:readings}). Clause (i) bounds expected selection bias across repeated draws of the store via mutual
information. Clause (ii) certifies the specific prompt $T$ deployed from a single run at confidence $1-\delta$, paying
maximal leakage across the pipeline. Clause (iii) certifies the expected risk of a randomized predictor $Q$ (for
instance, drawing a prompt recipe afresh per deployment query) simultaneously over all posteriors $Q\ll\Pi_u$, allowing
$Q$ itself to be chosen after inspecting $S$ at a pathwise price $\KL(Q\|\Pi_u)$ relative to a prior $\Pi_u$ fixed
beforehand. As shown in \cref{app:divergences}, all three prices arise as specializations of a single R\'enyi divergence
$D_\alpha(Q_u\,\|\,\Pstore_u\otimes R)$, interpolated across orders $\alpha\in(1,\infty)$ by Sibson's information
(\cref{prop:sibson}).

\begin{table}[t]
  \centering
  \caption{\textbf{Matching the certificate to the reported quantity.} The three clauses of \cref{thm:transfer} price
  adaptivity according to whether one certifies average bias, a single chosen prompt, or a randomized selector. A finite
  menu of $M$ candidates costs at most $\log M$ in (i)--(ii), whereas (iii) charges the realized divergence from a fixed
  prior.}
  \label{tab:readings}
  \small
  \setlength{\tabcolsep}{4pt}
  \begin{tabular}{L{0.18\linewidth}L{0.30\linewidth}L{0.21\linewidth}L{0.20\linewidth}}
    \toprule
    \headrow\textbf{guarantee} & \textbf{reported target} & \textbf{selection cost} & \textbf{recipe example}\\
    \midrule
    (i) expected gap & average optimism across stores and selections & mutual information $\MI(T;S\cond U)$ & bounded by $\log M$\\
    (ii) realized gap & risk of the selected prompt, with confidence $1-\delta$ & maximal leakage $\Leak(S\to T\cond U)$ & bounded by $\log M$\\
    (iii) posterior risk & risk averaged over a randomized choice, with confidence $1-\delta$ & logged $\KL(Q\|\Pi_u)$ & SCAR's recipe distribution\\
    \bottomrule
  \end{tabular}
\end{table}

\paragraph{Scaling and numerical calibration.}
Selection capacity acts directly as an inverse sample-size multiplier: doubling $\kappa$ inflates the surcharge by the
same factor as halving $\keff$. On a menu of $M=64$ recipes scored on $k=64$ binary-loss examples ($\sigma^2=1/4$),
clause (i) bounds expected optimism by $\sqrt{\log 64/128}=0.180$, while clause (ii), like clause (iii) at a point mass
on the empirical winner under a uniform prior and optimal $\lambda$, gives the $95\%$ high-probability surcharge
$\sqrt{(\log64+\log20)/128}=0.236$, recovering the classical union-bound constant. Spreading posterior mass across
competitive recipes reduces $\KL(Q\|\Pi_u)$ well below $\log M$, a mechanism exploited systematically in
\cref{sec:scar}.

\begin{proof}[Proof idea]
Every clause pairs \cref{lem:ghost} with a single change of measure (\cref{app:aux}; full proofs in
\cref{app:transfer}). In (i), applying the Donsker--Varadhan variational inequality between $Q_u$ and $\Qbar_u$ yields
$\lambda\E_{Q_u}\gap\le\KL(Q_u\|\Qbar_u)+\lambda^2\sigma^2/(2\keff)$; optimizing over $\lambda$ and applying Jensen's
inequality over $U$ gives the bound. In (ii), each fixed transcript satisfies
$\Pstore_u(\gap_u(t,S^\circ)>\epsilon)\le e^{-\keff\epsilon^2/(2\sigma^2)}$, and maximal leakage bounds the real-law
probability of $\{\gap_u(T,S)>\epsilon\}$ by $e^\kappa$ times this worst-case section probability. In (iii), Fubini's
theorem and Markov's inequality ensure $\E_{\Pi_u}\exp\{\lambda\gap_u(t,S)-\lambda^2\sigma^2/(2\keff)\}\le1/\delta$ with
probability at least $1-\delta$ over $S$, after which Donsker--Varadhan between $Q$ and $\Pi_u$ transfers the bound
simultaneously to every posterior $Q$.
\end{proof}

\subsection{Why the rate cannot be improved}

\begin{restatable}[The price is sharp]{theorem}{lowerthm}\label[theorem]{thm:lower}
\begin{enumerate}[label=(\roman*),leftmargin=2em]
\item There is a universal constant $c>0$ such that for every $k\ge16$, every $1\le\kappa\le k/16$, and every store
size $n\ge k$, there exist a distribution $P$, a finite library of prompts with losses in $\{0,1\}$, and a pipeline
that selects $k$ distinct store items with $\MI(T;S)\le\Leak(S\to T)\le\kappa$ and
$\E\bigl[L(T)-\Lhat_S(T)\bigr]\ge c\sqrt{\kappa/k}$.
\item If $k\to\infty$ with $\log k=o(\kappa)$ and $\kappa=o(k)$, the same construction has
$\E\bigl[L(T)-\Lhat_S(T)\bigr]\ge(1-o(1))\sqrt{2\sigma^2\kappa/k}$ with $\sigma^2=1/4$, and its gap exceeds
$(1-o(1))\sqrt{2\sigma^2\kappa/k}$ with probability tending to one: the constants of \cref{thm:transfer}(i), and of
(ii) at every fixed $\delta$, are exact.
\end{enumerate}
\end{restatable}

The witness pipeline is elementary: evaluate $N=\lfloor e^\kappa\rfloor$ prompts of identical risk $1/2$ on the first
$k$ items of the store, deploy the empirical winner, and ignore the remaining $n-k$ items. Because coordinates are
independent fair signs, the winner's gap equals half the maximum of $N$ independent Rademacher averages; a reverse
Chernoff bound yields the non-asymptotic rate in (i), and a sharp method-of-types binomial estimate recovers the exact
constant $\sqrt{2\sigma^2}$ in (ii) (\cref{app:lower}). Pure noise-chasing among equally capable prompts thus saturates
\cref{thm:transfer}, while the $n-k$ untouched items demonstrate that enlarging the ambient store without scoring more
examples cannot tighten any uniform certificate.

\section{Budgeting capacity from logs}\label{sec:calculus}

An information-theoretic bound becomes an engineering tool once its price can be read directly off a pipeline's
execution log. Modern prompting workflows compose discrete menu choices, candidate filtering, noisy or quantized score
checks, and iterative edits. The rules below decompose any such multi-stage transcript into an additive capacity ledger
without inspecting internal model or retriever weights.

\needspace{6\baselineskip}
\begin{proposition}[Capacity calculus, informal]\label[proposition]{prop:calculus}
Let the pipeline produce $T=(T_1,\dots,T_R)$ in $R$ adaptive rounds given $U$.
\begin{enumerate}[label=(\alph*),leftmargin=2em]
\item \itemhead{Rounds add.} Total selection capacity is at most the sum of the conditional capacities across rounds
(with equality for mutual information via the chain rule).
\item \itemhead{Post-processing is free.} Any transformation of $(T,U)$ that does not consult $S$ (formatting,
truncation, or deduplication) adds zero capacity.
\item \itemhead{Base charges.} Choosing from a fixed menu of $M$ candidates costs at most $\log M$ in both expected and
realized-gap capacity. In mutual information, releasing a score vector $a(S)+\mathcal N(0,\rho^{-1}I)$ of global
$\ell_2$-sensitivity $\Delta\defeq\sup_{s,s'}\|a(s)-a(s')\|_2$ costs at most $\rho\Delta^2/2$, and sampling from a
posterior $Q_S$ costs at most $\E\,\KL(Q_S\|\Pi)$ relative to a prior $\Pi$ fixed in advance. Restricting $Q_S$ to a
data-dependent shortlist (or \emph{gate}) $A(S)$ splits this divergence into
$\KL(Q_S\|\Pi_{A(S)})+\log(1/\Pi(A(S)))$, where $\Pi_A$ denotes $\Pi$ conditioned on $A$. Evaluating on an independent
split read by no round costs $0$.
\item \itemhead{Side information splits the price.} When the store is i.i.d.\ within a latent task $W$ that the pipeline
observes only through $S$, mutual information decomposes as $\MI(T;S\cond U)=\MI(T;W\cond U)+\MI(T;S\cond U,W)$, and
\cref{thm:transfer} holds conditionally on $(U,W)$ at the within-task capacity $\MI(T;S\cond U,W)$.
\end{enumerate}
\end{proposition}

\begin{figure}[!ht]
  \centering
  \begin{tikzpicture}[x=1pt,y=1pt,
    every node/.style={inner sep=0pt,outer sep=0pt,text=inkc},
    stage/.style={fill=panelc,draw=rulec,rounded corners=4pt,line width=.45pt,minimum width=62pt,minimum height=31pt},
    stitle/.style={anchor=base,font=\fontsize{7.2}{8.4}\selectfont\bfseries},
    sop/.style={anchor=base,font=\fontsize{6.4}{7.6}\selectfont,text=ink2c},
    chip/.style={draw=aquac,fill=aquac!10,rounded corners=6.5pt,line width=.6pt,minimum width=51pt,
                 minimum height=15pt,text height=6.8pt,text depth=3.4pt,inner ysep=2.4pt,font=\fontsize{7.6}{9}\selectfont},
    sumchip/.style={fill=inkc,rounded corners=6.5pt,minimum width=20pt,minimum height=15pt,
                    text height=6.8pt,text depth=3.4pt,inner ysep=2.4pt,font=\fontsize{8.4}{9}\selectfont,text=white},
    op/.style={anchor=base,font=\fontsize{8.6}{9}\selectfont},
    tag/.style={rounded corners=5pt,line width=.5pt,minimum height=12pt,inner xsep=5pt,font=\fontsize{6.4}{7.6}\selectfont},
    arr/.style={-{Stealth[length=3.3pt,width=2.8pt]},line width=.6pt,ink2c},
    lab/.style={anchor=base,font=\fontsize{6.3}{7.5}\selectfont,text=mutec}]
  \path[use as bounding box] (0,0) rectangle (396,100);
  \node[font=\fontsize{7.8}{9}\selectfont] (u) at (11,80) {$u,S$};
  \foreach \i/\name/\op in {1/{menu}/{one of $M$ templates},
                            2/{gate}/{shortlist $A(S)$},
                            3/{Gibbs choice}/{$Q\propto\Pi_{A}\,e^{-\lambda\widehat L_S}$},
                            4/{noisy scores}/{release $a(S)+\xi$},
                            5/{edit rounds}/{$R$ chosen edits}}{
    \node[stage] (s\i) at ({62+66*(\i-1)},80) {};
    \node[stitle] at ({62+66*(\i-1)},84) {\name};
    \node[sop] at ({62+66*(\i-1)},71.5) {\op};
  }
  \node[font=\fontsize{7.8}{9}\selectfont] (T) at (386,80) {$T$};
  \draw[arr] (u) -- (s1);
  \foreach \i [evaluate=\i as \j using int(\i+1)] in {1,...,4}{\draw[arr] (s\i) -- (s\j);}
  \draw[arr] (s5) -- (T);
  \foreach \i/\c in {1/{$\log M$},2/{$\log\frac{1}{\Pi(A)}$},3/{$\mathrm{KL}(Q\,\|\,\Pi_{A})$},
                     4/{$\frac{\rho}{2}\,\Delta^{2}$},5/{$\sum_{r\le R}C_r$}}{
    \node[chip] (c\i) at ({62+66*(\i-1)},46) {\c};
    \draw[aquac!70,line width=.5pt] (s\i.south) -- (c\i.north);
  }
  \foreach \i in {1,...,4}{\node[op,text=aquatext] at ({95+66*(\i-1)},44.3) {$+$};}
  \node[op] at (364.5,44.3) {$=$};
  \node[sumchip] at (382,46) {$\kappa$};
  \foreach \i/\r in {1/{finite menu},2/{gate charge},3/{logged at run time},4/{Gaussian channel},5/{chain rule}}{
    \node[lab] at ({62+66*(\i-1)},29) {\r};
  }
  \node[tag,fill=panelc,draw=rulec,text=ink2c,anchor=west] (f1) at (2,9) {post-processing $+\,0$};
  \node[tag,fill=panelc,draw=rulec,text=ink2c,anchor=west] (f2) at ([xshift=5pt]f1.east) {fresh split $+\,0$};
  \node[tag,fill=orangec!7,draw=orangec!60,text=orangetext,anchor=west] at ([xshift=5pt]f2.east)
       {unrestricted selection: potentially vacuous (Remark~\ref{rem:cherry})};
\end{tikzpicture}
  \caption{\textbf{Stage-by-stage capacity accounting from runtime logs.} Each logged operation contributes a closed-form
  entry to the capacity ledger (\cref{prop:calculus}); rounds compose additively while post-processing and fresh splits
  cost zero. The ledger total $\kappa$ bounds $\MI(T;S\cond U)$ in \cref{thm:transfer}(i), and for finite menus and edit
  rounds also bounds maximal leakage in (ii). Combining a shortlist gate with Gibbs sampling forms a single variational
  round whose exact price $\KL(Q\|\Pi)$ is logged by SCAR (\cref{thm:scar}).}
  \label{fig:ledger}
\end{figure}
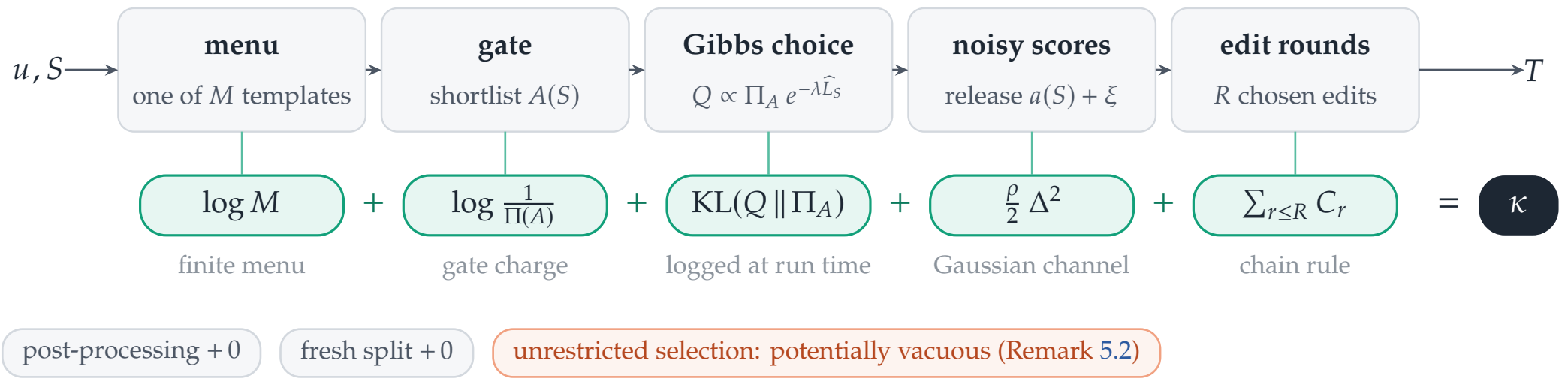

\Cref{prop:calculus} (stated and proved formally in \cref{app:proofs-calculus}) turns opaque pipeline design into
modular accounting (\cref{fig:ledger}). Discrete menus and multi-round edits control both mutual information and maximal
leakage, whereas Gaussian score releases and Gibbs posteriors control expected and PAC-Bayesian risk. In particular, the
gate decomposition exposes a pervasive pitfall in two-stage retrieve-then-rerank systems: narrowing a large library to a
promising shortlist $A(S)$ on the scoring data is never free preprocessing, as even a uniform or point choice inside
$A(S)$ must pay the surprise $\log(1/\Pi(A(S)))$ of landing in that shortlist.

\needspace{6\baselineskip}
\paragraph{A worked ledger.}
In our 64-recipe running example, choosing across four instruction templates, four retrieval rules, and four
demonstration layouts yields the ledger
\[
  \underbrace{\log4}_{\text{template}}+
  \underbrace{\log4}_{\text{retrieval rule}}+
  \underbrace{\log4}_{\text{configuration}}
  =\log64=4.16\ \text{nats},
\]
whether logged as one joint menu or as three sequential coordinate-descent rounds. At $k=64$ reused validation queries,
this $4.16$-nat budget yields an expected-gap bound of $0.180$ and a $95\%$ realized-gap correction of $0.236$. Replacing
hard selection by SCAR (\cref{sec:scar}) logs the realized posterior divergence instead, averaging $1.34$ nats on
Qwen2.5-7B-Instruct and $1.23$ nats on GPT-4.1-mini at $k=64$.

\paragraph{Why capacity must be budgeted.}
Are explicit capacity budgets genuinely necessary, or could a cleverer concentration inequality certify arbitrary
selection using $(n,k,\delta)$ alone? The following construction shows that without a restriction on selection capacity,
\emph{every} valid certificate becomes nearly vacuous: an unconstrained selector can cherry-pick $k$ store items on
which even a mediocre prompt suffers zero empirical loss, spending up to $\log\binom nk$ nats (or infinite capacity when
emitting continuous store contents verbatim). Non-vacuous evaluation therefore requires either fresh scoring data or an
explicit capacity budget.

\begin{restatable}[Cherry-picking]{remark}{cherryremark}\label[remark]{rem:cherry}
Fix one prompt whose loss is Bernoulli$(p)$ under $P$, and let the pipeline select $k$ store items on which that
loss is $0$. If $n(1-p)\ge2k$, then with probability at least $1-e^{-k/4}$ the selected score is $0$ while the risk
is $p$. A surcharge $b(n,k,\delta)$ with $\Pp(L\le\Lhat_S+b)\ge1-\delta$ for every $p\le1-2k/n$, at a level
$\delta<1-e^{-k/4}$, is therefore at least $1-2k/n$.
\end{restatable}

\section{Stability-constrained adaptive randomization}\label{sec:scar}

Committing deterministically to the single lowest-error candidate pays the full menu cost $\log M$, even when a dozen
prompt recipes achieve nearly indistinguishable validation performance. The PAC-Bayesian guarantee of
\cref{thm:transfer}(iii) suggests a more economical principle: instead of selecting a prompt heuristically and pricing
it afterward, choose the distribution over transcripts that \emph{directly minimizes the certified risk bound}.

Let $\Pi_u$ be a reference prior over candidate transcripts fixed before the scoring store $S$ is read (for instance,
uniform over our 64 prompt recipes), and let $A(S,u)\subset\calT$ be an optional data-dependent shortlist, or
\emph{gate}, produced by a first-stage filter, with conditioned prior $\Pi_{u,A}$. Given an inverse temperature
$\lambda>0$ fixed prior to scoring, \emph{stability-constrained adaptive randomization} (SCAR) samples transcripts from
the gated Gibbs posterior
\[
  Q_\lambda(\dd t\mid S,u)\;\propto\;\Pi_u(\dd t)\,e^{-\lambda\Lhat_S(t)}\,\mathbf 1\{t\in A(S,u)\}.
\]
By the Gibbs variational principle, $Q_\lambda$ is the exact minimizer of
$Q\mapsto\E_Q\Lhat_S+\lambda^{-1}\KL(Q\|\Pi_u)$ among all distributions supported on the gate $A(S,u)$. Fitting the
scoring data and conserving selection capacity thus become two halves of a single variational objective; SCAR logs the
resulting divergence and outputs the matching risk certificate (\cref{alg:scar}).

\needspace{6\baselineskip}
\subsection{A Gibbs selector with a logged information cost}

\begin{algorithm}[!htb]
  \caption{Stability-constrained adaptive randomization (SCAR)}
  \label{alg:scar}
  \begin{algorithmic}[1]
    \Require query $u$, store $S$, prior $\Pi_u$ fixed before any score is read, optional gate $A(S,u)$, inverse
      temperature $\lambda$ and level $\delta$ fixed in advance
    \State Score every transcript on the gate: $\Lhat_S(t)$ on its selected examples.
    \State Form $Q_\lambda(t)\propto\Pi_u(t)\,e^{-\lambda\Lhat_S(t)}$ on $A(S,u)$; sample $T\sim Q_\lambda$ afresh for each query.
    \State Log $\KL(Q_\lambda\|\Pi_u)=\KL(Q_\lambda\|\Pi_{u,A})+\log\bigl(1/\Pi_u(A)\bigr)$, the gate charge included.
    \State Report $\E_{Q_\lambda}\Lhat_S+\bigl(\KL(Q_\lambda\|\Pi_u)+\log\tfrac1\delta\bigr)/\lambda+\lambda\sigma^2/(2\keff)$ (\cref{thm:scar}).
  \end{algorithmic}
\end{algorithm}

\begin{restatable}[SCAR certificate and oracle inequality]{theorem}{scarthm}\label[theorem]{thm:scar}
Let \cref{ass:store} hold with $\keff$ taken over a set of $\Pi_u$-probability one, let $\Pi_u$ be fixed before $S$
is drawn, fix $\lambda>0$ and $\delta\in(0,1)$, and let $A=A(S,u)$ be any measurable data-dependent gate with
$\Pi_u(A)>0$ and $\int_{A}e^{-\lambda\Lhat_S}\dd\Pi_u<\infty$ (automatic for bounded losses). With probability at
least $1-\delta$,
\begin{align*}
  \E_{Q_\lambda}L_u&\le\E_{Q_\lambda}\Lhat_S+\frac{\KL(Q_\lambda\|\Pi_u)+\log(1/\delta)}{\lambda}
    +\frac{\lambda\sigma^2}{2\keff}\\
  &\le\inf_{Q:\ Q(A)=1}\Bigl\{\E_Q\Lhat_S+\frac{\KL(Q\|\Pi_{u,A})+\log\bigl(1/\Pi_u(A)\bigr)}{\lambda}\Bigr\}
    +\frac{\log(1/\delta)}{\lambda}+\frac{\lambda\sigma^2}{2\keff}.
\end{align*}
\end{restatable}

\paragraph{Variational optimality and two-stage gating.}
\Cref{thm:scar} pairs validity with an exact oracle inequality (\cref{app:scar}): on the high-probability event of
\cref{thm:transfer}(iii), SCAR achieves the smallest certified upper bound among \emph{all} posteriors supported on the
shortlist $A(S,u)$, including point masses on the empirical winner. The identity
$\KL(Q\|\Pi_u)=\KL(Q\|\Pi_{u,A})+\log(1/\Pi_u(A))$ separates the cost of reweighting within the shortlist from the
$\log(1/\Pi_u(A))$ price of restricting to it, rendering two-stage retrieve-then-rerank pipelines certifiable in a
single formula.

\paragraph{Temperature calibration and minimax optimality.}
The inverse temperature $\lambda$ governs how aggressively SCAR concentrates on low empirical loss: as $\lambda\to0$,
$Q_\lambda$ collapses to the conditioned prior $\Pi_{u,A}$; as $\lambda\to\infty$, it concentrates on empirical
minimizers inside $A$. Under a uniform prior over $M$ candidates and losses in $[0,1]$ ($\sigma^2=1/4$), balancing the
two surcharge terms against a point-mass comparator yields the a-priori schedule
\[
  \lambda^\star=\sqrt{8\keff(\log M+\log(1/\delta))},
\]
at which a point mass recovers the union-bound surcharge $\sqrt{(\log M+\log(1/\delta))/(2\keff)}$, while any spread in
$Q_{\lambda^\star}$ strictly improves the certificate. Moreover, at $\keff=k$ and $\lambda=\lambda^\star$, SCAR's
population risk lies within $\sqrt{2(\log M+\log(1/\delta))/k}$ of the best recipe in the library with probability at
least $1-2\delta$, matching the Fano minimax lower bound of $0.07\sqrt{\log M/k}$ for $4\le M\le e^{2k/3}$ up to
universal constants (\cref{app:scar-optimal}). Requiring certifiability thus incurs no penalty in the minimax selection
rate.

\subsection{Refining the certificate with loss variability}

First-order certificates treat every bounded loss through the worst-case sub-Gaussian proxy $\sigma^2=1/4$, making no
distinction between a prompt whose errors are consistently small and a ``brittle'' prompt whose erratic binary swings
happen to land favorably on a small validation sample. Our second-order selector replaces range with empirical variance
via a one-pass predictable plug-in estimator: along the transcript's scoring order, each loss is centered against a
running mean computed strictly from preceding examples, preserving a martingale structure under the ghost law while
allowing post-hoc optimization over a finite parameter grid $\Gamma$.

\needspace{6\baselineskip}
\begin{restatable}[Second-order SCAR]{theorem}{secondthm}\label[theorem]{thm:second}
Let \cref{ass:store}(a) hold with $\ell_t\in[0,1]$ and uniform weights over $k$ distinct selected examples. For a
fixed $t$ write $X_j(t)=\ell_t(Z_{i_j(t)})$ in transcript order, $\muhat_{j-1}(t)=\bigl(\tfrac12+\sum_{i<j}X_i(t)\bigr)/j$,
and $\Vhat_S(t)=\frac1k\sum_{j=1}^k\bigl(X_j(t)-\muhat_{j-1}(t)\bigr)^2$. Let $\Gamma\subset(0,1)$ be a finite grid
fixed in advance and $\psi(\gamma)=-\log(1-\gamma)-\gamma$. With probability at least $1-\delta$, simultaneously for
every $Q\ll\Pi_u$ and every $\gamma\in\Gamma$,
\[
  \E_QL_u\;\le\;\E_Q\Lhat_S+\frac{\psi(\gamma)}{\gamma}\,\E_Q\Vhat_S+\frac{\KL(Q\|\Pi_u)+\log(|\Gamma|/\delta)}{\gamma k}.
\]
For fixed $\gamma$, the second-order posterior $Q\propto\Pi_u\exp\{-k(\gamma\Lhat_S+\psi(\gamma)\Vhat_S)\}$ minimizes
the right-hand side.
\end{restatable}

Because $\psi(\gamma)/\gamma=\gamma/2+O(\gamma^2)$, optimizing over $\gamma$ balances the two correction terms near
$\gamma\approx\sqrt{2(\KL+\log(|\Gamma|/\delta))/(k\E_Q\Vhat_S)}$ and yields a leading Bernstein-type surcharge of
\[
  \sqrt{2\,\E_Q\Vhat_S\,(\KL+\log(|\Gamma|/\delta))/k},
\]
replacing the worst-case constant $1/4$ with the posterior's predictable loss variance $\E_Q\Vhat_S$ (\cref{app:second}).
A dyadic grid $\Gamma=\{0.9\cdot2^{-i}:0\le i\le\lceil\log_2k\rceil\}$ pays only $\log|\Gamma|=O(\log\log k)$ nats for
simultaneous tuning, and penalizing $\Vhat_S$ directly in the Gibbs exponent suppresses high-variance prompts that
deceive first-order empirical selection (\cref{fig:synthetic}d).

\section{Experiments}\label{sec:experiments}

We test three empirical predictions of the theory: whether selection optimism scales with $\sqrt{\log M/k}$, whether
variational randomization via SCAR tightens certified deployment risk without sacrificing accuracy, and how scoring
sample size, multi-round composition, shortlist gates, and loss variance individually shape post-selection validity.
Complete protocols, compute details, and supplementary ablations appear in \cref{app:experiments}.

\subsection{Recipe selection on Banking77}
We study 77-way online-banking intent classification on Banking77 \citep{casanueva2020efficient} across two frozen
instruction-tuned models run at temperature $0$: Qwen2.5-7B-Instruct \citep{qwen2025qwen25} served via vLLM
\citep{kwon2023efficient}, and GPT-4.1-mini. Prior to validation, 2{,}000 labeled training queries are embedded with
bge-small-en-v1.5 \citep{xiao2024cpack} to form a fixed public demonstration bank ($U$). Over this bank we construct a
library of 64 prompt recipes, combining 4 instruction templates, 4 retrieval rules (random, top-$m$, label-diverse
top-$m$, and maximal marginal relevance), and 4 demonstration layouts ($m\in\{4,8\}$, ordered most-similar first or
last).

Selection and risk evaluation operate on two disjoint test splits: a validation pool of 256 queries and an independent
holdout of 1{,}024 queries (512 for GPT-4.1-mini; holdout standard errors in \cref{app:banking}). Across a $6\times5$
grid of menu sizes $M\in\{2,4,8,16,32,64\}$ and scoring sample sizes $k\in\{16,32,64,128,256\}$, we draw 400 random
validation subsets and recipe sub-libraries per cell (reusing the full validation pool when $k=256$). Greedy selection
deploys the lowest-validation-error recipe and pays the menu charge $\log M$, whereas SCAR forms its Gibbs posterior
over all 64 recipes and pays the realized divergence $\KL(Q_{\lambda^\star}\|\Pi)$. All comparisons run over cached
outputs from 139{,}264 model evaluations.

\subsection{Selection bias and certificate size}
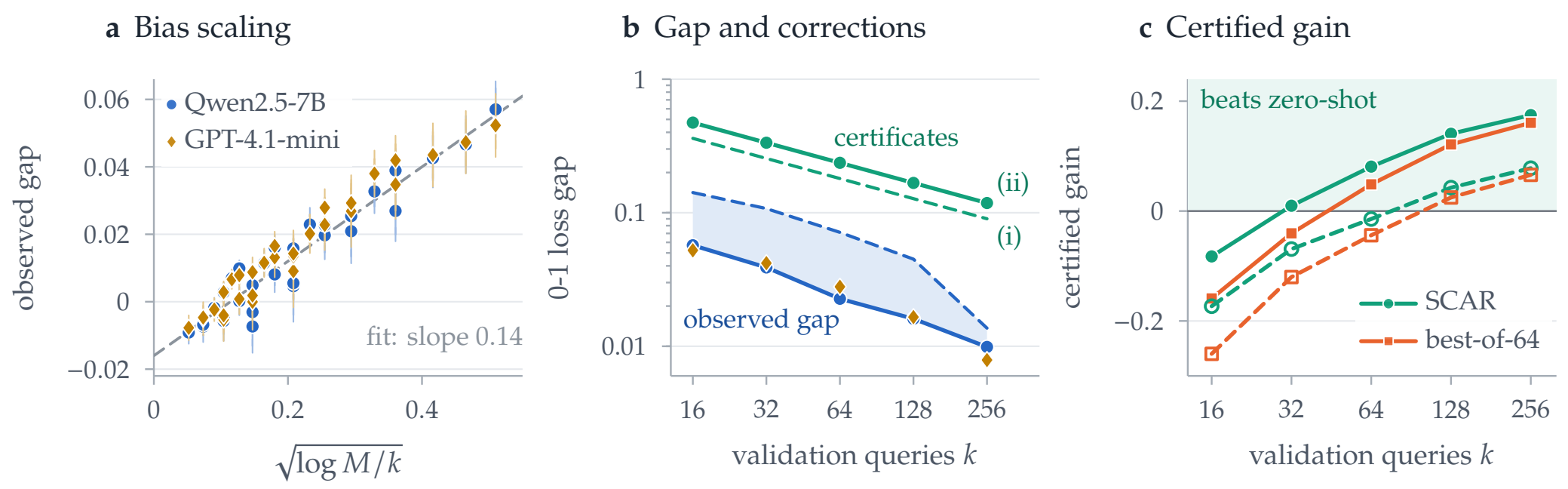
\begin{figure}[!ht]
  \centering
  \begin{tikzpicture}
\begin{groupplot}[group style={group size=3 by 1, horizontal sep=37pt},
    paper, width=92pt, height=74pt]
  \nextgroupplot[title={\paneltitle{a}{Bias scaling}},
    xlabel={$\sqrt{\log M/k}$}, ylabel={observed gap},
    xmin=0, xmax=0.55, ymin=-0.022, ymax=0.066, ytick={-0.02,0,0.02,0.04,0.06}, xtick={0,0.2,0.4},
    scaled ticks=false, yticklabel style={/pgf/number format/fixed,/pgf/number format/precision=2},
    xticklabel style={/pgf/number format/fixed},
    legend pos=north west, legend style={at={(0.02,1.0)}}]
  \addplot[mutec, line width=.7pt, densely dashed, domain=0:0.55, samples=2, forget plot] {0.140*x-0.016};
  \addplot[bluec, only marks, dot, error bars/.cd, y dir=both, y explicit,
    error bar style={bluec!45, line width=.45pt}, error mark options={draw=none}]
    table[col sep=comma, x=scale, y=mean_gap, y error=ci]{data/rp_gap_qwen.csv};
  \addlegendentry{Qwen2.5-7B}
  \addplot[amberc, only marks, diam, error bars/.cd, y dir=both, y explicit,
    error bar style={amberc!45, line width=.45pt}, error mark options={draw=none}]
    table[col sep=comma, x=scale, y=mean_gap, y error=ci]{data/rp_gap_gpt.csv};
  \addlegendentry{GPT-4.1-mini}
  \node[dlabel mute, anchor=west] at (axis cs:0.31,-0.011) {fit: slope $0.14$};
  \nextgroupplot[title={\paneltitle{b}{Gap and corrections}},
    xlabel={validation queries $k$}, ylabel={0-1 loss gap},
    xmode=log, log basis x=2, xtick={16,32,64,128,256}, xticklabels={16,32,64,128,256},
    xmin=13, xmax=420, ymode=log, ymin=0.006, ymax=1, ytick={0.01,0.1,1}, yticklabels={0.01,0.1,1}]
  \addplot[draw=none, name path=p95, forget plot] table[col sep=comma, x=k, y=p95]{data/rp_ksweep_qwen.csv};
  \addplot[draw=none, name path=mn, forget plot] table[col sep=comma, x=k, y=mean_gap]{data/rp_ksweep_qwen.csv};
  \addplot[bluec!14, forget plot] fill between[of=p95 and mn];
  \addplot[bluec, line width=.8pt, densely dashed] table[col sep=comma, x=k, y=p95]{data/rp_ksweep_qwen.csv};
  \addplot[bluec, data, dot] table[col sep=comma, x=k, y=mean_gap]{data/rp_ksweep_qwen.csv};
  \addplot[amberc, only marks, diam] table[col sep=comma, x=k, y=mean_gap]{data/rp_ksweep_gpt.csv};
  \node[dlabel blue, anchor=north west] at (axis cs:14,0.021) {observed gap};
  \addplot[aquac, line width=.8pt, densely dashed] table[col sep=comma, x=k, y=cert_expected]{data/rp_ksweep_qwen.csv};
  \addplot[aquac, data, dot] table[col sep=comma, x=k, y=cert_hp]{data/rp_ksweep_qwen.csv};
  \node[dlabel aqua, anchor=south west] at (axis cs:58,0.29) {certificates};
  \node[dlabel aqua, anchor=south west] at (axis cs:268,0.121) {(ii)};
  \node[dlabel aqua, anchor=north west] at (axis cs:268,0.088) {(i)};
  \nextgroupplot[title={\paneltitle{c}{Certified gain}},
    xlabel={validation queries $k$}, ylabel={certified gain},
    xmode=log, log basis x=2, xtick={16,32,64,128,256}, xticklabels={16,32,64,128,256},
    xmin=13, xmax=320, ymin=-0.3, ymax=0.24, ytick={-0.2,0,0.2},
    yticklabel style={/pgf/number format/fixed,/pgf/number format/precision=1},
    legend style={at={(0.99,0.03)}, anchor=south east}]
  \fill[aquac!9] (axis cs:13,0) rectangle (axis cs:320,0.24);
  \addplot[inkc!60, line width=.5pt, domain=13:320, samples=2, forget plot] {0};
  \node[dlabel aqua, anchor=north west] at (axis cs:14,0.232) {beats zero-shot};
  \addplot[aquac, data, dot] table[col sep=comma, x=k, y expr=\thisrow{zero_shot}-\thisrow{scar_cert}]{data/rp_scar_qwen.csv};
  \addlegendentry{SCAR}
  \addplot[orangec, data, sq] table[col sep=comma, x=k, y expr=\thisrow{zero_shot}-\thisrow{greedy_cert}]{data/rp_scar_qwen.csv};
  \addlegendentry{best-of-64}
  \addplot[aquac, data, densely dashed, opendot, forget plot] table[col sep=comma, x=k, y expr=\thisrow{zero_shot}-\thisrow{scar_cert}]{data/rp_scar_gpt.csv};
  \addplot[orangec, data, densely dashed, mark=square, mark size=1.45pt, mark options={solid, fill=white, line width=.7pt}, forget plot]
    table[col sep=comma, x=k, y expr=\thisrow{zero_shot}-\thisrow{greedy_cert}]{data/rp_scar_gpt.csv};
\end{groupplot}
\end{tikzpicture}
  \caption{\textbf{Selection optimism and certified deployment gains on Banking77.}
  (a) Mean holdout-minus-validation gap across 30 $(M,k)$ cells with $95\%$ intervals over validation draws
  (blue circles: Qwen2.5-7B-Instruct; amber diamonds: GPT-4.1-mini; dashed gray: linear fit in $\sqrt{\log M/k}$).
  (b) At $M=64$, Qwen2.5-7B-Instruct mean gap (solid blue) and 95th percentile (dashed blue, shaded band), alongside
  GPT-4.1-mini mean gap (amber diamonds), the expected-gap bound (dashed aqua), and the $95\%$ realized-gap certificate
  (solid aqua) of \cref{thm:transfer}.
  (c) Certified gain over zero-shot holdout error (zero-shot error minus certified risk) for SCAR (aqua) versus
  best-of-64 selection (orange), solid for Qwen2.5-7B-Instruct and dashed for GPT-4.1-mini; positive values (shaded)
  certify a strict improvement over zero-shot deployment.}
  \label{fig:real}
\end{figure}

Across all 30 $(M,k)$ cells, average selection optimism (holdout error minus selected validation error) tracks the
predicted $\sqrt{\log M/k}$ law closely, with least-squares slopes of $0.140$ on Qwen2.5-7B-Instruct and $0.141$ on
GPT-4.1-mini (\cref{fig:real}a).

Across all $12{,}000$ selection events per model ($24{,}000$ in total), empirical coverage is $100\%$: no realized gap
exceeds its $95\%$ certificate, with the most extreme draw reaching $95\%$ of the bound (\cref{fig:real}b). At $M=k=64$,
mean observed gaps are $0.023$ and $0.028$ against the expected-gap bound of $0.180$, and 95th-percentile gaps are
$0.071$ and $0.074$ against the realized-gap surcharge of $0.236$. This three-to-eightfold margin for $k\le128$ reflects
the geometry of real prompt libraries: \cref{thm:lower} proves that the certificate is exact when candidates have
identical risk and independent errors, whereas practical recipes share the same validation queries, exhibit substantial
error correlation across templates, and span genuine differences in population risk (from $0.092$ to $0.366$ on Qwen),
naturally compressing realized optimism inside the distribution-free envelope.

\FloatBarrier
\subsection{SCAR improves the certified risk}
At every sample size $k$ and on both model families, SCAR delivers the lowest certified deployment risk among all
compared procedures (\cref{tab:scar}), improving upon best-of-64 certification by $0.01$ to $0.09$ absolute risk points.
This advantage follows directly from the variational oracle inequality (\cref{thm:scar}): evaluated at the same
$\lambda^\star$, committing to a point mass on the winner recovers the $\log64=4.16$-nat menu bound, whereas SCAR retains
probability mass across top-performing recipes and expends only $0.86$--$1.78$ nats while matching the greedy winner's
holdout error within $0.01$.

\begin{table}[!htb]
  \centering
  \caption{\textbf{SCAR achieves the tightest certified risk across sample sizes and model families.} Top: $95\%$
  certified deployment risk (lower is better), averaged over 50 validation draws, comparing SCAR (\cref{thm:scar})
  against best-of-64 selection certified via maximal leakage (\cref{thm:transfer}(ii)) or approximate max-information
  (\cref{rem:maxinfo}); shaded cells beat zero-shot holdout error. Bottom: selection capacity expended in nats, comparing
  SCAR's logged $\KL(Q_{\lambda^\star}\|\Pi)$ against the fixed $\log64$ menu charge.}
  \label{tab:scar}
  \small
  \setlength{\tabcolsep}{3.6pt}
  \begin{tabular}{l ccccc c ccccc}
    \toprule
    \headrow & \multicolumn{5}{c}{\textbf{Qwen2.5-7B} (zero-shot $0.373$)} && \multicolumn{5}{c}{\textbf{GPT-4.1-mini} (zero-shot $0.282$)}\\
    \hhline{>{\arrayrulecolor{headc}}->{\arrayrulecolor{black}}----->{\arrayrulecolor{headc}}->{\arrayrulecolor{black}}-----}
    \headrow $k$ & 16 & 32 & 64 & 128 & 256 && 16 & 32 & 64 & 128 & 256\\
    \midrule
    SCAR   & 0.455 & \beat 0.363 & \beat 0.292 & \beat 0.232 & \beat 0.198 && 0.455 & 0.351 & 0.296 & \beat 0.239 & \beat 0.204\\
    best-of-64, (ii) & 0.532 & 0.413 & \beat 0.324 & \beat 0.251 & \beat 0.212 && 0.542 & 0.402 & 0.326 & \beat 0.257 & \beat 0.216\\
    best-of-64, max-info. & 0.659 & 0.503 & 0.388 & \beat 0.296 & \beat 0.244 && 0.669 & 0.492 & 0.390 & 0.302 & \beat 0.248\\
    \midrule
    $\kappa$, SCAR & 1.08 & 1.23 & 1.34 & 1.59 & 1.63 && 0.86 & 0.99 & 1.23 & 1.47 & 1.78\\
    $\kappa$, best-of-64 & \multicolumn{5}{c}{$\log64=4.16$ at every $k$} && \multicolumn{5}{c}{$\log64=4.16$ at every $k$}\\
    \bottomrule
  \end{tabular}
\end{table}

Economizing on selection capacity translates directly into earlier statistical certification of few-shot gains
(\cref{fig:real}c, \cref{tab:rates}). Across 4{,}000 independent validation draws per $k$, SCAR certifies a deployment
risk strictly below zero-shot holdout error in over $99\%$ of draws starting at $k=64$ for Qwen2.5-7B-Instruct and
$k=128$ for GPT-4.1-mini; at those same sample sizes, best-of-64 achieves $96\%$ and $89\%$ under maximal leakage, and
only $39\%$ and $9\%$ under approximate max-information.

Sequential coordinate search over the same library confirms the additive ledger of \cref{sec:calculus}: selecting the
instruction template, retrieval rule, and demonstration layout across three rounds on $k=64$ shared queries produces
monotonically growing mean gaps of $(0.004, 0.005, 0.015)$ on Qwen2.5-7B-Instruct and $(0.011, 0.011, 0.019)$ on
GPT-4.1-mini, remaining well within the composed $95\%$ certificates $(0.185, 0.212, 0.236)$ accumulated at $\log4$ nats
per round (\cref{tab:library}).

\subsection{Mechanism checks}
\begin{figure}[!ht]
  \centering
  \begin{tikzpicture}
\begin{groupplot}[group style={group size=2 by 2, horizontal sep=48pt, vertical sep=49pt},
    paper, width=146pt, height=76pt]
  \nextgroupplot[title={\paneltitle{a}{The denominator is $k$}}, xlabel={$\sqrt{\log N/k}$}, ylabel={gap of the winner},
    xmin=0, xmax=0.34, ymin=0, ymax=0.25, ytick={0,0.1,0.2}, xtick={0,0.1,0.2,0.3},
    scaled ticks=false, tick label style={/pgf/number format/fixed}]
  \addplot[aquac, data, domain=0:0.34, samples=2] {0.7071*x};
  \node[dlabel aqua, anchor=north west, align=left] at (axis cs:0.012,0.245) {Thm.~\ref*{thm:transfer}(i)\\[-1.5pt]with $k$};
  \addplot[orangec, line width=.9pt, densely dashed, domain=0:0.34, samples=2] {0.2236*x};
  \node[dlabel orange, anchor=north east, align=right] at (axis cs:0.338,0.034) {with $n=10k$};
  \addplot[bluec, only marks, dot] table[col sep=comma, x=scale, y=mean_gap]{data/context_gap_scaling.csv};
  \nextgroupplot[title={\paneltitle{b}{Rounds add}}, xlabel={$\sqrt{R\log 4/k}$},
    xmin=0.1, xmax=0.4, ymin=0, ymax=0.25, ytick={0,0.1,0.2}, xtick={0.1,0.2,0.3,0.4},
    scaled ticks=false, tick label style={/pgf/number format/fixed}]
  \addplot[aquac, data, domain=0.1:0.4, samples=2] {0.7071*x};
  \node[dlabel aqua, anchor=north west] at (axis cs:0.108,0.245) {Thm.~\ref*{thm:transfer}(i)};
  \addplot[bluec, only marks, dot] table[col sep=comma, x=predicted_scale, y=mean_gap]{data/multiround_prompt_search.csv};
  \node[dlabel blue, anchor=south east] at (axis cs:0.396,0.012) {$R=1,\dots,6$};
  \nextgroupplot[title={\paneltitle{c}{Gates must be charged}}, xlabel={candidates $N$}, ylabel={coverage},
    xmode=log, log basis x=2, xmin=0.8, xmax=30000, xtick={1,16,256,4096}, xticklabels={1,16,256,4096},
    ymin=0, ymax=1.06, ytick={0,0.5,0.95}, yticklabels={0,0.5,0.95}]
  \addplot[mutec, line width=.6pt, dotted, domain=0.8:30000, samples=2, forget plot] {0.95};
  \addplot[aquac, data, dot] table[col sep=comma, x=N, y=cover_leakage]{data/syn_negative_control.csv};
  \node[dlabel aqua, anchor=north east, align=right] at (axis cs:26000,0.88) {gate\\[-1.5pt]charged};
  \addplot[orangec, data, sq] table[col sep=comma, x=N, y=cover_naive]{data/syn_negative_control.csv};
  \node[dlabel orange, anchor=south west, align=left] at (axis cs:1.1,0.03) {gate\\[-1.5pt]ignored};
  \nextgroupplot[title={\paneltitle{d}{Variance is priced}}, xlabel={scored examples $k$},
    ylabel={brittle-pick rate},
    xmode=log, log basis x=2, xmin=6.5, xmax=620, xtick={8,32,128,512}, xticklabels={8,32,128,512},
    ymin=0, ymax=1.06, ytick={0,0.5,1}]
  \addplot[orangec, data, sq] table[col sep=comma, x=k, y=first_brittle]{data/syn_second_order.csv};
  \node[dlabel orange, anchor=north east] at (axis cs:600,0.99) {first order};
  \addplot[aquac, data, dot] table[col sep=comma, x=k, y=second_brittle]{data/syn_second_order.csv};
  \node[dlabel aqua, anchor=north east] at (axis cs:600,0.80) {Thm.~\ref*{thm:second}};
\end{groupplot}
\end{tikzpicture}
  \caption{\textbf{Controlled benchmarks isolating the four theoretical mechanisms.}
  (a) Equal-risk Rademacher library at $k=64$: empirical winner's gap (blue), the expected-gap bound scaling with $1/\sqrt{k}$
  (aqua), and the invalid curve obtained by substituting store size $n=10k$ into the denominator (orange).
  (b) Multi-round search (four edits per round, $R=1,\dots,6$): mean gap versus $\sqrt{R\log4/k}$ alongside the composed
  ledger bound (aqua).
  (c) Empirical $95\%$ coverage for a data-dependent singleton winner gate when its prior-mass charge $\log N$ is
  included (aqua) versus omitted (orange; dotted line: nominal $0.95$).
  (d) Probability of selecting a high-variance brittle prompt under first-order (orange) versus second-order SCAR (aqua).
  Trials per point: 4{,}000 in (a, c), 2{,}500 in (b), and 3{,}000 in (d).}
  \label{fig:synthetic}
\end{figure}
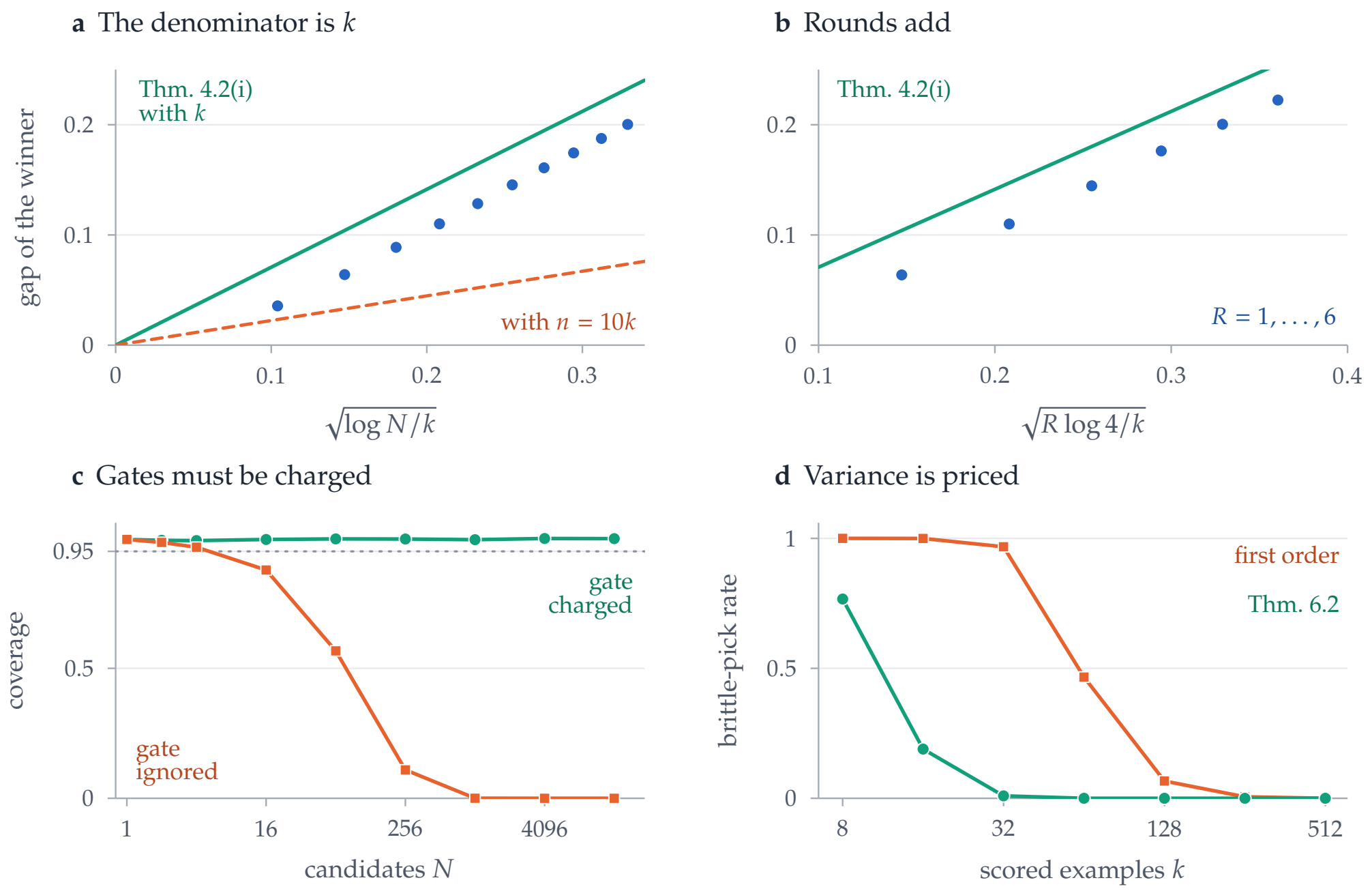

Controlled synthetic benchmarks isolate each structural element of the theory in a setting where candidate correlation
and risk profiles are prescribed exactly (\cref{fig:synthetic}).

\paragraph{Scored examples versus store size.}
When $N$ equal-risk prompts have independent Rademacher losses (\cref{thm:lower}), the winner's mean gap reaches
$48$--$86\%$ of the theoretical bound (\cref{fig:synthetic}a), confirming near-exactness in the independent worst case.
Replacing $k$ by the ambient store size $n=10k$ severely underpredicts the true optimism at every point: unselected store
items confer zero concentration.

\paragraph{Additive composition across rounds.}
Iterating four candidate edits across $R=1,\dots,6$ adaptive rounds on the same $k=64$ examples produces linear growth
against $\sqrt{R\log4/k}$ (\cref{fig:synthetic}b): because every round queries the same validation examples, selection
capacities accumulate across the editing history.

\paragraph{Necessity of shortlist gate charges.}
Restricting selection to a data-dependent gate containing the empirical winner exposes the failure of naive two-stage
accounting (\cref{fig:synthetic}c): omitting the gate charge $\log(1/\Pi(A))=\log N$ collapses nominal $95\%$ coverage
to $57\%$ at $N=64$, $0.025\%$ ($1/4{,}000$) at $N=1{,}024$, and $0\%$ from $N=4{,}096$, whereas charging the gate
maintains $\ge99\%$ coverage throughout.

\needspace{6\baselineskip}
\paragraph{Variance penalization against brittle prompts.}
In a mixed library of low-variance stable prompts (mean risk $\approx0.30$) and high-variance Bernoulli brittle prompts
(mean risk $\approx0.45$), extreme upward fluctuations cause first-order selection at $k=32$ to pick a suboptimal brittle
prompt in $97\%$ of trials; second-order SCAR (\cref{thm:second}) prices predictable variance and cuts that mistake rate
to $0.9\%$ (\cref{fig:synthetic}d), yielding strictly tighter certificates from $k=64$ onward ($0.446$ vs.\ $0.509$ at
$k=64$, and $0.290$ vs.\ $0.356$ at $k=512$; temperature ablations and GPT-5.2 validation-leakage stress tests appear in
\cref{app:experiments}).

\FloatBarrier

\section{Discussion}\label{sec:discussion}

\paragraph{Two budgets behind every reported score.}
Every empirical prompt evaluation spends from two budgets at once: the effective sample size $\keff$ of the examples it
scores, and the selection capacity $\kappa$ expended in choosing both the prompt and those examples. The transfer
principle (\cref{thm:transfer}) links them through the universal scale $\sqrt{\kappa/\keff}$, while the capacity calculus
(\cref{prop:calculus}) turns runtime logs into an auditable balance sheet across menus, shortlist gates, noisy score
releases, and iterative edits. Pairing a reported metric with its $\keff$, demonstration count $m$, capacity ledger, and
confidence level makes evaluation hygiene transparent, and clarifies why a fresh split read by zero stages resets the
selection surcharge to zero.

\paragraph{Randomization as evaluation-aware selection.}
Hard optimization on a small validation set pays for certainty it does not possess: committing to a single winning
recipe incurs the full $\log M$ menu price even when several candidates are statistically tied. By directly minimizing
the PAC-Bayesian bound, SCAR converts near-ties into information savings: on Banking77 it matches the greedy winner's
holdout error within $0.01$ while cutting selection cost from $4.16$ nats to $0.86$--$1.78$ nats and certifying
deployment gains at half the validation budget.

\paragraph{Retrieval quality versus score reliability.}
Improving a predictor and certifying its reported score are distinct statistical problems. We make this separation exact
for Bayesian linear in-context regression in \cref{sec:linear}: query-aligned retrieval genuinely lowers deployment risk
by concentrating the selected design's directional precision along the test query (quantified up to nuisance leakage via
an exact Schur complement in \cref{prop:schur,cor:topk}). Yet the moment a practitioner tunes the retriever, similarity
metric, or context length on the same data used to score the resulting prompt, the reported number incurs the adaptive
surcharge characterized here.

\paragraph{Open directions.}
Three theoretical questions emerge naturally from our results. First, while our $\sqrt{\kappa/\keff}$ certificates are
exact for independent equal-risk candidates (\cref{thm:lower}), structured prompt libraries induce strong cross-recipe
correlations; developing covariance-aware or chaining-based transfer bounds could close the constant-factor gap between
worst-case menus and correlated templates. Second, unrestricted deterministic index selection forces a nearly vacuous
$1-2k/n$ surcharge (\cref{rem:cherry}); certifying deterministic dense retrievers when they are scored on their own
retrieved pool will require pricing score-margin stability or embedding smoothness rather than combinatorial index
counts. Third, extending ghost decoupling beyond block-independent stores (\cref{ass:store}(b)) to corpora with
overlapping dependency graphs or mixing structure would broaden certification to web-scale retrieval stores.

\bibliographystyle{plainnat}
\bibliography{refs}

\clearpage
\appendix
\crefalias{section}{appendix}\crefalias{subsection}{appendix}
\pretocmd{\subsection}{\needspace{5\baselineskip}}{}{}
\pretocmd{\subsubsection}{\needspace{5\baselineskip}}{}{}
\addtocontents{toc}{\protect\addvspace{5pt}\protect\noindent{\protect\sffamily\protect\small\protect\bfseries\protect\color{muted}Appendix}\protect\par}
\noindent{\color{refcool}\rule{\linewidth}{1.6pt}}\par\vspace{8pt}
\noindent{\sffamily\bfseries\LARGE\color{navy}Appendix}\par\vspace{3pt}
\noindent{\sffamily\small\color{muted}The Price of Choosing Examples: Context Overfitting in Adaptive In-Context Learning}\par\vspace{6pt}
\noindent{\color{rulegray}\rule{\linewidth}{0.5pt}}\par\vspace{4pt}
\suppressfloats[t]

\begin{figure}[ht]
  \centering
  \begin{tikzpicture}[x=1pt,y=1pt,
    every node/.style={inner sep=0pt,outer sep=0pt},
    res/.style={rounded corners=3pt,line width=.5pt,minimum width=58pt,minimum height=21pt,align=center,
                font=\fontsize{6.2}{7.2}\selectfont,text=ink2c},
    main/.style={res,line width=.8pt},
    cert/.style={draw=aquac},         certmain/.style={main,draw=aquac,fill=aquac!13},
    sharp/.style={draw=orangec},      sharpmain/.style={main,draw=orangec,fill=orangec!11},
    design/.style={draw=bluec!80},
    dep/.style={-{Stealth[length=3.2pt,width=2.8pt]},line width=.55pt,draw=ink2c!85},
    lab/.style={font=\fontsize{6.2}{7.2}\selectfont,text=ink2c,anchor=west}]
  \def\rs#1#2{{\bfseries\color{inkc}#1}\\[-.4pt]#2}
  \path[use as bounding box] (0,-198) rectangle (396,11);
  \node[res,sharp] (B4) at (286,0) {\rs{Lem.~\ref*{lem:revchernoff}}{reverse Chernoff}};
  \node[res,sharp] (B6) at (350,0) {\rs{Lem.~\ref*{lem:types}}{method of types}};
  \node[res,cert] (D2) at (31,-36)  {\rs{Lem.~\ref*{lem:plugin}}{supermartingale}};
  \node[certmain] (L41) at (198,-36) {\rs{Lem.~\ref*{lem:ghost}}{ghost concentration}};
  \node[res,sharp] (B5) at (286,-36) {\rs{Lem.~\ref*{lem:radmax}}{maximal fluctuation}};
  \node[certmain] (T62) at (70,-72)  {\rs{Thm.~\ref*{thm:second}}{second-order SCAR}};
  \node[certmain] (T42) at (198,-72) {\rs{Thm.~\ref*{thm:transfer}}{score transfer}};
  \node[res,sharp] (B7) at (330,-72) {\rs{Lem.~\ref*{lem:radmax-sharp}}{sharp fluctuation}};
  \node[res,cert] (RB1) at (31,-114)  {\rs{Rem.~\ref*{rem:maxinfo}}{max-information}};
  \node[res,cert] (PB2) at (95,-114)  {\rs{Prop.~\ref*{prop:sibson}}{Sibson ladder}};
  \node[certmain] (C1) at (159,-114)  {\rs{Prop.~\ref*{prop:calculus}\,/\,\ref*{prop:calculus-formal}}{capacity calculus}};
  \node[res,cert] (PB3) at (223,-114) {\rs{Prop.~\ref*{prop:subgamma}}{sub-gamma transport}};
  \node[sharpmain] (T43) at (287,-114) {\rs{Thm.~\ref*{thm:lower}}{the price is sharp}};
  \node[sharpmain] (R52) at (31,-150) {\rs{Rem.~\ref*{rem:cherry}}{cherry-picking}};
  \node[certmain] (T61) at (159,-150) {\rs{Thm.~\ref*{thm:scar}}{SCAR certificate}};
  \node[res,sharp] (RB8) at (322,-150) {\rs{Rem.~\ref*{rem:exact-scale}}{exact constants}};
  \node[res,design] (E1) at (31,-186)  {\rs{Prop.~\ref*{prop:linear}}{retrieval and design}};
  \node[res,design] (E2) at (95,-186)  {\rs{Prop.~\ref*{prop:schur}}{Schur complement}};
  \node[res,design] (E3) at (159,-186) {\rs{Cor.~\ref*{cor:topk}}{top-$k$ retrieval}};
  \node[res,cert] (D1) at (262,-186)  {\rs{Prop.~\ref*{prop:scar-optimal}}{rate optimality}};
  \draw[dep] (D2) -- (T62);
  \draw[dep] (L41) -- (T42);
  \draw[dep] (B4) -- (B5);
  \draw[dep] (B5) -- (B7);
  \draw[dep] (B6) -- (B7);
  \draw[dep] (T42) -- (RB1);
  \draw[dep] (T42) -- (PB2);
  \draw[dep] (T42) -- (C1);
  \draw[dep] (T42) -- (PB3);
  \draw[dep] (T42) -- (T43);
  \draw[dep] (B7) -- (T43);
  \draw[dep] (B7) -- (RB8);
  \draw[dep] (PB3) -- (RB8);
  \draw[dep] (C1) -- (T61);
  \draw[dep] (T61) -- (D1);
  \draw[dep,rounded corners=4pt] (B6.east) -- (391,0) -- (391,-186) -- (D1.east);
  \draw[dep] (E1) -- (E2);
  \draw[dep] (E2) -- (E3);
  \begin{scope}[shift={(0,4)}]
    \node[res,draw=aquac,fill=aquac!13,minimum width=9pt,minimum height=6pt,line width=.6pt] at (6,0) {};
    \node[lab] at (13,0) {certificate};
    \node[res,draw=orangec,fill=orangec!11,minimum width=9pt,minimum height=6pt,line width=.6pt] at (62,0) {};
    \node[lab] at (69,0) {sharpness};
    \node[res,draw=bluec!80,minimum width=9pt,minimum height=6pt] at (116,0) {};
    \node[lab] at (123,0) {retrieval design};
    \node[lab] at (1.5,-11) {filled: main text;\ \ outlined: appendix};
  \end{scope}
\end{tikzpicture}
  \caption{\textbf{Logical architecture of the formal results.} Directed edges trace proof dependencies (shown in
  transitive reduction), with \cref{thm:transfer} serving as the central hub. Setup primitives (\cref{sec:setup}) and
  standard literature tools (\cref{app:aux}) are left implicit.}
  \label{fig:deps}
\end{figure}
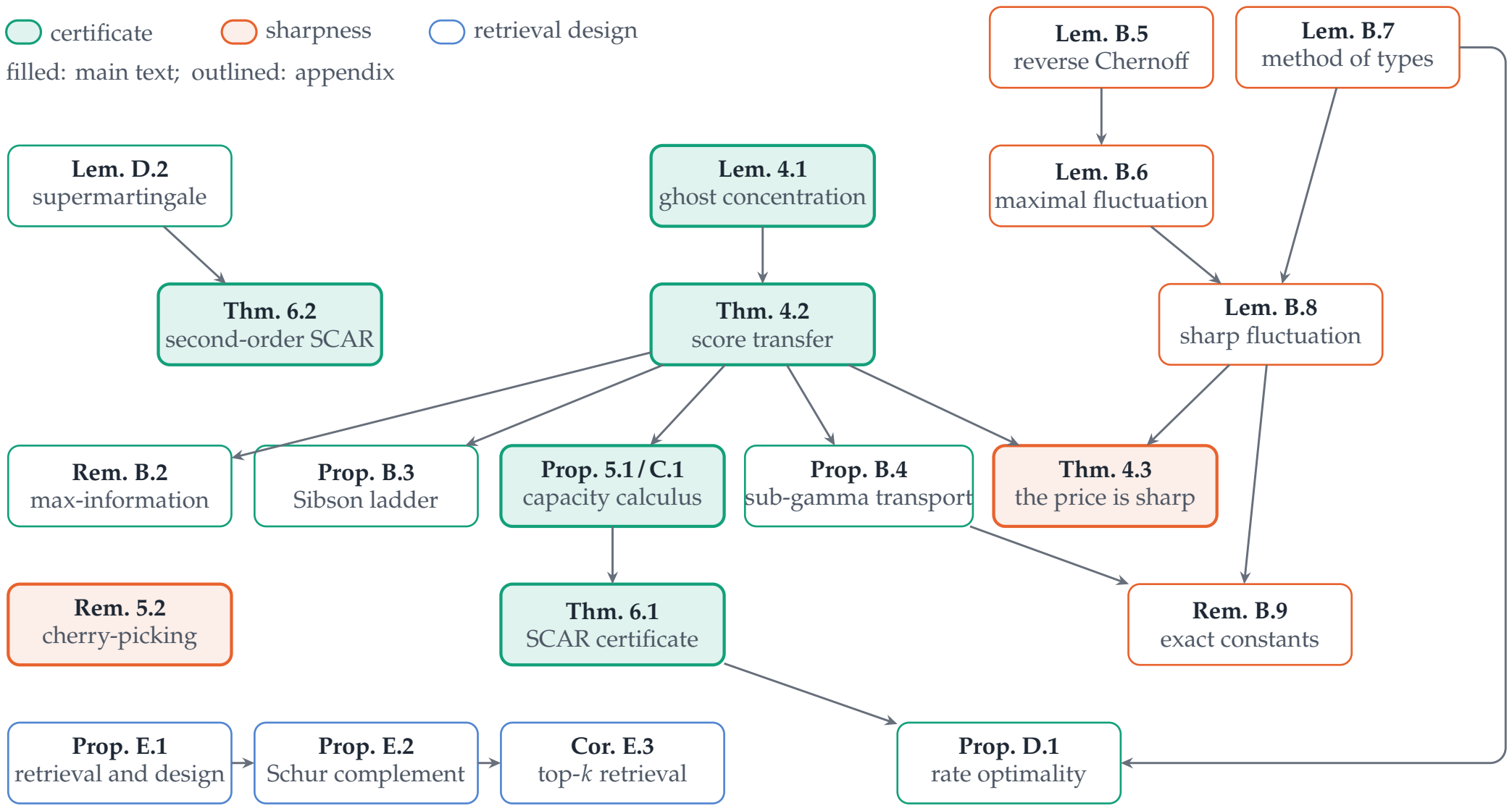

\paragraph{Organization.} The appendix is organized into four parts. \emph{Preliminaries} (\cref{app:prelim}) summarizes
notation and collects the variational, information-theoretic, concentration, and minimax tools invoked in the proofs.
\emph{Theory} (\cref{app:proofs-transfer,app:proofs-calculus,app:proofs-scar,app:linear}) restates and proves every
formal claim of the main text alongside R\'enyi/Sibson and sub-gamma extensions, minimax optimality of SCAR, and the
exact Schur-complement analysis of Bayesian linear in-context regression. \emph{Experiments} (\cref{app:experiments})
documents full evaluation protocols, compute, finite-sample uncertainty, and supplementary ablations. \emph{Practice}
(\cref{app:practice}) translates the capacity calculus into runtime logging tables, an audit checklist, deployment
diagnostics, scope boundaries, and broader impact. \Cref{fig:deps} maps the logical dependencies among the formal
results.

\section{Preliminaries}\label{app:prelim}

This appendix fixes notation (\cref{app:notation}) and restates the results from the literature that the proofs
invoke (\cref{app:aux}).

\subsection{Notation}\label{app:notation}

\Cref{tab:notation} collects the notation used throughout. The formal definitions are in \cref{sec:setup}
(\cref{def:transcript,def:laws} and \cref{ass:store}) and in \cref{sec:scar}.

\begin{table}[ht]
  \centering
  \caption{\textbf{Notation.}}
  \label{tab:notation}
  \footnotesize
  \renewcommand{\arraystretch}{1.12}
  \setlength{\tabcolsep}{3.4pt}
  \begin{tabular}{L{0.18\linewidth}L{0.25\linewidth}L{0.24\linewidth}L{0.26\linewidth}}
    \toprule
    \headrow\textbf{symbol} & \textbf{meaning} & \textbf{symbol} & \textbf{meaning}\\
    \midrule
    $U$, $u$ & query, public side information & $\Lhat_S(t)$ & selected score, $\sum_jw_j(t)\ell_t(Z_{i_j(t)})$\\
    $S=(Z_1,\dots,Z_n)$ & scoring store, law $\Pstore_u$ given $U=u$ & $\gap_u(t,S)$ & $L_u(t)-\Lhat_S(t)$\\
    $K_u(\dd t\mid s)$ & pipeline kernel & $k$, $\keff(t)$ & scoring sample size; $\keff(t)=1/\sum_g\pi_g(t)^2$\\
    $T$, $\calT$ & transcript and its space & $\MI(T;S\cond U)$ & expected selection capacity\\
    $Q_u$, $\Qbar_u$ & real and ghost laws of $(S,T)$ & $\Leak(S\to T\cond U=u)$ & realized-gap capacity (maximal leakage)\\
    $\Kbar_u$ & marginal law of $T$ & $\Pi_u$, $Q$ & prior and posterior on $\calT$\\
    $\ell_t$, $L_u(t)$ & loss of prompt $t$ and its risk & $A(S,u)$, $\Pi_{u,A}$ & gate and conditioned prior\\
    $i_j(t)$, $w_j(t)$ & selected indices and weights & $\lambda$, $\gamma$, $\Gamma$ & inverse temperatures and grid\\
    $B_g$, $\pi_g(t)$ & blocks and selected block weight & $\Vhat_S(t)$, $\psi(\gamma)$ & predictable variance, $-\log(1-\gamma)-\gamma$\\
    \bottomrule
  \end{tabular}
\end{table}

\subsection{Auxiliary results}\label{app:aux}

The expected-gap argument uses information-theoretic decoupling
\citep{russo2016controlling,xu2017information}; the realized-gap argument uses maximal-leakage transfer
\citep{issa2020operational,esposito2021generalization}; and the posterior argument uses Catoni-type PAC-Bayes
\citep{catoni2007pacbayesian}. The following results state these tools, together with the concentration,
composition, and lower-bound facts, in the form used by the proofs.

\paragraph{Change of measure and information measures.}

Every certificate returns from the ghost law to the real law through one of these facts. The variational principle
drives \cref{thm:transfer}(i),(iii) and \cref{thm:scar,thm:second}. The leakage facts drive \cref{thm:transfer}(ii),
\cref{prop:sibson}, and the capacity calculus (\cref{prop:calculus-formal}).

\begin{lemma}[{Donsker--Varadhan; Gibbs variational principle \citep{donsker1975asymptotic},
\citep[Lemma~1.1.3]{catoni2007pacbayesian}}]\label[lemma]{lem:dv}
Let $Q\ll P$ be probability measures on a common space.
\begin{enumerate}[label=(\roman*),leftmargin=2em]
\item If $f$ is measurable with $\E_Pe^f<\infty$, then $\E_Qf\le\KL(Q\|P)+\log\E_Pe^f$.
\item If $f$ is measurable with $Z\defeq\E_Pe^{-f}<\infty$, let $\dd P_f/\dd P=e^{-f}/Z$. For every $Q$ with
$\KL(Q\|P)<\infty$ and $\E_Q|f|<\infty$,
\[
  \E_Qf+\KL(Q\|P)=-\log Z+\KL(Q\|P_f),
\]
so $\E_Qf+\KL(Q\|P)\ge-\log Z$, with equality if and only if $Q$ is the Gibbs law $P_f$.
\end{enumerate}
\end{lemma}

Catoni states (ii) for bounded $f$; the same computation gives it under the integrability stated here.

\begin{lemma}[{Golden formula \citep[Thm.~4.1]{polyanskiy2025information}}]\label[lemma]{lem:golden}
Let $X\sim P_X$, let $Y$ given $X$ have law $P_{Y\mid X}$ with marginal $P_Y$, and let $R$ be any law on the range of
$Y$. Then
\[
  \E_{X}\KL\bigl(P_{Y\mid X}\,\|\,R\bigr)=\MI(X;Y)+\KL(P_Y\|R).
\]
In particular $\MI(X;Y)=\min_R\E_X\KL(P_{Y\mid X}\|R)$, attained at $R=P_Y$.
\end{lemma}

\begin{lemma}[{Maximal leakage \citep{issa2020operational}}]\label[lemma]{lem:leakage}
Let $X$ and $Y$ take values in standard Borel spaces, and let the law of $Y$ given $X=x$ have a jointly measurable
density $k(y\mid x)$ with respect to a $\sigma$-finite measure $\mu$. The maximal leakage
\[
  \Leak(X\to Y)=\log\int\esssup_{x\sim P_X}k(y\mid x)\,\mu(\dd y)
\]
does not depend on $\mu$ or on the version of $k$, and:
\begin{enumerate}[label=(\roman*),leftmargin=2em]
\item $\Leak(X\to Y)\ge\MI(X;Y)$;
\item if $Y$ takes at most $M$ values, then $\Leak(X\to Y)\le\log M$;
\item if $X$, $Y$, $Z$ form a Markov chain $X-Y-Z$, then $\Leak(X\to Z)\le\Leak(X\to Y)$.
\end{enumerate}
\end{lemma}

\begin{lemma}[{Transfer through maximal leakage and Sibson's information \citep{esposito2021generalization}}]\label[lemma]{lem:leak-transfer}
Let $(X,Y)\sim P_{XY}$, let $E$ be an event, and let $E_y=\{x:(x,y)\in E\}$ be its sections.
\begin{enumerate}[label=(\roman*),leftmargin=2em]
\item $P_{XY}(E)\le e^{\Leak(X\to Y)}\,\esssup_{y\sim P_Y}P_X(E_y)$.
\item For $\alpha\in(1,\infty)$, let $\Leak^{(\alpha)}(X\to Y)=\min_RD_\alpha(P_{XY}\|P_X\otimes R)$ be Sibson's information
of order $\alpha$ \citep{sibson1969information}, where $D_\alpha$ is the R\'enyi divergence of order $\alpha$. Then
\[
  P_{XY}(E)\le\exp\Bigl\{\frac{\alpha-1}{\alpha}\Leak^{(\alpha)}(X\to Y)\Bigr\}\Bigl(\esssup_{y\sim P_Y}P_X(E_y)\Bigr)^{\frac{\alpha-1}{\alpha}}.
\]
\end{enumerate}
\end{lemma}

\paragraph{Concentration.}

Ghost concentration (\cref{lem:ghost}) rests on Hoeffding's lemma. The second-order certificate (\cref{thm:second})
rests on the inequality of Fan et al. \Cref{rem:cherry} and \cref{cor:topk} rest on the multiplicative Chernoff bound.

\begin{lemma}[{Hoeffding's lemma \citep{hoeffding1963probability}; tensorization \citep{boucheron2013concentration}}]\label[lemma]{lem:hoeffding}
If $X\in[a,b]$ almost surely, then $\log\E e^{\lambda(X-\E X)}\le\lambda^2(b-a)^2/8$ for all $\lambda\in\R$. If
$X_1,\dots,X_m$ are independent and centered with $\log\E e^{\lambda X_j}\le\nu_j\lambda^2/2$, then $\sum_jw_jX_j$ is
centered with variance proxy $\sum_jw_j^2\nu_j$.
\end{lemma}

\begin{lemma}[{\citet[proof of Lemma~4.1, eq.~(4.12)]{fan2015exponential}}]\label[lemma]{lem:fan}
If $\xi\ge-1$ and $\gamma\in[0,1)$, then $\exp\{\gamma\xi-\psi(\gamma)\xi^2\}\le1+\gamma\xi$, where $\psi(\gamma)=-\log(1-\gamma)-\gamma$.
\end{lemma}

\begin{lemma}[{Multiplicative Chernoff bound \citep[Thm.~1.10.5]{doerr2018probabilistic}}]\label[lemma]{lem:chernoff}
Let $B$ be a sum of independent random variables with values in $[0,1]$, and let $\mu$ be the mean of $B$. For every
$\eta\in[0,1]$,
\[
  \Pp\bigl(B\le(1-\eta)\mu\bigr)\le e^{-\eta^2\mu/2};\qquad\text{in particular}\qquad\Pp(B\le\mu/2)\le e^{-\mu/8}.
\]
\end{lemma}

\paragraph{Lower-bound tools.}

The reverse Chernoff bound gives the winner's-curse lower bound of \cref{thm:lower} (through \cref{lem:revchernoff}),
and Fano's inequality gives the minimax lower bound of \cref{prop:scar-optimal}.

\begin{lemma}[{Reverse Chernoff bound for the symmetric binomial \citep[Prop.~7.3.2]{matousek2008probabilistic}}]\label[lemma]{lem:mv}
Let $n$ be even and let $X\sim\mathrm{Bin}(n,1/2)$. For every integer $t\in[0,n/8]$,
\[
  \Pp\bigl(X\ge n/2+t\bigr)\ge\frac1{15}\,e^{-16t^2/n}.
\]
\end{lemma}

\begin{lemma}[{Fano's inequality \citep[\S2.10]{cover2006elements}, for a uniform prior}]\label[lemma]{lem:fano}
Let $J$ be uniform on $\{1,\dots,M\}$ with $M\ge2$, and let $\hat J$ be any estimate of $J$ computed from an observation
$X$ and randomness independent of $(J,X)$. Then
\[
  \Pp(\hat J\ne J)\ge1-\frac{\MI(J;X)+\log2}{\log M}.
\]
\end{lemma}

\section{The transfer principle}\label{app:proofs-transfer}

This appendix proves the results of \cref{sec:transfer} and refines them. Ghost concentration
(\cref{app:ghost}) and one change of measure per reading (\cref{app:transfer}) give the expected, realized, and
posterior certificates. \Cref{app:sibson,app:subgamma} interpolate between the readings and adapt them to the
variance of the losses. \Cref{app:lower} shows that the rate cannot be improved. We restate results of the main text
before their proofs.

\subsection{Ghost concentration}\label{app:ghost}

\ghostlemma*

\begin{proof}
Fix $u$ and $t$. The selected indices $i_1(t),\dots,i_m(t)$ are distinct constants, and $S^\circ$ is drawn from the
store law, independently of $t$. By definition,
\[
  \gap_u(t,S^\circ)=\sum_jw_j(t)\bigl(L_u(t)-\ell_t(Z^\circ_{i_j(t)})\bigr),
\]
and each selected item has marginal law $P_u$, so the gap is centered.

\emph{Case (a).} The selected items are i.i.d.\ from $P_u$, so the summands are independent, centered, and
$\sigma^2$-sub-Gaussian; \cref{lem:hoeffding} gives the variance proxy $\sigma^2\sum_jw_j(t)^2=\sigma^2/\keff(t)$.

\emph{Case (b).} Group the selected items by block, so that $\Lhat_{S^\circ}(t)=\sum_gA_g$, where
\[
  A_g\defeq\sum_{j:\,i_j(t)\in B_g}w_j(t)\,\ell_t(Z^\circ_{i_j(t)})\in[0,\pi_g(t)]
\]
because $\ell_t\in[0,1]$. Each $A_g$ is a function of block $g$ only, so the $A_g$ are independent, and Hoeffding's
lemma gives $\log\E e^{\lambda(\E A_g-A_g)}\le\lambda^2\pi_g(t)^2/8$. Summing over independent blocks, the gap is
sub-Gaussian with variance proxy
\[
  \frac14\sum_g\pi_g(t)^2=\frac{\sigma^2}{\keff(t)},\qquad\sigma^2=\frac14 .\qedhere
\]
\end{proof}

\subsection{Three readings of one change of measure}\label{app:transfer}

\transferthm*

\begin{proof}
Each part combines \cref{lem:ghost} with one change of measure. Throughout, $\keff\le\keff(t)$ for every $u$ and
every $t\in\calT_0$, where $K_u(\calT_0\mid s)=1$ in (i)--(ii) and $\Pi_u(\calT_0)=1$ in (iii). Every supremum over $t$
ranges over $\calT_0$.

\emph{(i) Expected gap.} Fix $u$ and write $I_u=\KL(Q_u\|\Qbar_u)$, the mutual information of $T$ and $S$ given
$U=u$. If $I_u=\infty$, there is nothing to prove. For $\lambda\in\R$, \cref{lem:dv}(i) with $f(s,t)=\lambda\gap_u(t,s)$
gives
\[
  \lambda\,\E_{Q_u}\gap_u(T,S)\le I_u+\log\E_{t\sim\Kbar_u}\E_{S^\circ}e^{\lambda\gap_u(t,S^\circ)}
  \le I_u+\frac{\lambda^2\sigma^2}{2\keff},
\]
because under $\Qbar_u$ the store is independent of the transcript, $\Kbar_u(\calT_0)=1$, and \cref{lem:ghost}
applies to every $t\in\calT_0$. Optimizing over $\lambda>0$ and over $\lambda<0$ gives
\[
  \bigl|\E[\gap_U(T,S)\mid U=u]\bigr|\le\sqrt{2\sigma^2I_u/\keff}.
\]
Averaging over $U$ and applying Jensen's inequality to the concave square root gives (i).

\emph{(ii) Realized gap.} Fix $u$ and let
\[
  \epsilon=\sqrt{2\sigma^2\bigl(\kappa+\log(1/\delta)\bigr)/\keff},\qquad
  E=\{(s,t):\gap_u(t,s)>\epsilon\},\qquad E_t=\{s:(s,t)\in E\}.
\]
For every $t\in\calT_0$, Chernoff's bound and \cref{lem:ghost} give
\[
  \Pstore_u(E_t)=\Pp\bigl(\gap_u(t,S^\circ)>\epsilon\bigr)\le e^{-\keff\epsilon^2/(2\sigma^2)}=\delta e^{-\kappa}.
\]
Take $k_u$ and $\mu$ as in \cref{def:laws}, with $k_u(t\mid s)=0$ for $t\notin\calT_0$, which is possible because
$K_u(\calT_0\mid s)=1$. Then
\begin{align*}
  Q_u(E)&=\int\mu(\dd t)\int_{E_t}k_u(t\mid s)\,\Pstore_u(\dd s)
  \le\int\mu(\dd t)\,\esssup_{s}k_u(t\mid s)\,\Pstore_u(E_t)\\
  &\le e^{\Leak(S\to T\cond U=u)}\,\delta e^{-\kappa}\le\delta .
\end{align*}
Averaging over $U$ proves the upper tail, and the same argument applied to $-\gap$ proves the lower tail. This is the
transfer of \cref{lem:leak-transfer}(i), applied to the selected score.

\emph{(iii) Posterior risk.} Fix $u$, $\lambda$, and $\delta$, and let
\[
  f_S(t)=\lambda\gap_u(t,S)-\frac{\lambda^2\sigma^2}{2\keff}.
\]
For every $t\in\calT_0$, \cref{lem:ghost} gives $\E_Se^{f_S(t)}\le1$. Because $\Pi_u$ is fixed before $S$ is drawn,
Fubini's theorem gives $\E_S\E_{t\sim\Pi_u}e^{f_S(t)}\le1$. Markov's inequality then gives
$\Pp\bigl(\E_{\Pi_u}e^{f_S}>1/\delta\bigr)\le\delta$. On the complementary event, \cref{lem:dv}(i) with $P=\Pi_u$ gives,
for every $Q\ll\Pi_u$, including any $Q$ that depends on $S$,
\[
  \lambda\,\E_{t\sim Q}\gap_u(t,S)-\frac{\lambda^2\sigma^2}{2\keff}\le\KL(Q\|\Pi_u)+\log(1/\delta).
\]
Dividing by $\lambda$ proves (iii).
\end{proof}

\begin{remark}[Approximate max-information]\label[remark]{rem:maxinfo}
Suppose that, in place of the leakage bound of \cref{thm:transfer}(ii), for every event $E$,
\[
  Q_u(E)\le e^\kappa\,\Qbar_u(E)+\beta,
\]
that is, the $\beta$-approximate max-information, measured in nats, is at most $\kappa$ \citep{dwork2015generalization}. The same
computation with $\Qbar_u(E)\le\sup_{t\in\calT_0}\Pstore_u(E_t)\le\delta e^{-\kappa}$ gives (ii) with probability at least
$1-\delta-\beta$. This variant suits differentially private releases, whose max-information on i.i.d.\ stores can be
far smaller than their maximal leakage. For a menu of $M$ candidates, the variant costs $\log(M/\beta)$
(\cref{app:maxinfo-rules}) rather than $\log M$. The \emph{max-information} rows of \cref{tab:scar,tab:rates} use this
variant.
\end{remark}

\subsection{A common divergence representation}\label{app:divergences}
All three prices are one quantity (\cref{tab:divergences}): the R\'enyi divergence $D_\alpha(Q_u\,\|\,\Pstore_u\otimes R)$ of the real law from the
store law paired with a reference law $R$ on transcripts. Two choices specialize it: the order $\alpha$, and whether $R$
is optimized or fixed in advance. Optimizing $R$ at order $1$ gives the mutual information (the optimal reference is
$\Kbar_u$). Optimizing at order $\infty$ gives the maximal leakage. Fixing $R=\Pi_u$ at order $1$ gives
$\E_S\KL(Q_S\|\Pi_u)$ with $Q_S=K_u(\cdot\mid S)$, which (iii) prices pathwise. The orders in between are Sibson's informations
\citep{sibson1969information}. The argument of (ii) transfers at each of them and trades capacity against
confidence (\cref{app:sibson}).

\begin{table}[t]
  \centering
  \caption{\textbf{Three readings of one divergence.} Each certificate is ghost concentration at scale
  $\sigma^2/\keff$ (\cref{lem:ghost}) plus a price for returning to the real law. The price is the order-$\alpha$ divergence of that law from the
  store law paired with a reference law $R$ on transcripts; $R$ is optimized or fixed in advance.
  The last two columns give the price of a menu of $M$ fixed candidates and the certificate for losses in $[0,1]$
  ($\sigma^2=\tfrac14$).
  \cref{prop:subgamma,thm:second} give variance-adaptive forms of (i) and (iii).}
  \label{tab:divergences}
  \small
  \setlength{\tabcolsep}{3.4pt}
  \begin{tabular}{lllllc}
    \toprule
    \headrow\textbf{reading} & \textbf{certifies} & \textbf{price} & {\boldmath$(\alpha,R)$} & \textbf{menu of {\boldmath$M$}} & \textbf{certificate}\\
    \midrule
    (i) expected & $\E\,\gap$ & $\MI(T;S\cond U)$ & $1$, optimized & $\log M$ & $\sqrt{\MI/(2\keff)}$\\
    (ii) realized & $\gap$, w.p.\ $1-\delta$ & $\Leak(S\to T\cond U)$ & $\infty$, optimized & $\log M$ &
      $\sqrt{(\Leak+\log\frac1\delta)/(2\keff)}$\\
    (iii) posterior & $\E_Q L$ $\forall Q$, w.p.\ $1-\delta$ & $\KL(Q\,\|\,\Pi)$ & $1$, fixed ($\Pi$) & logged &
      $\frac{\KL+\log(1/\delta)}{\lambda}+\frac{\lambda}{8\keff}$\\
    \bottomrule
  \end{tabular}
\end{table}

\subsection{The Sibson ladder}\label{app:sibson}
For $\alpha\in(1,\infty)$, Sibson's information of order $\alpha$ \citep{sibson1969information} is
\begin{align*}
  \Leak^{(\alpha)}(S\to T\cond U=u)&\defeq\min_R D_\alpha\bigl(Q_u\,\|\,\Pstore_u\otimes R\bigr)\\
  &=\frac{\alpha}{\alpha-1}\log\int_{\calT}\Bigl(\int k_u(t\mid s)^\alpha\,\Pstore_u(\dd s)\Bigr)^{1/\alpha}\mu(\dd t),
\end{align*}
where $D_\alpha$ is the R\'enyi divergence of order $\alpha$ and $R$ ranges over laws on $\calT$, with $k_u$ and $\mu$
as in \cref{def:laws}. Sibson's information is nondecreasing in $\alpha$, tends to $\Leak(S\to T\cond U=u)$ as $\alpha\to\infty$, and tends to $\MI(T;S\cond U=u)$ as
$\alpha\downarrow1$ whenever it is finite for some $\alpha>1$ \citep{esposito2021generalization}. At order $1$, the
minimizing $R$ is $\Kbar_u$, by the golden formula (\cref{lem:golden}).

\begin{proposition}[Transfer at every order]\label[proposition]{prop:sibson}
Under the hypotheses of \cref{thm:transfer}, if $\Leak^{(\alpha)}(S\to T\cond U=u)\le\kappa$ for $P_U$-almost every $u$,
then for every $\delta\in(0,1)$, with probability at least $1-\delta$,
\[
  \gap_U(T,S)\le\sqrt{\frac{2\sigma^2}{\keff}\Bigl(\kappa+\frac{\alpha}{\alpha-1}\log\frac1\delta\Bigr)}.
\]
\end{proposition}

\begin{proof}
Let $E$, $E_t$, and $k_u$ be as in the proof of \cref{thm:transfer}(ii), now with $\epsilon$ defined by
\[
  \frac{\keff\epsilon^2}{2\sigma^2}=\kappa+\frac{\alpha}{\alpha-1}\log\frac1\delta .
\]
As in the argument of \cref{lem:leak-transfer}(ii), H\"older's inequality with exponents $\alpha$ and
$\alpha/(\alpha-1)$ in the inner integral gives
\begin{align*}
  Q_u(E)&\le\int\mu(\dd t)\Bigl(\int k_u(t\mid s)^\alpha\,\Pstore_u(\dd s)\Bigr)^{1/\alpha}\Pstore_u(E_t)^{\frac{\alpha-1}{\alpha}}\\
  &\le e^{\frac{\alpha-1}{\alpha}\Leak^{(\alpha)}(S\to T\cond U=u)}\sup_{t\in\calT_0}\Pstore_u(E_t)^{\frac{\alpha-1}{\alpha}}.
\end{align*}
Since $\sup_t\Pstore_u(E_t)\le e^{-\keff\epsilon^2/(2\sigma^2)}$, the right side is at most
\[
  \exp\Bigl\{\frac{\alpha-1}{\alpha}\Bigl(\kappa-\frac{\keff\epsilon^2}{2\sigma^2}\Bigr)\Bigr\}=\delta .\qedhere
\]
\end{proof}

Smaller orders charge less capacity but multiply the confidence term by $\alpha/(\alpha-1)$, which diverges as
$\alpha\downarrow1$. At $\alpha=\infty$, the proposition is \cref{thm:transfer}(ii).

\subsection{Variance-adaptive transport}\label{app:subgamma}

\begin{proposition}[Sub-gamma transport]\label[proposition]{prop:subgamma}
Suppose that for every $u$ and every $t\in\calT_0$, where $\Kbar_u(\calT_0)=1$, the ghost cumulant satisfies
$\log\E_{S^\circ}e^{\lambda\gap_u(t,S^\circ)}\le\lambda^2v/(2(1-b\lambda))$ for $\lambda\in[0,1/b)$, with $1/0=\infty$.
With $\MI=\MI(T;S\cond U)$,
\[
  \E\,\gap_U(T,S)\;\le\;\inf_{\lambda\in(0,1/b)}\Bigl\{\frac{\MI}{\lambda}+\frac{\lambda v}{2(1-b\lambda)}\Bigr\}
  \;=\;\sqrt{2v\,\MI}+b\,\MI .
\]
\end{proposition}

\begin{proof}
With $I_u$ the mutual information of $T$ and $S$ given $U=u$, the argument of \cref{thm:transfer}(i), run with the
sub-gamma envelope in place of the sub-Gaussian one, gives
\[
  \E[\gap_U\mid U=u]\le\inf_{\lambda\in(0,1/b)}\Bigl\{\frac{I_u}{\lambda}+\frac{\lambda v}{2(1-b\lambda)}\Bigr\}.
\]
For $I_u>0$ the objective is convex on $(0,1/b)$ and is minimized at
\[
  \lambda^\star=\frac{\sqrt{2I_u}}{\sqrt v+b\sqrt{2I_u}},\qquad\text{where}\qquad
  \frac{\lambda^\star}{1-b\lambda^\star}=\sqrt{\frac{2I_u}{v}},
\]
and its minimum is $\sqrt{2vI_u}+bI_u$, which is also the infimum $0$ at $I_u=0$
(cf.\ \citealp[Lemma~2.4]{boucheron2013concentration}). The minimum is a concave function of $I_u$, so Jensen's inequality
over $U$ completes the proof. This is the bound of \citet[Thm.~2]{jiao2017dependence} and
\citet[Thm.~1]{bu2020tightening} for the ghost cumulant of the selected score.
\end{proof}

Consider case (a) of \cref{ass:store} with uniform weights over $k$ items and losses in $[0,1]$ whose variance under
$P_u$ is at most $s^2$. Each summand $(L_u(t)-\ell_t(Z))/k$ is centered, is at most $1/k$, and has variance at most
$s^2/k^2$. Bernstein's inequality therefore gives the envelope with $v=s^2/k$ and $b=1/(3k)$, and the bound becomes
\[
  \E\,\gap_U(T,S)\le\sqrt{\frac{2s^2\MI}{k}}+\frac{\MI}{3k},
\]
so the variance $s^2$ of a prompt's losses replaces the sub-Gaussian proxy $\sigma^2=1/4$. For an $R$-round pipeline,
\cref{prop:calculus-formal}(a) lets us replace $\MI$ by the capacity profile
$\sum_{r}\MI(T_r;S\cond U,T_{<r})$.

\subsection{Sharpness: the winner's-curse construction}\label{app:lower}

The construction selects the best of $N$ equal-risk prompts, so its gap is half the maximum of $N$ independent
Rademacher averages. Four lemmas bound that maximum from below, first up to a constant and then with
the exact constant, and the theorem follows from them.

\begin{lemma}[Reverse Chernoff bound for Rademacher sums]\label[lemma]{lem:revchernoff}
Let $k\ge16$, let $0\le a\le1/32$, and let $S_k$ be a sum of $k$ independent uniform signs. Then
\[
  \Pp(S_k\ge ka)\ge\frac{1}{15}\,e^{-3.2}\exp\Bigl\{-\frac{64}{15}\,ka^2\Bigr\}.
\]
\end{lemma}

\begin{proof}
\proofstep{Step 1 (an even number of signs).} Let $k'=2\lfloor k/2\rfloor\ge k-1\ge15$, so $k'\ge16$, and let $S'$ be
the sum of the first $k'$ signs. Then $S_k\ge S'-1$.

\proofstep{Step 2 (an integer deviation).} Let $r=\lceil(ka+1)/2\rceil$, an integer with $2r\ge ka+1$ and
$r\le ka/2+3/2$. Since $a\le1/32$ and $k\ge16$,
\[
  \frac{ka}2+\frac32\le\frac{k}{64}+\frac32\le\frac{k-1}8\le\frac{k'}8 .
\]

\proofstep{Step 3 (the reverse Chernoff bound).} With $X'=(S'+k')/2\sim\mathrm{Bin}(k',1/2)$,
\[
  \Pp(S_k\ge ka)\ge\Pp(S'\ge2r)=\Pp(X'\ge k'/2+r)\ge\tfrac1{15}e^{-16r^2/k'}
\]
by \cref{lem:mv}, which applies to even $k'$ and integer $r\in[0,k'/8]$.

\proofstep{Step 4 (the exponent).} Using $k/(k-1)\le16/15$ and $a\le1/32$,
\[
  \frac{16r^2}{k'}\le\frac{16(ka/2+3/2)^2}{k-1}=\frac{4k^2a^2+24ka+36}{k-1}\le\frac{64}{15}\,ka^2+0.8+2.4 .\qedhere
\]
\end{proof}

\begin{lemma}[Maximal Rademacher fluctuation]\label[lemma]{lem:radmax}
There is a universal $c_0>0$ such that the following holds. Let $X_j=k^{-1}\sum_{m=1}^k\epsilon_{mj}$ for $j\le N$ and
independent uniform signs $\epsilon_{mj}$, with $k\ge16$ and $2\le N\le e^{k/16}$. Then
\[
  \E\max_{j\le N}X_j\ge c_0\sqrt{\log N/k}.
\]
\end{lemma}

\begin{proof}
We pass from the tail of one average to the mean of the maximum, treating small $N$ separately.

\proofstep{Step 1 (one average).} Let $a=\sqrt{\log N/(64k)}$, so that $a\le1/32$. By \cref{lem:revchernoff},
\[
  p\defeq\Pp(X_1\ge a)\ge c_1N^{-1/15},\qquad c_1=e^{-3.2}/15 .
\]

\proofstep{Step 2 (the maximum rarely stays low).} By independence,
\[
  \eta_N\defeq\Pp\bigl(\max_jX_j<a\bigr)\le(1-p)^N\le\exp\{-c_1N^{14/15}\}\eqqcolon\bar\eta_N .
\]

\proofstep{Step 3 (from the tail to the mean).} Since $\max_jX_j\ge X_1$,
\[
  \max_jX_j\ge a\,\mathbf 1\{\max_jX_j\ge a\}+X_1\mathbf 1\{\max_jX_j<a\},
\]
and the Cauchy--Schwarz inequality with $\E X_1^2=1/k$ gives
\[
  \E\max_jX_j\ge a(1-\eta_N)-\sqrt{\eta_N/k}=k^{-1/2}\Bigl[\frac{(1-\eta_N)\sqrt{\log N}}{8}-\sqrt{\eta_N}\Bigr].
\]

\proofstep{Step 4 (large $N$).} Since $\eta_N\le\bar\eta_N$, the bracket minus $\sqrt{\log N}/16$ is at least
\[
  \sqrt{\log N}\,\Bigl(\frac1{16}-\frac{\bar\eta_N}8\Bigr)-\sqrt{\bar\eta_N},
\]
which is nondecreasing in $N$ once $\bar\eta_N<1/2$ and is positive at $N_0\defeq3000$, where $\bar\eta_{N_0}<0.009$.
Hence the bracket is at least $\sqrt{\log N}/16$ for $N\ge N_0$.

\proofstep{Step 5 (small $N$).} For $2\le N<N_0$, use $\E\max_jX_j\ge\E\max(X_1,X_2)=\tfrac12\E|X_1-X_2|$. The
difference $\epsilon_{m1}-\epsilon_{m2}$ has the law of a sum of two signs, so $X_1-X_2$ has the law of
$k^{-1}S_{2k}$. Hölder's inequality with $\E S_{2k}^4\le3(\E S_{2k}^2)^2$ gives
\[
  \E|S_{2k}|\ge(\E S_{2k}^2)^{3/2}(\E S_{2k}^4)^{-1/2}\ge\sqrt{2k/3}.
\]
Hence $\E\max_jX_j\ge(6k)^{-1/2}\ge\sqrt{\log N/k}\,/\sqrt{6\log N_0}$.

Therefore $c_0=\min\{1/16,(6\log N_0)^{-1/2}\}=1/16$ works for both ranges.
\end{proof}

\begin{lemma}[Binomial tails by the method of types]\label[lemma]{lem:types}
For integers $0\le m\le k$,
\[
  \Pp\bigl(\mathrm{Bin}(k,1/2)\ge m\bigr)\ge\frac{1}{k+1}\exp\bigl\{-k\,D(m/k\,\|\,1/2)\bigr\},
\]
where $D(p\|q)$ is the binary relative entropy. Moreover, $D(\tfrac12+x\,\|\,\tfrac12)\le2x^2/(1-4x^2)$ for $|x|<1/2$.
\end{lemma}

\begin{proof}
Let $p=m/k$. The value $m$ is a mode of $\mathrm{Bin}(k,p)$: if $0<m<k$ then $(k+1)p\in(m,m+1)$, and if $m\in\{0,k\}$
the law is a point mass at $m$. Therefore $m$ has probability at least $1/(k+1)$ under $\mathrm{Bin}(k,p)$, and the likelihood
ratio at $m$ is
\[
  \frac{\Pp(\mathrm{Bin}(k,1/2)=m)}{\Pp(\mathrm{Bin}(k,p)=m)}=2^{-k}p^{-m}(1-p)^{m-k}=e^{-kD(p\|1/2)},
\]
with $0^0=1$ \citep[\S11.1]{cover2006elements}. Hence
\[
  \Pp\bigl(\mathrm{Bin}(k,1/2)\ge m\bigr)\ge\Pp\bigl(\mathrm{Bin}(k,1/2)=m\bigr)\ge\frac{e^{-kD(p\|1/2)}}{k+1}.
\]
For the second claim,
\[
  D(\tfrac12+x\,\|\,\tfrac12)=\sum_{j\ge1}\frac{(2x)^{2j}}{2j(2j-1)}\le\frac12\sum_{j\ge1}(2x)^{2j}=\frac{2x^2}{1-4x^2}.\qedhere
\]
\end{proof}

\begin{lemma}[Sharp maximal fluctuation]\label[lemma]{lem:radmax-sharp}
Let $X_j$ for $j\le N$ be as in \cref{lem:radmax}. If $k\to\infty$ and $N=N_k$ satisfy $\log k=o(\log N)$ and
$\log N=o(k)$, then for every $\eta\in(0,1)$,
\[
  \Pp\Bigl(\max_jX_j<(1-\eta)\sqrt{2\log N/k}\Bigr)\to0,\qquad\E\max_jX_j\ge(1-o(1))\sqrt{2\log N/k}.
\]
\end{lemma}

\begin{proof}
We follow the proof of \cref{lem:radmax}, with \cref{lem:types} in place of \cref{lem:revchernoff}.

\proofstep{Step 1 (one average).} Fix $\eta$, let $a=(1-\eta)\sqrt{2\log N/k}$, which tends to $0$, let
$m=\lceil k(1+a)/2\rceil$, and let $x=m/k-1/2\in[a/2,a/2+1/k]$. Since $X_1\ge a$ exactly when at least $k(1+a)/2$ of
its signs are positive, \cref{lem:types} gives
\[
  p\defeq\Pp(X_1\ge a)\ge\frac{1}{k+1}\exp\Bigl\{-\frac{2kx^2}{1-4x^2}\Bigr\},\qquad
  2kx^2\le\frac{ka^2}{2}+2a+\frac2k=(1-\eta)^2\log N+o(1).
\]

\proofstep{Step 2 (the maximum).} Because $x\to0$ and $\log(k+1)=o(\log N)$, Step 1 gives
$\log p\ge-(1-\eta)^2(1+o(1))\log N$, so
\[
  Np\ge N^{1-(1-\eta)^2(1+o(1))}\to\infty,\qquad\Pp\bigl(\max_jX_j<a\bigr)\le(1-p)^N\le e^{-Np}\to0 .
\]

\proofstep{Step 3 (the mean).} As in Step 3 of the proof of \cref{lem:radmax},
\[
  \E\max_jX_j\ge a\,\Pp\bigl(\max_jX_j\ge a\bigr)-k^{-1/2}\,\Pp\bigl(\max_jX_j<a\bigr)^{1/2}=a(1-o(1)),
\]
because $a\sqrt k\to\infty$. Since $\eta$ is arbitrary, the claim follows.
\end{proof}

\lowerthm*

\begin{proof}
We build a pipeline of capacity at most $\kappa$ whose gap is of order $\sqrt{\kappa/k}$, and find its exact constant in
the regime of (ii).

\emph{(i) The order.}
\proofstep{Step 1 (the construction).} Since $1\le\kappa\le k/16$, we have $k\ge16$. Let $N=\lfloor e^\kappa\rfloor\ge2$,
so that
\[
  (1-\log2)\,\kappa\le\kappa-\log2\le\log N\le\kappa\le k/16 .
\]
Let $\calZ=\{-1,+1\}^N$ carry the uniform law $P$ and let prompt $j$ have loss $\ell_j(z)=(1+z_j)/2\in\{0,1\}$, so every
prompt has risk $1/2$. For every $n\ge k$, let the pipeline read only $Z_1,\dots,Z_k$, select the indices $(1,\dots,k)$,
and output $T=(J,(1,\dots,k))$, where
\[
  J=\argmin_j\frac1k\sum_{m\le k}\ell_j(Z_m),
\]
with ties broken by the smallest index. The unused items $Z_{k+1},\dots,Z_n$ change neither the capacity nor the gap.

\proofstep{Step 2 (its capacity).} The transcript takes at most $N$ values, so \cref{lem:leakage}(i),(ii) give
$\MI(T;S)\le\Leak(S\to T)\le\log N\le\kappa$.

\proofstep{Step 3 (its gap).} The gap of the winner is half the maximum of $N$ independent Rademacher averages,
\[
  L(J)-\Lhat_S(J)=\frac12\max_j\frac1k\sum_{m\le k}(-Z_{m,j}),
\]
so \cref{lem:radmax} gives
\[
  \E\bigl[L(T)-\Lhat_S(T)\bigr]\ge\frac{c_0}2\sqrt{\frac{\log N}k}\ge\frac{c_0}{2}\sqrt{\frac{(1-\log2)\,\kappa}k},
\]
and $c=\sqrt{1-\log2}/32\approx0.017$ works.

\emph{(ii) The constant.} Because $\log k=o(\kappa)$, we have $\kappa\to\infty$ and $\log N\ge\log(e^\kappa-1)=\kappa-o(1)$.
Hence the hypotheses of \cref{lem:radmax-sharp} hold for the construction of (i), and the lemma, applied to the gap of
Step 3, gives
\[
  \E\bigl[L(T)-\Lhat_S(T)\bigr]\ge\frac12(1-o(1))\sqrt{\frac{2\log N}k}=(1-o(1))\sqrt{\frac{2\sigma^2\kappa}k},
  \qquad\sigma^2=\frac14,
\]
and the same lower bound for the gap itself with probability tending to one. Since $\Leak(S\to T)\le\log N\le\kappa$
and $\keff=k$, \cref{thm:transfer} bounds the expected gap by $\sqrt{2\sigma^2\kappa/k}$ and, at any fixed level
$\delta$, the gap by $\sqrt{2\sigma^2(\kappa+\log(1/\delta))/k}$. Both ratios to $\sqrt{2\sigma^2\kappa/k}$ tend to one.
\end{proof}

\begin{remark}[Exactness at every scale]\label[remark]{rem:exact-scale}
Suppose instead that $\kappa/k\to c\in(0,1/16]$, and define $a^\star$ by $D(\tfrac{1+a^\star}{2}\|\tfrac12)=c$. The
argument of \cref{lem:radmax-sharp}, run at a fixed $a<a^\star$ with the exact exponent $kD(m/k\,\|\,\tfrac12)$ of
\cref{lem:types} in place of its quadratic bound, gives an expected gap of at least $(1-o(1))\,a^\star/2$. This lower
bound matches the bound of \cref{prop:subgamma}'s argument run with the true ghost cumulant
$\lambda\mapsto k\log\cosh(\lambda/(2k))$ of the Rademacher gap, whose inverse conjugate at $\kappa$ tends to
$a^\star/2$. Hence the transfer inequality itself is asymptotically an equality, and at this scale the only loss in
\cref{thm:transfer}(i) is the sub-Gaussian proxy $\sigma^2=1/4$.
\end{remark}

\section{The capacity calculus}\label{app:proofs-calculus}

This appendix states the capacity calculus of \cref{sec:calculus} in full (\cref{app:calculus-formal}) and proves
its four rules (\cref{app:calculus-rules}). It then proves the lower bound for deterministic selection
(\cref{app:cherry}) and records how approximate max-information composes (\cref{app:maxinfo-rules}).

\subsection{Formal statement}\label{app:calculus-formal}

\begin{proposition}[Capacity calculus, formal version of \cref{prop:calculus}]\label[proposition]{prop:calculus-formal}
Let $T=(T_1,\dots,T_R)$ be produced in $R$ adaptive rounds given $U$.
\begin{enumerate}[label=(\alph*),leftmargin=2em]
\item \itemhead{Rounds add.} Let $\Leak_r(t_{<r})$ be the maximal leakage of round $r$ with the earlier rounds fixed at
$t_{<r}$, its essential suprema taken over $\Pstore_u$. Then
\[
  \MI(T;S\cond U)=\sum_{r=1}^R\MI(T_r;S\cond U,T_{<r}),\qquad
  \Leak(S\to T\cond U=u)\le\sum_{r=1}^R\sup_{t_{<r}}\Leak_r(t_{<r}).
\]
\item \itemhead{Post-processing is free.} If $T'$ is generated from $(T,U)$ without reading $S$, then
\[
  \MI(T';S\cond U)\le\MI(T;S\cond U),\qquad\Leak(S\to T'\cond U)\le\Leak(S\to T\cond U).
\]
\item \itemhead{Base charges.} A round that outputs one of $M$ fixed candidates costs at most $\log M$ in either
capacity. A round that releases $a(S)+\xi$ with $\xi\sim\mathcal N(0,\rho^{-1}I)$ costs at most
$\tfrac\rho2\E\|a(S)-a(S^\circ)\|_2^2$ in mutual information, with $S^\circ$ an independent copy of $S$, and at most $\rho\Delta^2/2$ when
$\|a(s)-a(s')\|_2\le\Delta$ for all stores $s,s'$, whatever filters follow. A
round that samples from a law $Q_S$ costs at most $\E\,\KL(Q_S\,\|\,\Pi)$ in mutual information for any law $\Pi$ fixed in advance. If
$Q_S$ is supported on a data-dependent gate $A(S)$, then
$\KL(Q_S\|\Pi)=\KL(Q_S\|\Pi_{A(S)})+\log\bigl(1/\Pi(A(S))\bigr)$, where $\Pi_A$ is $\Pi$ conditioned on $A$. This charge
prices the sampled transcript, whereas a logged gate is a separate release. A
score on a split that no round has read costs $0$.
\item \itemhead{Side information splits the price.} Let a latent task $W$ be such that, given $(U,W)$, the store
satisfies \cref{ass:store} with item law $P_{U,W}$ and the deployment item is $Z_\star\sim P_{U,W}$, independent of $S$. Let the
pipeline read $W$ only through $(U,S)$, that is, $T\perp W\mid(U,S)$. Then
$\MI(T;S\cond U)=\MI(T;W\cond U)+\MI(T;S\cond U,W)$, and \cref{thm:transfer} holds conditionally on $(U,W)$ with the
within-task capacity $\MI(T;S\cond U,W)$ for the within-task risk.
\end{enumerate}
\end{proposition}

\subsection{The four rules}\label{app:calculus-rules}

We prove the rules of \cref{prop:calculus-formal} in turn. All statements are conditional on $U=u$, and we drop $u$
from the notation. The unconditional versions follow by averaging.

\paragraph{(a) Rounds add.}
The chain rule for mutual information \citep{cover2006elements}, applied $R-1$ times, gives
\[
  \MI(T_1,\dots,T_R;S)=\sum_{r=1}^R\MI(T_r;S\cond T_{<r}).
\]
For maximal leakage, it suffices to treat two rounds and induct. Write the density of $(T_1,T_2)$ given $s$ as
$k_1(t_1\mid s)\,k_2(t_2\mid s,t_1)$. Since
\[
  \esssup_sk_1(t_1\mid s)\,k_2(t_2\mid s,t_1)\le\esssup_sk_1(t_1\mid s)\cdot\esssup_sk_2(t_2\mid s,t_1),
\]
integrating over $t_2$ and then over $t_1$ gives
\[
  \Leak\bigl(S\to(T_1,T_2)\bigr)\le\Leak(S\to T_1)+\sup_{t_1}\Leak_2(t_1),\qquad
  \Leak_2(t_1)=\log\int\esssup_sk_2(t_2\mid s,t_1)\,\mu_2(\dd t_2),
\]
where $\mu_2$ is the reference measure of the second round and $\Leak_2(t_1)$ is the leakage of that round with its
history fixed. The essential suprema run over the store law throughout, so no conditioning on the history is needed.
Induction over rounds proves (a) \citep{issa2020operational,esposito2021generalization}.

\paragraph{(b) Post-processing is free.}
Let $T'=f(T,R)$ with randomness $R$ independent of $(S,T)$, and let $G(\cdot\mid t)$ be the law of $f(t,R)$. Mutual
information does not increase under data processing \citep{cover2006elements}. For maximal leakage, assume
$\Leak(S\to T)<\infty$ and, with $k$ and $\mu$ as in \cref{def:laws}, define the law
\[
  R^\ast(\dd t)=e^{-\Leak(S\to T)}\esssup_sk(t\mid s)\,\mu(\dd t)
\]
on $\calT$. For almost every $s$ and every event $A$, the kernel $K'$ of $T'$ satisfies
\[
  K'(A\mid s)=\int G(A\mid t)\,k(t\mid s)\,\mu(\dd t)\le e^{\Leak(S\to T)}\,(R^\ast G)(A),
\]
so $K'(\cdot\mid s)$ has a density at most $e^{\Leak(S\to T)}$ with respect to the law $R^\ast G$ obtained by passing
$R^\ast$ through $G$, and
$\Leak(S\to T')\le\Leak(S\to T)$. This is the data-processing property of \cref{lem:leakage}(iii), proved here in the
form that we use.

\paragraph{(c) Base charges.}
\emph{Menus.} A round with at most $M$ outputs has, for every history,
\[
  \MI(T_r;S\cond T_{<r})\le H(T_r\cond T_{<r})\le\log M,\qquad
  \log\sum_y\esssup_s\Pp(T_r=y\mid S=s)\le\log M,
\]
where the second quantity is its maximal leakage with the counting measure (\cref{lem:leakage}(ii)).

\emph{Gaussian releases.} Let $Y=a(S)+\xi$ with $\xi\sim\mathcal N(0,\rho^{-1}I_m)$ independent of $S$, write $P_s$ for
the law $\mathcal N(a(s),\rho^{-1}I_m)$, and let $P_Y=\E_{S^\circ}P_{S^\circ}$. Convexity of $\KL(P_s\|\cdot)$ gives
\[
  \KL(P_s\|P_Y)\le\E_{S^\circ}\KL(P_s\|P_{S^\circ})=\frac\rho2\,\E_{S^\circ}\|a(s)-a(S^\circ)\|_2^2,
\]
and averaging over $S$ bounds $\MI(Y;S)$ as claimed. Any filter applied to $Y$ is post-processing, so
(b) covers it.

\emph{Gibbs rounds and gates.} For a round that samples from $Q_S$ with marginal $\bar Q=\E_SQ_S$ and any law $\Pi$
fixed in advance, the golden formula (\cref{lem:golden}) reads
\[
  \E_S\KL(Q_S\|\Pi)=\E_S\KL(Q_S\|\bar Q)+\KL(\bar Q\|\Pi)\ge\E_S\KL(Q_S\|\bar Q)=\MI(T_r;S).
\]
If $Q$ is supported on $A$ and $Q\ll\Pi_A=\Pi(\cdot\cap A)/\Pi(A)$, then $\dd\Pi_A/\dd\Pi=\mathbf 1_A/\Pi(A)$, which
gives the gate identity
\[
  \KL(Q\|\Pi)=\E_Q\log\frac{\dd Q}{\dd\Pi_A}+\E_Q\log\frac{\dd\Pi_A}{\dd\Pi}=\KL(Q\|\Pi_A)+\log\frac1{\Pi(A)}.
\]

\emph{Gibbs rounds in realized-gap capacity.} With $\pi$ a density of $\Pi$, a Gibbs round
$q_s(t)\propto\pi(t)e^{-\lambda\Lhat_s(t)}$ on scores in $[0,1]$ has normalizer at least $e^{-\lambda}$, so
$q_s(t)\le e^\lambda\pi(t)$ and its maximal leakage is at most $\min\{\lambda,\log|\supp\Pi|\}$, the bound of the
exponential mechanism. With a data-dependent gate $A(s)$, the
normalizer is at least $e^{-\lambda}\Pi(A(s))$, so the bound becomes
\[
  \min\Bigl\{\lambda+\esssup_s\log\frac1{\Pi(A(s))},\ \log|\supp\Pi|\Bigr\};
\]
a singleton gate at the winner leaks $\log|\supp\Pi|$ at every $\lambda$. At the inverse temperatures that SCAR uses,
this bound is no better than the menu bound, so \cref{thm:scar} prices SCAR by its logged divergence instead.

\emph{Fresh splits.} If a score is computed on a split $S'$ that is independent of $(S,T)$ and read by no round, then
$\MI(T;S'\cond U)=0$, and \cref{thm:transfer} for the score on $S'$ reduces to ghost concentration.

\paragraph{(d) Side information.}
The chain rule in two orders gives
\[
  \MI(T;S,W\cond U)=\MI(T;S\cond U)+\MI(T;W\cond U,S)=\MI(T;W\cond U)+\MI(T;S\cond U,W).
\]
Because $T\perp W\mid(U,S)$, the term $\MI(T;W\cond U,S)$ vanishes, and equating the two expansions gives the
identity. Conditional on $(U,W)=(u,w)$, the store satisfies \cref{ass:store} with item law $P_{u,w}$ and
$Z_\star\sim P_{u,w}$. Hence \cref{thm:transfer} applies with $U$ replaced by $(U,W)$, and it bounds the gap to the
within-task risk $L_{u,w}(T)$. We apply the theorem given $(U,W)$ because, given $U$ alone, the store is a mixture
over tasks and need not be i.i.d.

\subsection{Deterministic selection}\label{app:cherry}

Deterministic retrieval that places raw store text in the transcript lies outside the calculus. For such retrieval, a
surcharge that depends only on $(n,k,\delta)$ must cover nearly the whole loss range.

\cherryremark*

\begin{proof}
The number $B$ of zero-loss items is $\mathrm{Bin}(n,1-p)$ with mean $\mu=n(1-p)\ge2k$, so \cref{lem:chernoff}
gives
\[
  \Pp(B<k)\le\Pp(B\le\mu/2)\le e^{-\mu/8}\le e^{-k/4}.
\]
On $\{B\ge k\}$, the selected score is $0$ and the risk is $p$. If a surcharge $b$ satisfied $b<p$, the certified event
$\{L\le\Lhat_S+b\}$ would be contained in $\{B<k\}$, whose probability is at most $e^{-k/4}<1-\delta$. Hence $b\ge p$
for every $p\le1-2k/n$.
\end{proof}

\subsection{Approximate max-information under composition}\label{app:maxinfo-rules}

For the variant of \cref{rem:maxinfo}, the three rules below hold with additive slacks, and composition across rounds
needs a stronger hypothesis than for maximal leakage.

\emph{Menus.} Let $T$ take at most $M$ values, let $q_y=\Pp(T=y)$, and call $y$ light if $q_y<\beta/M$ and heavy
otherwise. Light outcomes have total probability below $\beta$. For an event $E$ with sections $E_y$ and for a heavy $y$,
\[
  \Pp(S\in E_y,\,T=y)\le\Pp(S\in E_y)\le\frac M\beta\,q_y\,\Pp(S^\circ\in E_y),
  \qquad\text{so}\qquad Q(E)\le\frac M\beta\,\Qbar(E)+\beta
\]
\citep[Thm.~6]{dwork2015generalization}. The slack is necessary: for $T=\mathbf 1\{S\in A\}$ with $\Pp(S\in A)=p$, the
event $\{S\in A,T=1\}$ has probability $p$ under the real law and $p^2$ under the ghost law, although the maximal
leakage of $T$ is $\log2$.

\emph{Post-processing.} Let $T'=f(T,R)$ as in rule (b). For an event $E'$ of $(s,t')$ and each value $r$ of $R$, the set
$E_r=\{(s,t):(s,f(t,r))\in E'\}$ is an event of $(s,t)$, so
\[
  \Pp\bigl((S,T')\in E'\bigr)=\E_RQ(E_R)\le e^\kappa\,\Pp\bigl((S^\circ,T')\in E'\bigr)+\beta .
\]

\emph{Rounds.} Composition requires that we measure each round against the conditional law of the store given the
history. Suppose that for every event and almost every history,
\[
  P_{ST_r\mid t_{<r}}(E)\le e^{\kappa_r}\bigl(P_{S\mid t_{<r}}\otimes P_{T_r\mid t_{<r}}\bigr)(E)+\beta_r .
\]
Write $P_{ST_1}=m_1+e_1$ with $m_1=\min\{P_{ST_1},e^{\kappa_1}P_S\otimes P_{T_1}\}$, and split the second-round kernel
as $K_2(\cdot\mid s,t_1)=m_2+e_2$ with $m_2=\min\{K_2(\cdot\mid s,t_1),e^{\kappa_2}P_{T_2\mid t_1}\}$. Then
\[
  P_{ST_1T_2}\le m_1\otimes m_2+e_1\otimes K_2+P_{ST_1}\otimes e_2,
\]
with $m_1\otimes m_2\le e^{\kappa_1+\kappa_2}P_S\otimes P_{T_1T_2}$ and the other two terms of mass at most $\beta_1$ and
$\beta_2$. Induction gives slack $\sum_r\beta_r$ \citep[Lemma~5]{dwork2015generalization}.

\emph{Why the conditional law is needed.} The conditional law cannot be replaced by the store law. For a one-item
store uniform on $\{1,\dots,8\}$, let $T_1=\mathbf 1\{S\le4\}$ and let the second round output
$T_2=t_1\mathbf 1\{S\in\{1,5,6,7,8\}\}$ given $T_1=t_1$. Against the store law, $T_1$ has max-information $\log2$, and the
second round has max-information at most $\log(8/3)$ for each fixed $t_1$, whereas $(T_1,T_2)$ has $\log8$. Maximal leakage
avoids this subtlety because it depends on the store law only through its null sets.

\section{Stability-constrained adaptive randomization}\label{app:proofs-scar}

This appendix proves the SCAR certificate of \cref{sec:scar} and its oracle inequality (\cref{app:scar}). It then shows
that SCAR is a minimax-rate selector (\cref{app:scar-optimal}) and proves the second-order certificate
(\cref{app:second}). We restate results of the main text before their proofs.

\subsection{The certificate and its oracle inequality}\label{app:scar}

\scarthm*

\begin{proof}
Both inequalities hold on the event of \cref{thm:transfer}(iii).

\proofstep{Step 1 (the certificate).} Fix $u$ and work on the event of probability at least $1-\delta$ on which
\cref{thm:transfer}(iii) holds for every $Q\ll\Pi_u$. The gate $A=A(S,u)$ and the posterior $Q_\lambda$ depend on $S$.
Because (iii) holds simultaneously for all such laws, the first inequality is (iii) at $Q=Q_\lambda$.

\proofstep{Step 2 (the oracle inequality).} On laws with $Q(A)=1$, let
\[
  F(Q)=\E_Q\Lhat_S+\lambda^{-1}\KL(Q\|\Pi_u)=\E_Q\Lhat_S+\lambda^{-1}\KL(Q\|\Pi_{u,A})+\lambda^{-1}\log\frac1{\Pi_u(A)},
\]
where the second form is the gate identity of \cref{app:calculus-rules}. The Gibbs variational principle (\cref{lem:dv}(ii))
shows that $Q\mapsto\E_Q\Lhat_S+\lambda^{-1}\KL(Q\|\Pi_{u,A})$ is minimized by
$Q\propto\Pi_{u,A}e^{-\lambda\Lhat_S}$, which is $Q_\lambda$. Hence $F(Q_\lambda)\le F(Q)$ for every $Q$ supported on the
gate. After adding $\lambda^{-1}\log(1/\delta)+\lambda\sigma^2/(2\keff)$, this is the second inequality.
\end{proof}

\subsection{Rate optimality}\label{app:scar-optimal}

\begin{proposition}[Rate optimality of SCAR]\label[proposition]{prop:scar-optimal}
Let \cref{ass:store}(a) hold with $\ell_t\in[0,1]$ and uniform weights over $k$ distinct selected examples, let the
library have $M\ge2$ prompt recipes scored on the same $k$ reused examples, and let $L^\star=\min_tL_u(t)$.
\begin{enumerate}[label=(\roman*),leftmargin=2em]
\item \itemhead{Upper bound.} Take the uniform prior, no gate, $\sigma^2=\tfrac14$, and the inverse temperature
\[
  \lambda^\star=\sqrt{8k\bigl(\log M+\log(1/\delta)\bigr)}.
\]
With probability at least $1-2\delta$, SCAR at $\lambda=\lambda^\star$ satisfies
\[
  \E_{Q_\lambda}L_u\le L^\star+\sqrt{2(\log M+\log(1/\delta))/k}.
\]
\item \itemhead{Lower bound.} If $4\le M\le e^{2k/3}$, there are laws $P_1,\dots,P_M$ such that every selection rule
that maps the $k$ reused examples to a law $Q$ on the library has
\[
  \max_j\E_j\bigl[\E_QL_j-L_j^\star\bigr]\ge\frac14\sqrt{\frac{3\log M}{32k}}\ge0.07\sqrt{\frac{\log M}k}.
\]
\end{enumerate}
\end{proposition}

\begin{proof}
Part (i) uses the oracle inequality at the best recipe, and part (ii) reduces selection to identifying the best of
$M$ options.

\emph{(i) Upper bound.} Let $t^\star$ attain $L^\star$. On the event of \cref{thm:scar}, the oracle inequality at the point
mass $Q=\delta_{t^\star}$, for which $\KL(\delta_{t^\star}\|\Pi_u)=\log M$, gives
\[
  \E_{Q_\lambda}L_u\le\Lhat_S(t^\star)+\frac{\log M+\log(1/\delta)}{\lambda^\star}+\frac{\lambda^\star}{8k}
  =\Lhat_S(t^\star)+\sqrt{\frac{\log M+\log(1/\delta)}{2k}}.
\]
Hoeffding's inequality for the fixed recipe $t^\star$ gives $\Lhat_S(t^\star)\le L^\star+\sqrt{\log(1/\delta)/(2k)}$ with
probability at least $1-\delta$. The two surcharges sum to at most $\sqrt{2(\log M+\log(1/\delta))/k}$.

\emph{(ii) Lower bound.}
\proofstep{Step 1 (a hard family).} Let $\calZ=\{0,1\}^M$ and let recipe $t$ have loss $\ell_t(z)=z_t$. Under $P_j$, let
the coordinates be independent, with coordinate $j$ Bernoulli$(\tfrac12-\epsilon)$ and the others Bernoulli$(\tfrac12)$,
where $\epsilon^2=3\log M/(32k)\le1/16$. Then $L_j^\star=\tfrac12-\epsilon$ and $\E_QL_j-L_j^\star=\epsilon(1-Q(j))$.

\proofstep{Step 2 (reduction to identification).} Draw $J$ uniformly on $[M]$, the examples from $P_J^{\otimes k}$, and
$\hat J$ from $Q$. Then the average excess risk is $\epsilon\,\Pp(\hat J\ne J)$.

\proofstep{Step 3 (information).} By the golden formula (\cref{lem:golden}) with the law $P_0$ of $M$ fair coordinates as
reference, and by \cref{lem:types},
\[
  \MI(J;\text{examples})\le\max_j\KL\bigl(P_j^{\otimes k}\,\|\,P_0^{\otimes k}\bigr)=kD(\tfrac12-\epsilon\,\|\,\tfrac12)
  \le\frac{2k\epsilon^2}{1-4\epsilon^2}\le\frac83\,k\epsilon^2=\frac14\log M .
\]

\proofstep{Step 4 (Fano).} Fano's inequality (\cref{lem:fano}) gives, for $M\ge4$,
\[
  \Pp(\hat J\ne J)\ge1-\frac{\tfrac14\log M+\log2}{\log M}\ge\frac14,
\]
so some $j$ has expected excess risk at least $\epsilon/4$.
\end{proof}

The upper bound is the certificate of \cref{thm:scar}, converted into a risk guarantee, and the lower bound is the
classical sample complexity of identifying the best of $M$ options. For fixed $\delta$ and $4\le M\le e^{2k/3}$, the two
bounds match up to constants. Hence the certified selector is a minimax-rate selector up to a constant factor, and
certification does not worsen the rate.

\subsection{The second-order certificate}\label{app:second}

The certificate replaces the range of the losses by a predictable variance. Its proof uses a supermartingale built
from the inequality of \cref{lem:fan}.

\begin{lemma}[Predictable plug-in supermartingale]\label[lemma]{lem:plugin}
Let $X_1,\dots,X_k\in[0,1]$ be independent with common mean $\mu$, and let each $\muhat_{j-1}\in[0,1]$ be a function
of $X_1,\dots,X_{j-1}$. For every $\gamma\in[0,1)$,
\[
  \E\exp\Bigl\{\gamma\sum_{j=1}^k(\mu-X_j)-\psi(\gamma)\sum_{j=1}^k(X_j-\muhat_{j-1})^2\Bigr\}\le1 .
\]
\end{lemma}

\begin{proof}
Let $\mathcal F_{j-1}=\sigma(X_1,\dots,X_{j-1})$ and $\xi_j=\muhat_{j-1}-X_j\in[-1,1]$. Then
$\gamma(\mu-X_j)=\gamma(\mu-\muhat_{j-1})+\gamma\xi_j$ and $(X_j-\muhat_{j-1})^2=\xi_j^2$, so by \cref{lem:fan}
\[
  \E\bigl[e^{\gamma(\mu-X_j)-\psi(\gamma)(X_j-\muhat_{j-1})^2}\bigm|\mathcal F_{j-1}\bigr]
  \le e^{\gamma(\mu-\muhat_{j-1})}\bigl(1+\gamma(\muhat_{j-1}-\mu)\bigr)=e^{-a}(1+a)\le1,
\]
with $a=\gamma(\muhat_{j-1}-\mu)$. The tower property over $j=k,k-1,\dots,1$ proves the claim. This is the
predictable plug-in construction of \citet[eq.~(13)]{waudbysmith2024estimating} in the mirrored form $\mu-X_j$.
\end{proof}

\secondthm*

\begin{proof}
The proof follows that of \cref{thm:transfer}(iii), with \cref{lem:plugin} in place of \cref{lem:ghost}.

\proofstep{Step 1 (one transcript).} Fix $u$ and $\gamma\in\Gamma$. For a fixed $t$, the indices are distinct constants,
so under $S\sim P_u^n$ the selected losses $X_j(t)$ are i.i.d.\ with mean $L_u(t)$. Moreover,
$\muhat_{j-1}(t)\in[0,1]$ depends only on $X_1(t),\dots,X_{j-1}(t)$. \Cref{lem:plugin} gives $\E_SM_\gamma(t,S)\le1$ for
\[
  M_\gamma(t,S)=\exp\bigl\{\gamma k\bigl(L_u(t)-\Lhat_S(t)\bigr)-\psi(\gamma)\,k\,\Vhat_S(t)\bigr\}.
\]

\proofstep{Step 2 (all transcripts and all grid points).} By Fubini's theorem, $\E_S\E_{\Pi_u}M_\gamma\le1$, so
$\Pp(\E_{\Pi_u}M_\gamma>|\Gamma|/\delta)\le\delta/|\Gamma|$. A union bound over the grid gives an event of probability
at least $1-\delta$ on which $\E_{\Pi_u}M_\gamma\le|\Gamma|/\delta$ for every $\gamma\in\Gamma$.

\proofstep{Step 3 (change of measure).} On that event, \cref{lem:dv}(i) gives, for every $Q\ll\Pi_u$ and every
$\gamma\in\Gamma$,
\[
  \gamma k\bigl(\E_QL_u-\E_Q\Lhat_S\bigr)-\psi(\gamma)\,k\,\E_Q\Vhat_S\le\KL(Q\|\Pi_u)+\log\frac{|\Gamma|}\delta,
\]
and dividing by $\gamma k$ proves the bound.

\proofstep{Step 4 (the minimizer).} For fixed $\gamma$, the right-hand side equals, up to a constant,
\[
  \E_Q\Bigl[\Lhat_S+\frac{\psi(\gamma)}{\gamma}\,\Vhat_S\Bigr]+\frac{\KL(Q\|\Pi_u)}{\gamma k},
\]
which the Gibbs variational principle (\cref{lem:dv}(ii)) minimizes at $Q\propto\Pi_u\exp\{-k(\gamma\Lhat_S+\psi(\gamma)\Vhat_S)\}$.
\end{proof}

\begin{remark}[Reading the certificate]\label[remark]{rem:bernstein-form}
Since $\psi(\gamma)/\gamma=\sum_{m\ge2}\gamma^{m-1}/m\le\gamma/(2(1-\gamma))$, the surcharge at $\gamma$ is at most
\[
  \frac{\gamma\,\E_Q\Vhat_S}{2(1-\gamma)}+\frac{C}{\gamma k},\qquad C=\KL(Q\|\Pi_u)+\log\frac{|\Gamma|}\delta .
\]
At $\gamma^\circ=\sqrt{2C/(k\E_Q\Vhat_S)}$, this bound is $(1+O(\gamma^\circ))\sqrt{2\E_Q\Vhat_S\,C/k}$. The dyadic grid of
\cref{sec:scar} contains a point within a factor $2$ of $\gamma^\circ$ whenever
$\gamma^\circ\in[0.9\cdot2^{-\lceil\log_2k\rceil},0.9]$. Using that point costs at most a factor $5/4$ in the leading term. The
predictable variance satisfies
\[
  \E\Vhat_S(t)=v\Bigl(1+\frac{H_k-H_k^{(2)}}k\Bigr)+\Bigl(\frac12-L_u(t)\Bigr)^2\frac{H_k^{(2)}}k,
\]
with $v=\Var\,\ell_t(Z)$, $H_k=\sum_{j\le k}j^{-1}$, and $H_k^{(2)}=\sum_{j\le k}j^{-2}\le\pi^2/6$, that is,
$v(1+O(\log k/k))+O(1/k)$. The last term reflects the initial guess $\muhat_0=1/2$. The certificate reported in
\cref{sec:experiments} is the exact minimum over $\Gamma$.
\end{remark}

\section{Bayesian linear in-context regression}\label{app:linear}
\label{sec:linear}
Bayesian linear in-context regression makes exact the separation between retrieval quality and certification,
stated in \cref{sec:discussion}. Let $\theta\sim\mathcal N(0,\tau^2I_d)$ and
$x_i\sim\mathcal N(0,I_d)$, and let $y_i=x_i^\top\theta+\xi_i$ with $\xi_i\sim\mathcal N(0,\sigma_\xi^2)$. Consider a query
$x_\star$ with label $y_\star$ and direction $u=x_\star/\|x_\star\|$. Given $k$ selected examples $(X_I,y_I)$, the Bayes
predictor $\widehat y_\star$ is the posterior mean, which is ridge regression with penalty $\sigma_\xi^2/\tau^2$.

\begin{proposition}[Retrieval improves the design]\label[proposition]{prop:linear}
For every fixed selected design $X_I$,
\[
  \E[(y_\star-\widehat y_\star)^2\mid X_I,x_\star]=\sigma_\xi^2+x_\star^\top\bigl(\tau^{-2}I_d+\sigma_\xi^{-2}X_I^\top X_I\bigr)^{-1}x_\star .
\]
If $k^{-1}X_I^\top X_I\succeq\alpha\,uu^\top$, the excess risk is at most $\|x_\star\|^2/(\tau^{-2}+k\alpha/\sigma_\xi^2)$.
\end{proposition}

\begin{proof}
Condition on $X_I$. The Gaussian prior and likelihood give the posterior covariance
\[
  \Sigma_I=\bigl(\tau^{-2}I_d+\sigma_\xi^{-2}X_I^\top X_I\bigr)^{-1},
\]
which does not depend on $y_I$. The prediction error of
$y_\star=x_\star^\top\theta+\xi_\star$ is the noise plus the posterior variance, $\sigma_\xi^2+x_\star^\top\Sigma_Ix_\star$. If
$k^{-1}X_I^\top X_I\succeq\alpha uu^\top$, then
\[
  \Sigma_I^{-1}\succeq B\defeq\tau^{-2}I_d+\frac{k\alpha}{\sigma_\xi^2}\,uu^\top,
\]
so $\Sigma_I\preceq B^{-1}$, and $u$ is an eigenvector of $B$ with eigenvalue $\tau^{-2}+k\alpha/\sigma_\xi^2$. Multiplying
$u^\top B^{-1}u$ by $\|x_\star\|^2$ gives the bound.
\end{proof}

Query-aligned retrieval lowers the risk by raising the directional mass $\alpha_I=u^\top(k^{-1}X_I^\top X_I)u$ of the
design along $u$. A Schur complement expresses the risk exactly through $\alpha_I$ and measures how much of the gain leaks into
nuisance directions (\cref{prop:schur}). Top-$k$ retrieval by $(u^\top x_i)^2$ attains
$\alpha_I\ge F_{\chi^2_1}^{-1}(1-2k/n)$, a chi-squared quantile, with probability at least $1-e^{-k/4}$
(\cref{cor:topk}). These gains do not change certification. If the retrieval rule, the similarity metric, or $k$ is
chosen on the same data that scores the result, the reported score still pays the capacity of \cref{thm:transfer}.

\begin{proposition}[Schur-complement form of the query precision]\label[proposition]{prop:schur}
Let $A_I=k^{-1}X_I^\top X_I$, with directional mass $\alpha_I=u^\top A_Iu$, cross term $b_I=(I_d-uu^\top)A_Iu$, and
nuisance block $C_I=(I_d-uu^\top)A_I(I_d-uu^\top)$, and define the effective directional precision
\[
  \varpi_I\defeq\tau^{-2}+\frac{k}{\sigma_\xi^2}\alpha_I-\frac{k^2}{\sigma_\xi^4}\,
  b_I^\top\Bigl(\tau^{-2}(I_d-uu^\top)+\frac{k}{\sigma_\xi^2}C_I\Bigr)^{\dagger}b_I .
\]
Then $\E[(y_\star-\widehat y_\star)^2\mid X_I,x_\star]=\sigma_\xi^2+\|x_\star\|^2/\varpi_I$. The subtracted term is the precision
lost to nuisance directions that the selected examples do not orthogonalize.
\end{proposition}

\begin{proof}
In the decomposition $\R^d=\mathrm{span}\{u\}\oplus u^\perp$, the precision $\Sigma_I^{-1}=\tau^{-2}I_d+(k/\sigma_\xi^2)A_I$
has the block form
\[
  \Sigma_I^{-1}=\begin{pmatrix}\tau^{-2}+k\alpha_I/\sigma_\xi^2 & k\,b_I^\top/\sigma_\xi^2\\[2pt]
  k\,b_I/\sigma_\xi^2 & \tau^{-2}(I_d-uu^\top)+k\,C_I/\sigma_\xi^2\end{pmatrix}.
\]
The $u^\perp$ block is invertible on $u^\perp$, with inverse given by the pseudo-inverse. The $(u,u)$ entry of
$\Sigma_I$ is the inverse of the Schur complement of the $u^\perp$ block, which is $\varpi_I$. Finally,
$x_\star^\top\Sigma_Ix_\star=\|x_\star\|^2u^\top\Sigma_Iu$, and \cref{prop:linear} completes the proof.
\end{proof}

\begin{corollary}[Top-$k$ Gaussian retrieval]\label[corollary]{cor:topk}
Let $x_i\sim\mathcal N(0,I_d)$ be i.i.d.\ and independent of $x_\star$, and let the retriever select the $k$ examples
with the largest $(u^\top x_i)^2$, with $2k<n$. Let $q_{n,k}=F_{\chi^2_1}^{-1}(1-2k/n)$. With probability at least
$1-e^{-k/4}$, the directional mass satisfies $\alpha_I\ge q_{n,k}$. Hence, on that event, if the subtracted term of
\cref{prop:schur} is at most a fraction $\eta\in[0,1)$ of $\tau^{-2}+kq_{n,k}/\sigma_\xi^2$, then
\[
  \E[(y_\star-\widehat y_\star)^2\mid X_I,x_\star]\le\sigma_\xi^2+\frac{\|x_\star\|^2}{(1-\eta)(\tau^{-2}+kq_{n,k}/\sigma_\xi^2)}.
\]
\end{corollary}

\begin{proof}
Selection uses only the inputs, so it suffices to bound the directional mass.

\proofstep{Step 1 (selection does not bias the regression).} The selected set $I$ is a function of
$(x_1,\dots,x_n,x_\star)$ only. The parameter $\theta$ and the noises are independent of these variables, so
$y_I=X_I\theta+\xi_I$ with $\xi_I\sim\mathcal N(0,\sigma_\xi^2I_k)$ independent of $\theta$. \Cref{prop:linear,prop:schur}
therefore apply conditionally on $X_I$.

\proofstep{Step 2 (the directional mass).} The scores $(u^\top x_i)^2$ are i.i.d.\ $\chi^2_1$, so the number of them at
least $q_{n,k}$ is $\mathrm{Bin}(n,2k/n)$ with mean $2k$. Hence \cref{lem:chernoff} bounds by
$e^{-k/4}$ the probability that fewer than $k$ scores exceed $q_{n,k}$. Otherwise the top $k$ scores all exceed $q_{n,k}$, and so does their
average $\alpha_I$.

\proofstep{Step 3 (the risk).} The risk bound follows from \cref{prop:schur}, since
$\varpi_I\ge(1-\eta)(\tau^{-2}+kq_{n,k}/\sigma_\xi^2)$.
\end{proof}

\needspace{6\baselineskip}%
\section{Experimental protocols and additional results}\label{app:experiments}

\subsection{The Banking77 pipeline}\label{app:banking}

\paragraph{Data.}
Banking77 \citep{casanueva2020efficient} (CC BY 4.0) contains online-banking queries labeled with 77 intents. With
seed 20260727, we draw a demonstration bank of 2{,}000 labeled queries from the training split. From the test split, we draw a
validation pool of 256 queries and a disjoint holdout of 1{,}024 queries. The GPT-4.1-mini replication uses the first
512 holdout queries. The demonstration bank supplies only demonstrations and is fixed before any score is read, so it is public side
information in the sense of \cref{sec:setup}. The certified statistic, the selected score of \cref{def:transcript}, is
the error on $k$ queries drawn from the validation pool.

\paragraph{Recipes.}
A recipe fixes an instruction template, a retrieval rule, and a demonstration configuration (\cref{sec:experiments}).
We embed the demonstration bank and the queries with bge-small-en-v1.5
\citep{xiao2024cpack} (MIT license), normalize the embeddings, and compare them by cosine similarity. The four retrieval
rules select $m$ demonstrations per query: uniformly at random (seeded per recipe and query); the $m$ most similar;
greedily in similarity order skipping labels already chosen (label-diverse); and maximal marginal relevance with weight
$0.5$ over the 50 most similar. The four demonstration configurations are $m\in\{4,8\}$ with the most similar
demonstration first or last. The four instruction templates are:
\begin{itemize}[leftmargin=2em]
  \item[T0] \emph{You are an intent classifier for a banking assistant. Classify the customer query into exactly one intent label from the list below. Answer with the label only.}
  \item[T1] \emph{Task: banking intent detection. Read the customer message and output the single best-matching intent label from the allowed list. Output only the label, nothing else.}
  \item[T2] \emph{As a precise banking support router, choose the one intent label that best describes the user query. The label must be copied verbatim from the allowed labels. Respond with only that label.}
  \item[T3] \emph{Classify the following online-banking customer query. Pick exactly one intent from the allowed intent labels and reply with just that intent label in lowercase.}
\end{itemize}
The system message is the template followed by the list of the 77 labels. The user message lists the demonstrations as
\texttt{Query:}/\texttt{Intent:} pairs followed by the test query. The four zero-shot baselines use the templates without
demonstrations.

\paragraph{Models and parsing.}
Qwen2.5-7B-Instruct \citep{qwen2025qwen25} (Apache 2.0) runs in vLLM \citep{kwon2023efficient} in bf16 at temperature $0$
with at most 12 new tokens. GPT-4.1-mini runs through the OpenAI chat-completions API at temperature $0$ with a fixed
seed and at most 12 tokens. We lower-case the first line of the output and map non-alphanumeric characters to
underscores. We accept the result if it equals a label or matches exactly one label as a substring, and otherwise
count it as an error (0.73\% and 0.10\% of outputs).

\paragraph{Selection, certificates, and SCAR.}
We run 400 repetitions in each of 30 cells, one per menu size $M\in\{2,4,8,16,32,64\}$ and scoring sample size
$k\in\{16,32,64,128,256\}$. Each repetition draws $k$ validation queries (all 256 when $k=256$) and $M$ of the 64 recipes
(all when $M=64$). It selects the recipe with the lowest validation error, breaking ties at random, and records its gap,
the holdout error minus the validation error. The certificate of \cref{thm:transfer}(i) is
$\sqrt{\log M/(2k)}$, and at level $\delta=0.05$ the certificates of (ii) and of the approximate max-information variant
of \cref{rem:maxinfo} are
\[
  \sqrt{\frac{\log M+\log(1/\delta)}{2k}}\qquad\text{and}\qquad
  \sqrt{\frac{\log(M/\beta)+\log(1/\delta')}{2k}},\quad\beta=\delta'=0.025 .
\]
SCAR (\cref{alg:scar}) samples a recipe from a Gibbs posterior with the uniform prior over the 64 recipes and the
inverse temperature, fixed in advance,
\[
  \lambda^\star=\sqrt{8k(\log64+\log20)} ;
\]
for each $k$, we average the certificate of \cref{thm:scar} over 50 validation subsets. We evaluate the best-of-64
winner on the same subsets. We certify it either by \cref{thm:transfer}(ii), with surcharge
\[
  \sqrt{\frac{\log64+\log20}{2k}},
\]
equal to the point-mass certificate of (iii) at the same $\lambda^\star$, or by approximate max-information. Each
per-subset certificate is valid at level $0.05$, and \cref{tab:scar} reports their averages. \Cref{tab:rates} reports
how often the per-subset certificate lies below the zero-shot holdout error over 4{,}000 further subsets per $k$.

\paragraph{Multi-round logging.}
The multi-round variant (\cref{tab:library}) logs the pipeline as three rounds on the same validation queries. The
first round chooses the template with top-$m$ retrieval and configuration ($4$, most similar first) held fixed, the
second chooses the retrieval rule, and the third chooses the configuration.

\begin{table}[ht]
  \centering
  \caption{\textbf{The multi-round gaps stay inside the composed certificates.} Zero-shot error averages the four
  templates, and recipe errors are the holdout errors of the 64 recipes. Multi-round gaps are means over 400 subsets
  at $k=64$. By \cref{prop:calculus-formal}, the joint transcript of $R$ rounds leaks
  at most $\log4$ per round. Hence \cref{thm:transfer}(ii) gives the certificate $\sqrt{(R\log4+\log(1/\delta))/(2k)}$
  with $\delta=0.05$ in the last row.}
  \label{tab:library}
  \footnotesize
  \begin{tabular}{lccccc}
    \toprule
    \headrow & & & \multicolumn{3}{c}{\textbf{multi-round gap at {\boldmath$k=64$}}}\\
    \hhline{>{\arrayrulecolor{headc}}--->{\arrayrulecolor{black}}---}
    \headrow\textbf{model} & \textbf{zero-shot error} & \textbf{recipe errors, range (mean)} & $R=1$ & $R=2$ & $R=3$\\
    \midrule
    Qwen2.5-7B-Instruct & 0.373 & 0.092--0.366 (0.192) & 0.004 & 0.005 & 0.015\\
    GPT-4.1-mini & 0.282 & 0.104--0.277 (0.161) & 0.011 & 0.011 & 0.019\\
    \midrule
    certificate, both models & & & 0.185 & 0.212 & 0.236\\
    \bottomrule
  \end{tabular}
\end{table}
\begin{table}[ht]
  \centering
  \caption{\textbf{SCAR's certificate falls below the zero-shot holdout error most often.} Each cell is a percentage over
  4{,}000 validation subsets of size $k$. At $k=256$ each subset is the whole pool, so each entry is $0$ or $100$.
  Rows: SCAR at $\lambda^\star$, and the best-of-64 winner certified by \cref{thm:transfer}(ii) or by approximate max-information (\cref{rem:maxinfo}). Level
  $0.05$ throughout.}
  \label{tab:rates}
  \footnotesize
  \setlength{\tabcolsep}{4pt}
  \begin{tabular}{l ccccc c ccccc}
    \toprule
    \headrow & \multicolumn{5}{c}{\textbf{Qwen2.5-7B} (zero-shot $0.373$)} && \multicolumn{5}{c}{\textbf{GPT-4.1-mini} (zero-shot $0.282$)}\\
    \hhline{>{\arrayrulecolor{headc}}->{\arrayrulecolor{black}}----->{\arrayrulecolor{headc}}->{\arrayrulecolor{black}}-----}
    \headrow $k$ & 16 & 32 & 64 & 128 & 256 && 16 & 32 & 64 & 128 & 256\\
    \midrule
    SCAR & 7.8 & 65.5 & 99.7 & 100 & 100 && 0 & 3.3 & 40.9 & 99.2 & 100\\
    best-of-64, (ii) & 0 & 30.2 & 96.1 & 100 & 100 && 0 & 0 & 5.4 & 89.2 & 100\\
    best-of-64, max-info. & 0 & 0 & 38.8 & 100 & 100 && 0 & 0 & 0 & 8.6 & 100\\
    \bottomrule
  \end{tabular}
\end{table}

\paragraph{Sampling variability and holdout precision.}
Error bars in \cref{fig:real}(a) show $95\%$ normal intervals across the 400 validation draws within each $(M,k)$ cell,
capturing selection variability conditional on the fixed pools. Deployment risk for each recipe is estimated on the
disjoint holdout split with binomial standard error at most $0.016$ ($1{,}024$ queries on Qwen2.5-7B-Instruct) and
$0.022$ ($512$ queries on GPT-4.1-mini); the zero-shot baseline carries standard errors of $0.015$ and $0.020$,
respectively. Because SCAR's average certified margin below zero-shot at $k=32$ on Qwen ($0.010$) is comparable to this
single-holdout standard error, \cref{sec:experiments} highlights per-draw win rates starting at $k=64$, where margins
substantially exceed holdout estimation noise. Across all $30\times400$ resampled selection events per model
(\cref{sec:experiments}; $M=64$ slice in \cref{fig:real}b), the tightest margin below the $95\%$ certificate is $0.019$
(roughly one holdout standard error); at $k=256$, each draw uses the entire validation pool and varies only the
sampled sub-library and random tie-breaking.

\paragraph{Compute and reproduction.}
Embedding took under a minute on one T4 GPU. The Qwen2.5-7B-Instruct evaluation of 87{,}040 prompts ran as nine
parallel H100 jobs and took 11 minutes of wall-clock time, including container build and model loading. The 52{,}224
GPT-4.1-mini calls took 18 minutes. All analyses run on a CPU in seconds from the cached outputs. The supplement
contains the scripts for data preparation, embedding, prompt construction, and evaluation, and the cached model
outputs. Its analysis script, \texttt{real\_pipeline\_analysis.py}, regenerates every number and plot of this section
from the cache.

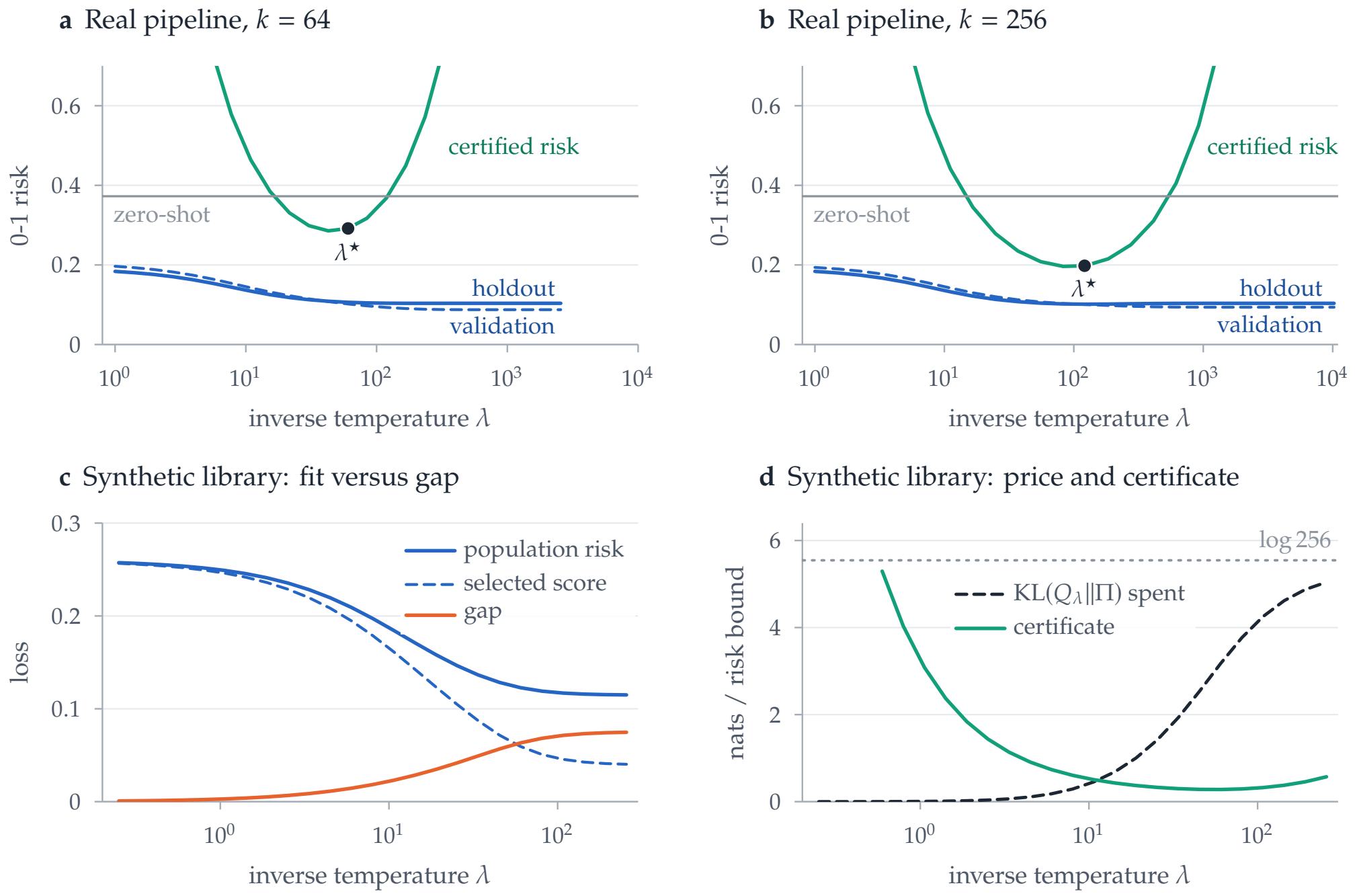
\begin{figure}[t]
  \centering
  \begin{tikzpicture}
\begin{groupplot}[group style={group size=2 by 2, horizontal sep=46pt, vertical sep=50pt},
    paper, width=150pt, height=78pt, xminorticks=false]
  \nextgroupplot[title={\paneltitle{a}{Real pipeline, $k=64$}}, xlabel={inverse temperature $\lambda$}, ylabel={0-1 risk},
    xmode=log, xmin=0.8, xmax=10000, ymin=0, ymax=0.7, ytick={0,0.2,0.4,0.6}]
  \addplot[aquac, data] table[col sep=comma, x=lam, y=scar_cert]{data/rp_scarfront_qwen_k64.csv};
  \node[dlabel aqua, anchor=west] at (axis cs:330,0.5) {certified risk};
  \addplot[bluec, data] table[col sep=comma, x=lam, y=scar_hold]{data/rp_scarfront_qwen_k64.csv};
  \node[dlabel blue, anchor=south east] at (axis cs:2500,0.11) {holdout};
  \addplot[bluec, line width=.8pt, densely dashed] table[col sep=comma, x=lam, y=scar_val]{data/rp_scarfront_qwen_k64.csv};
  \node[dlabel blue, anchor=north east] at (axis cs:2500,0.082) {validation};
  \addplot[mutec, line width=.6pt, domain=0.8:10000, samples=2, forget plot] {0.3726};
  \node[dlabel mute, anchor=north west] at (axis cs:0.9,0.362) {zero-shot};
  \addplot[inkc, only marks, mark=*, mark size=2pt, mark options={draw=white, line width=.5pt}, forget plot] coordinates {(60.52,0.2917)};
  \node[dlabel, text=inkc, anchor=north] at (axis cs:60.52,0.27) {$\lambda^\star$};
  \nextgroupplot[title={\paneltitle{b}{Real pipeline, $k=256$}}, xlabel={inverse temperature $\lambda$}, ylabel={0-1 risk},
    xmode=log, xmin=0.8, xmax=11000, ymin=0, ymax=0.7, ytick={0,0.2,0.4,0.6}]
  \addplot[aquac, data] table[col sep=comma, x=lam, y=scar_cert]{data/rp_scarfront_qwen_k256.csv};
  \node[dlabel aqua, anchor=west] at (axis cs:1000,0.5) {certified risk};
  \addplot[bluec, data] table[col sep=comma, x=lam, y=scar_hold]{data/rp_scarfront_qwen_k256.csv};
  \node[dlabel blue, anchor=south east] at (axis cs:9000,0.11) {holdout};
  \addplot[bluec, line width=.8pt, densely dashed] table[col sep=comma, x=lam, y=scar_val]{data/rp_scarfront_qwen_k256.csv};
  \node[dlabel blue, anchor=north east] at (axis cs:9000,0.086) {validation};
  \addplot[mutec, line width=.6pt, domain=0.8:11000, samples=2] {0.3726};
  \node[dlabel mute, anchor=north west] at (axis cs:0.9,0.362) {zero-shot};
  \addplot[inkc, only marks, mark=*, mark size=2pt, mark options={draw=white, line width=.5pt}] coordinates {(121.05,0.1979)};
  \node[dlabel, text=inkc, anchor=north] at (axis cs:121.05,0.178) {$\lambda^\star$};
  \nextgroupplot[title={\paneltitle{c}{Synthetic library: fit versus gap}}, xlabel={inverse temperature $\lambda$}, ylabel={loss},
    xmode=log, log basis x=10, xmin=0.2, xmax=300, ymin=0, ymax=0.3, ytick={0,0.1,0.2,0.3},
    scaled ticks=false, tick label style={/pgf/number format/fixed},
    legend style={at={(axis cs:290,0.295)}, anchor=north east}]
  \addplot[bluec, data] table[col sep=comma, x=eta, y=population_loss, restrict x to domain=-1:3]{data/scar_temperature_ablation.csv};
  \addlegendentry{population risk}
  \addplot[bluec, line width=.8pt, densely dashed] table[col sep=comma, x=eta, y=empirical_loss, restrict x to domain=-1:3]{data/scar_temperature_ablation.csv};
  \addlegendentry{selected score}
  \addplot[orangec, data] table[col sep=comma, x=eta, y=gap, restrict x to domain=-1:3]{data/scar_temperature_ablation.csv};
  \addlegendentry{gap}
  \nextgroupplot[title={\paneltitle{d}{Synthetic library: price and certificate}}, xlabel={inverse temperature $\lambda$},
    ylabel={nats / risk bound},
    xmode=log, log basis x=10, xmin=0.2, xmax=300, ymin=0, ymax=6.4, ytick={0,2,4,6},
    legend style={at={(axis cs:1.5,5.3)}, anchor=north west}]
  \addplot[inkc, line width=.9pt, densely dashed] table[col sep=comma, x=eta, y=kl_to_uniform, restrict x to domain=-1:3]{data/scar_temperature_ablation.csv};
  \addlegendentry{$\KL(Q_\lambda\|\Pi)$ spent}
  \addplot[aquac, data, restrict y to domain=0:6.4] table[col sep=comma, x=eta, y=certificate, restrict x to domain=-1:3]{data/scar_temperature_ablation.csv};
  \addlegendentry{certificate}
  \addplot[mutec, dotted, line width=.7pt, domain=0.2:300, samples=2, forget plot] {5.545};
  \node[dlabel mute, anchor=south east] at (axis cs:290,5.6) {$\log256$};
\end{groupplot}
\end{tikzpicture}
  \caption{\textbf{SCAR trades fit against the capacity it spends, so its certificate is smallest at an intermediate
  inverse temperature.} (a, b) Qwen2.5-7B-Instruct on Banking77: certified risk of \cref{thm:scar} across $\lambda$,
  with the realized holdout risk, the validation risk, zero-shot risk (gray), and the a-priori $\lambda^\star$ (dot);
  averages over 50 validation subsets. (c, d) A synthetic library of 256 recipes with risks
  $0.08+0.72\,\mathrm{Beta}(2,6)$ and $k=64$ (2{,}500 trials). Raising $\lambda$ lowers the selected score faster than
  the population risk, so the gap grows while the divergence spent approaches $\log256$.}
  \label{fig:app-scar}
\end{figure}

\subsection{Synthetic protocols}\label{app:synthetic}
All synthetic scripts use fixed seeds and run on a CPU.

\emph{Winner's curse} (\cref{fig:synthetic}a). The winner of $N\in\{2,4,\dots,1024\}$ prompts whose losses are
independent Rademacher coordinates mapped to $\{0,1\}$, $k=64$, 4{,}000 trials.

\emph{Rounds} (\cref{fig:synthetic}b). Four edits per round, $R=1,\dots,6$, the winner among $4^R$ terminal prompts on
the same $k=64$ examples, 2{,}500 trials.

\emph{Gate charge} (\cref{fig:synthetic}c). The winner of $N\in\{1,2,4,16,\dots,16384\}$ prompts of risk $1/2$, $k=64$,
4{,}000 trials. The gate, a data-dependent restriction of the candidates (\cref{sec:scar}), contains only the winner.
The certificates are
\[
  \sqrt{\frac{\log(1/\delta)}{2k}}\quad\text{(gate ignored)}\qquad\text{and}\qquad
  \sqrt{\frac{\log N+\log(1/\delta)}{2k}}\quad\text{(gate charged)},
\]
where the gate-charged certificate is both \cref{thm:scar} at $\lambda^\star$ for the point mass and
\cref{thm:transfer}(ii). The max-information variant $\sqrt{(\log(N/\beta)+\log(1/\delta'))/(2k)}$ with
$\beta=\delta'=\delta/2$ covers in every trial.

\emph{Brittle library} (\cref{fig:synthetic}d). The library has 50 stable and 150 brittle prompts. A stable prompt
has risk drawn from $\mathcal N(0.30,0.015^2)$ clipped to $[0.22,0.38]$, and its losses equal the risk plus
$\mathrm{Unif}[-0.035,0.035]$. A brittle prompt has risk drawn from $\mathcal N(0.45,0.035^2)$ clipped to $[0.32,0.58]$
and Bernoulli losses. The prior is uniform over the 200 prompts, and each $k$ uses 3{,}000 trials. The first-order
selector keeps the lowest score and reports $\sqrt{(\log200+\log(1/\delta))/(2k)}$ above it. The second-order selector
minimizes the certificate of \cref{thm:second} over its grid $\Gamma$. Both certificates held in at least $99.9\%$ of
trials, and \cref{tab:second} lists their sizes.

\begin{table}[ht]
  \centering
  \caption{\textbf{Charging variance instead of range tightens the certificate from 64 examples on.} Brittle library,
  means over 3{,}000 trials. Risk is the population risk of the selected prompt. Bold marks the tighter of the two
  certificates at each $k$.}
  \label{tab:second}
  \footnotesize
  \begin{tabular}{lccccccc}
    \toprule
    \headrow $k$ & 8 & 16 & 32 & 64 & 128 & 256 & 512\\
    \midrule
    first order, risk & 0.435 & 0.426 & 0.410 & 0.328 & 0.273 & 0.267 & 0.266\\
    first order, certificate & \textbf{0.749} & \textbf{0.639} & \textbf{0.569} & 0.509 & 0.445 & 0.393 & 0.356\\
    second order, risk & 0.395 & 0.297 & 0.268 & 0.267 & 0.267 & 0.266 & 0.266\\
    second order, certificate & 1.467 & 0.945 & 0.620 & \textbf{0.446} & \textbf{0.358} & \textbf{0.313} & \textbf{0.290}\\
    \bottomrule
  \end{tabular}
\end{table}

\subsection{Real-LLM stress tests}\label{app:api}
\begin{figure}[ht]
  \centering
  \begin{tikzpicture}
\begin{groupplot}[group style={group size=2 by 1, horizontal sep=40pt}, paper, width=130pt, height=62pt, xminorticks=false]
  \nextgroupplot[title={\paneltitle{a}{Leaked items inflate validation accuracy}},
    xlabel={validation items leaked into the demonstrations},
    ylabel={accuracy}, xtick={0,2,4,8}, xmin=-0.5, xmax=8.5, ymin=0, ymax=1.06, ytick={0,0.5,1}, legend pos=north west]
  \addplot[orangec, data, sq] table[col sep=comma, x=leak_count, y=validation_accuracy]{data/openai_llm_leaky_context.csv};
  \addlegendentry{validation (reused)}
  \addplot[bluec, data, dot] table[col sep=comma, x=leak_count, y=holdout_accuracy]{data/openai_llm_leaky_context.csv};
  \addlegendentry{holdout}
  \nextgroupplot[title={\paneltitle{b}{SCAR over the four recipes}}, xlabel={inverse temperature $\lambda$},
    ylabel={accuracy / normalized KL},
    xmode=log, log basis x=10, xmin=0.2, xmax=300, ymin=0, ymax=1.06, ytick={0,0.5,1}, legend style={at={(0.98,0.55)}, anchor=east}]
  \addplot[orangec, data] table[col sep=comma, x=eta, y=validation_accuracy, restrict x to domain=-1:3]{data/openai_llm_leaky_scar_temperature.csv};
  \addlegendentry{validation}
  \addplot[bluec, data] table[col sep=comma, x=eta, y=holdout_accuracy, restrict x to domain=-1:3]{data/openai_llm_leaky_scar_temperature.csv};
  \addlegendentry{holdout}
  \addplot[inkc, line width=.9pt, densely dashed] table[col sep=comma, x=eta, y expr=\thisrow{kl_to_uniform}/1.3862944, restrict x to domain=-1:3]{data/openai_llm_leaky_scar_temperature.csv};
  \addlegendentry{$\KL/\log4$}
\end{groupplot}
\end{tikzpicture}
  \caption{\textbf{On a real model, leaked validation items inflate validation accuracy and drive holdout accuracy to
  zero.} GPT-5.2 on the nonce-label task: (a) validation and holdout accuracy of the four recipes; (b) SCAR over the
  four recipes, with the divergence normalized by the menu price $\log4$.}
  \label{fig:app-leak}
\end{figure}
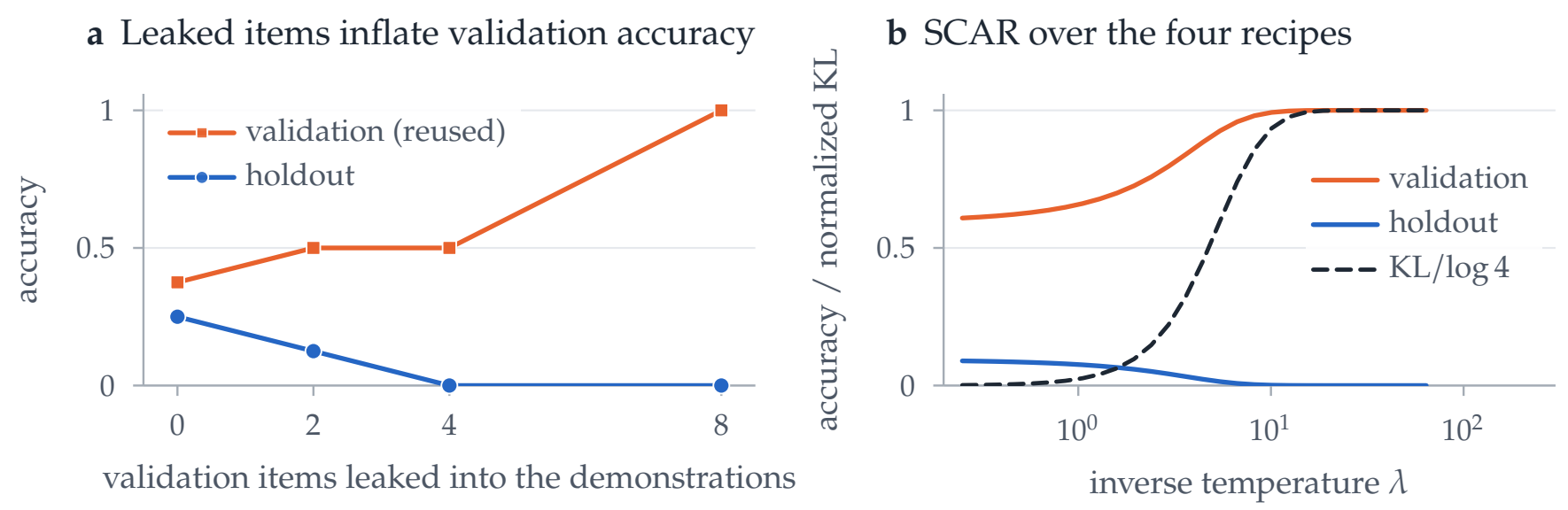

To complement the quantitative Banking77 study with direct mechanism probes on a frontier model, we run two controlled
protocols with GPT-5.2 via the OpenAI API using synthetic nonce labels. The first is a semantic null control: eight
prompt recipes all achieve $1.0$ accuracy on both an 8-item validation split and an 8-item holdout, confirming zero
empirical optimism when true risk vanishes across the library. The second is a deliberate validation-leakage stress
test: four recipes leak $0$, $2$, $4$, or $8$ validation items directly into the prompt's demonstrations. As leaked
items accumulate, validation accuracy climbs to $1.0$ while holdout accuracy collapses to $0$ (\cref{fig:app-leak}a);
increasing SCAR's inverse temperature $\lambda$ concentrates posterior mass onto the fully leaked recipe and drives
$\KL(Q_\lambda\|\Pi)$ toward the $\log4$ menu ceiling (\cref{fig:app-leak}b), illustrating how the logged divergence
automatically tracks severe context overfitting on a live model.

\needspace{6\baselineskip}%
\section{Using the certificates in practice}\label{app:practice}

\subsection{Accounting rules}
\Cref{tab:accounting} translates \cref{prop:calculus-formal} into a stage-by-stage reference sheet, pairing each
closed-form capacity charge with the runtime log entry that makes it auditable without inspecting internal model or
retriever weights.

Realized-gap capacity is the maximal leakage of \citet{issa2020operational}: $e^{\Leak}$ is the largest multiplicative
factor by which observing the transcript can boost the probability of guessing any randomized function of the store, and
consequently bounds the real-law probability of any joint event $(S,T)\in E$ by $e^{\Leak}$ times the worst-case
ghost-section probability \citep{esposito2021generalization}.

\begin{table}[ht]
  \centering
  \caption{\textbf{Each pipeline stage has a capacity charge, and its log makes the charge auditable.} Expected
  capacity enters \cref{thm:transfer}(i), realized-gap capacity (maximal leakage) enters (ii), and the logged divergence
  enters (iii) and \cref{thm:scar}.}
  \label{tab:accounting}
  \footnotesize
  \renewcommand{\arraystretch}{1.15}
  \begin{tabular}{L{0.25\linewidth}L{0.155\linewidth}L{0.165\linewidth}L{0.30\linewidth}}
    \toprule
    \headrow\textbf{stage} & \textbf{expected capacity} & \textbf{realized-gap capacity} & \textbf{required log}\\
    \midrule
    choose one of $M$ fixed templates or recipes & $\log M$ & $\log M$ & candidate set, fixed before scoring; chosen identifier\\
    release scores $a(S)+\mathcal N(0,\rho^{-1}I)$ & $\rho\Delta^2/2$ & --- & noise precision and sensitivity bound $\Delta$\\
    Gibbs or SCAR sampling from prior $\Pi$ & $\E\,\KL(Q_S\|\Pi)$ & at most $\log|\supp\Pi|$, and at most $\lambda$ without a gate (\cref{app:calculus-rules}) & prior, temperature, posterior weights\\
    data-dependent gate $A(S)$ & $\E\log(1/\Pi(A))$, inside the KL & --- & gate rule and the prior mass of every realized gate\\
    $R$ adaptive rounds & $\sum_rC_r$ & $\sum_r\kappa_r$ & per-round candidates and choices\\
    formatting, truncation, deduplication & $0$ & $0$ & none (post-processing)\\
    evaluation on a split no stage read & $0$ & $0$ & split identifier and a no-reuse guarantee\\
    \bottomrule
  \end{tabular}
\end{table}

\subsection{Assumptions in deployment}
\Cref{tab:assumptions} lists why each assumption is used, when it holds in deployment, and how to weaken or check it.
The bounded-loss row has a variance-adaptive form. For losses in $[0,1]$ with variance at most $s^2$ and uniform weights
over $k$ items, the ghost-cumulant condition of \cref{prop:subgamma} holds with $v=s^2/k$ and $b=1/(3k)$ (Bernstein's
inequality).

\begin{table}[ht]
  \centering
  \caption{\textbf{Each assumption maps to a deployment practice and to a way to weaken or check it.}}
  \label{tab:assumptions}
  \footnotesize
  \renewcommand{\arraystretch}{1.15}
  \begin{tabular}{L{0.165\linewidth}L{0.18\linewidth}L{0.27\linewidth}L{0.255\linewidth}}
    \toprule
    \headrow\textbf{assumption} & \textbf{why it is used} & \textbf{when it holds in deployment} & \textbf{how to weaken or check}\\
    \midrule
    frozen learner after $T$ & separates prompt adaptivity from training & a version-pinned model through an API or fixed weights during the search window & record model-in-the-loop edits in $T$\\
    bounded losses & sub-Gaussian ghost concentration with $\sigma^2=\tfrac14$ & misclassification, exact-match failure, task failure & clip scores; use \cref{prop:subgamma} for variance\\
    complete transcript & makes the price auditable & retrieval stacks that log candidates, scores, filters, seeds & audit with \cref{app:checklist}\\
    i.i.d.\ or block store & ghost concentration at $\keff$ & independent items or independent document blocks, with chunks of one document grouped as a block & block by document or source; evaluate on a fresh split\\
    prior fixed in advance & PAC-Bayes change of measure & a base retriever or a recipe library chosen before scoring & condition only on public metadata or another split; log $\Pi(A)$\\
    \bottomrule
  \end{tabular}
\end{table}

The store assumption allows dependence inside public blocks (case (b) of \cref{ass:store}). Scraped corpora, however,
contain near-duplicates and cross-document dependence. Where such dependence crosses blocks, it requires mixing or
graph-dependent concentration, which lies outside the present results. An independent evaluation split avoids selection reuse; concentration within that split still depends
on its sampling law.

\subsection{An audit checklist}\label{app:checklist}
A practitioner should be able to answer these questions before using a selected score as evidence about
deployment risk:
\begin{enumerate}[leftmargin=2em,label=\arabic*.]
  \item Which data were read to construct the final transcript, and which split scores it?
  \item How many templates, recipes, retrievers, and rerankers were tried, and was the candidate set fixed before scoring?
  \item Were validation examples, labels, critiques, or manual edits reused across rounds?
  \item Is the final context deterministic top-$k$, a randomized draw, or a posterior sample, and what prior defines its KL?
  \item Which gates restricted the candidates, and what was their prior mass?
  \item Is there a split that no stage has read?
\end{enumerate}
A complete report states the scoring sample size $k$ (or $\keff$), the number of demonstrations per prompt, the capacity ledger (the per-stage
charges of \cref{tab:accounting}), the certificate with its level, and the split hygiene (which split each stage
read).

\subsection{Limitations}\label{app:limitations}
Our framework treats the prompted model as a frozen map from transcript to loss, isolating the statistical cost of
adaptive evaluation from pretraining or weight-update dynamics. Non-vacuous certification requires both an auditable
selection capacity and a store that is conditionally i.i.d.\ or independent across public blocks; when selection over a
pool of size $n$ is unrestricted and deterministic, any distribution-free certificate must span at least $1-2k/n$
(\cref{rem:cherry}).

Because \cref{thm:transfer} protects uniformly against the independent equal-risk configuration of \cref{thm:lower}, its
surcharges exceed observed selection gaps by factors of roughly three to eight for $k\le128$ (and nine to twelve at
$k=256$) on correlated recipe libraries such as Banking77, where candidates share queries and differ substantially in
true risk. Empirically, our real-LLM evaluation focuses on one benchmark (Banking77), two model families, and a
64-recipe design space resampled from fixed validation and holdout pools, complemented by synthetic benchmarks that
isolate individual mechanisms under exact ground-truth laws.

\subsection{Broader impact}\label{app:impact}
Auditable capacity ledgers are designed to curb inadvertent overclaiming in retrieval-augmented generation, few-shot
benchmarking, and automated prompt optimization whenever evaluation splits are reused across design iterations. The
primary operational risk is performative reporting, in which a system publishes a small nominal capacity budget while
omitting informal or unlogged rounds of prompt search. Enforcing transcript completeness (\cref{def:transcript}) alongside the split-hygiene
audit in \cref{app:checklist} directly addresses this failure mode. All experiments rely on synthetic benchmarks and a
public intent-classification dataset distributed under CC BY 4.0.

\end{document}